\documentclass{article}

\usepackage{microtype}
\usepackage{graphicx}
\usepackage{subfigure}
\usepackage{booktabs} 
\usepackage{hyperref}
\usepackage{fullpage}

\usepackage{hyperref}
\hypersetup{
	colorlinks=true,
	linkcolor=blue,
	citecolor=blue,
	filecolor=blue,      
	urlcolor=blue,
}

\usepackage{amsmath,fullpage}
\usepackage{amssymb}
\usepackage{mathtools}
\usepackage{amsthm}
\usepackage{float}

\usepackage{booktabs}
\usepackage{multirow}
\usepackage{diagbox}
\usepackage{hhline}
\usepackage{bbm}
\usepackage{extarrows}
\usepackage{natbib}
\usepackage[capitalize,noabbrev]{cleveref}
\usepackage[textsize=tiny]{todonotes}
\usepackage{algorithm}%
\usepackage{algorithmicx}%
\usepackage{algpseudocode}%

\newtheorem{theorem}{Theorem}
\newtheorem{proposition}[theorem]{Proposition}
\newtheorem{lemma}{Lemma}%
\newtheorem{remark}{Remark}%
\newtheorem{definition}{Definition}%

\newcommand{\U}{{\mathbf U}}
\newcommand{\X}{{\mathbf X}}
\newcommand{\Y}{{\mathbf Y}}
\newcommand{\Z}{{\mathbf Z}}
\newcommand{\F}{{\mathbb F}}
\newcommand{\E}{{\mathbb E}}
\newcommand{\R}{{\mathbb R}}
\newcommand{\I}{{\mathcal I}}
\newcommand{\Ss}{{\mathcal S}}
\newcommand{\T}{{\mathcal T}}
\newcommand{\vv}{\boldsymbol{v}}
\newcommand{\vu}{\boldsymbol{u}}
\newcommand{\diag}{{\rm diag}}
\newcommand{\sign}{{\rm sign}}

\allowdisplaybreaks
\numberwithin {equation} {section}

\title{Fairness via Independence: A (Conditional) Distance Covariance Framework}

\author{Ruifan Huang\thanks{School of Mathematics and Statistics, Huazhong University of Science and Technology, Wuhan, Hubei, China. Email: ruifanhuang@hust.edu.cn.}, 
Haixia Liu\thanks{School of Mathematics and Statistics  \& Institute of Interdisciplinary Research for Mathematics and Applied Science \& Hubei Key Laboratory of Engineering Modeling and Scientific Computing, Huazhong University of Science and Technology, Wuhan, Hubei, China. Email: liuhaixia@hust.edu.cn. The work of H.X. Liu was supported in part by NSFC 11901220, Interdisciplinary Research Program of HUST 2024JCYJ005, National Key Research and Development Program of China 2023YFC3804500. The authors gratefully acknowledge the computational resources provided by the HPC Platform of Huazhong University of Science and Technology.}}

\begin{document}
\maketitle

\begin{abstract}
    We explore fairness from a statistical perspective by selectively utilizing either conditional distance covariance or distance covariance statistics as measures to assess the independence between predictions and sensitive attributes. We boost fairness with independence by adding a distance covariance-based penalty to the model's training. Additionally, we present the matrix form of empirical (conditional) distance covariance for parallel calculations to enhance computational efficiency. Theoretically, we provide a proof for the convergence between empirical and population (conditional) distance covariance, establishing necessary guarantees for batch computations. 
Through experiments conducted on a range of real-world datasets, we have demonstrated that our method effectively bridges the fairness gap in machine learning. Our code is available at \url{https://github.com/liuhaixias1/Fair_dc/}.
\end{abstract}

% \keywords{Fairness Learning, Distance Covariance, Convergence in Probability}

\section{Introduction}

Despite the success of modern deep neural networks (DNNs) applied to various tasks in recent years, it may raise many ethical and legal concerns during model training. In real-world classification and decision-making tasks, biased datasets can influence machine learning models, resulting in unfair predictions \cite{bellamyAIFairness3602018}.
With some online models and algorithms, unfair prediction results can lead users to make biased choices, and behavioral biases can create even more biased data, creating a cycle of bias
\cite{mehrabiSurveyBiasFairness2022}.
Therefore, ensuring fairness is crucial when applying machine learning to tasks like computer vision, natural language processing, classification, and regression, requiring careful consideration of ethical and legal risks.

Fairness in machine learning can be broadly categorized into two levels: (1) group-level fairness, emphasizing equitable treatment among different groups, and (2) individual-level fairness, with the goal of providing similar predictions for similar individuals. 
In this paper, our focus is on group-level fairness. A fair machine learning model should avoid producing biased outputs based on sensitive attributes such as ethnicity, gender, and age. While these sensitive attributes may not be explicitly present in the training data features, deep learning models often work with high-dimensional and complex data, which contains a wealth of information. Some of this information may inadvertently correlate with sensitive attributes and result in biased outcomes \citep{kimLearningNotLearn2019, parkLearningDisentangledRepresentation2021}.

A natural and intuitive idea to enhance fairness is to promote the statistical independence between the model predictions $(\hat \Y)$ and sensitive attributes $(\Z)$. When treating model predictions and sensitive attributes as random variables, the fairness criterion Demographic Parity (DP) is satisfied only when $\hat \Y$ and $\Z$ are independent. Similarly, the fairness criterion Equalized Odds (EO) holds when there is conditional independence between $\hat \Y$ and $\Z$ given the ground truth labels $(\Y)$.
Measuring the independence of random variables is a fundamental goal in statistical data analysis, and various methods are available for this purpose, including Pearson correlation, Hirschfeld-Gebelein-R\'enyi maximal correlation (HGR), and Mutual Information (MI).

Among these methods, Pearson correlation coefficients are commonly used to measure the correlation between random variables, and has been extensively researched in the context of fair machine learning \citep{zafarFairnessConstraintsMechanisms2017a, zhaoFairMetaLearningFewShot2020}, yielding positive results and providing evidence for the effectiveness of this idea. 
However, Pearson correlation coefficients are only effective for capturing linear correlations and struggle to capture non-linear relationships which is also important in fairness learning. Additionally, relationships in complex datasets are often non-linear and can be further complicated by non-linear transformations performed by machine learning models. Therefore, a more robust statistical measure is needed to bridge the fairness gap.

In this paper, we introduce so-called distance covariance and conditional distance covariance based methods for fair classification. Distance covariance (DC) \citep{szekelyMeasuringTestingDependence2007} and Conditional Distance Covariance (CDC) \citep{wang2015conditional} are robust methods for quantifying both linear and non-linear correlations between two or three random variable (vectors). A smaller (conditional) distance covariance value indicates a weaker relationship between the random variables, and it equals zero if and only if the two random variables are (conditional) independent. 
However, directly computing DC or CDC can be challenging as it requires knowledge of the analytical form of the distribution function and involves integration. To overcome this limitation, we employ empirical (conditional) distance covariance, which can be computed directly from samples, as a surrogate loss for fairness constraint. We incorporate them as a regularization term during the model training process.

Among the estimations of DC, CDC, HGR, and MI used to characterize the independence, empirical (conditional) distance covariance (DC) stands out due to the following reasons, considering that we can only rely on estimates instead of exact computations: (1) MI-related methods \citep{gupta2021controllable} are limited to DP fairness. For EO fairness notions, conditional mutual information should be employed instead. Furthermore, MI-related methods typically begin by estimating upper and lower bounds. The distributions used to calculate these bounds are assumed as a standard normal distribution or a diagonal gaussian distribution, with the mean and variance parameterized by neural networks. However, this assumption may not always be appropriate. Furthermore, the process is generally time-consuming in training (see Table \ref{tab:time}).  
(2) The estimation of HGR \citep{maryFairnessAwareLearningContinuous2019a} heavily relies on the approximation space. In practical computations, it's infeasible to explore all functions. Classical methods approximate HGR by constraining it to linear spaces or Reproducing Kernel Hilbert Spaces. Consequently, there may be a noticeable difference between the approximated and true HGR.  (3) The empirical DC and CDC together are good choices for DP \& EO fairnesses. The population CDC is equivalent to independence (see Theorem 3 (i) in \cite{szekelyMeasuringTestingDependence2007}), which is used to characterize DP fairness. Similarly, the population DC is equivalent to conditional independence (see Theorem 1(i) in \cite{wang2015conditional}), which is used to characterize EO fairness. The discrepancy between the empirical DC and the population DC is attributed to statistic error, which can be reduced by increasing the sample size. For CDC, the discrepancy between the empirical CDC and the population CDC can also be reduced by increasing the sample size if we set the bandwidth of the Gaussian kernel tends to zeros when the sample size tends to infinity. Additionally, (empirical) DC/CDC is very easy to apply to tensor-valued data (including vector, matrix or multi-way tensor). Naturally extends to measure dependence between two random vectors of possibly different dimensions. You can measure the dependence between a variable in $\R^{d_1}$ and another in $\R^{d_2}$ just as easily as between two scalars, where $d_1$ and $d_2$ may be different. The algorithm remains the same: compute all pairwise distances within each sample set.  Theoretically, MI also applies to vectors. However, the curse of dimensionality makes estimation even more difficult and unreliable in high dimensions. For HGR, the classical definition is for single variables. Generalizing it to multivariate data necessitates an intermediate mapping from a vector to a scalar. For detailed comparisons with state-of-the-art methods, please refer to Tables \ref{tab:ComparisionWithSota} and \ref{tab:time}.

In summary, our contributions can be outlined as follows. 

(1) {\bf We introduce the utilization of empirical (conditional) distance covariance as a feasible penalty term in the machine learning process to promote independence.} 
To the best of our knowledge, we are the first to incorporate conditional distance covariance with machine learning.

(2) {\bf we present the matrix form of empirical (conditional) distance covariance for parallel calculations to enhance computational efficiency.}

(3) {\bf We provide theoretical proofs of the convergence in probability between the population and empirical (conditional) distance covariance with respect to the sample size. } Although the empirical distance covariance converges almost surely to the population distance covariance or empirical CDC converges in probability to the population CDC, these properties hold for large samples as the sample size approaches infinity. In deep learning, various batch gradient descent methods are commonly employed due to limitations on GPU memory. These results provide essential theoretical insights for small-batch computations and indicate approximately what batch size is necessary to achieve the desired level of accuracy.

(4) {\bf The numerical experiments conducted in this paper provide evidence of the versatility of our proposed methods, showcasing their applicability across diverse datasets and tasks, while achieving competitive performance. } Our method does not necessitate any prior knowledge about the model or existing bias, rendering it widely applicable. Furthermore, it is not restricted to binary sensitive attributes, but can be extended to encompass any number of sensitive attributes or subgroups. This adaptability stems from the fact that (conditional) distance covariance can be applied to random vectors of any dimension. 
Moreover, our method is a plug-and-play approach, allowing for easy application across various domains, datasets, and neural network models with minimal computational cost. This characteristic enhances its practicality for real-world applications.

Throughout the paper, we use bold capital letter, capital letter to represent random vectors, sample matrices, respectively.

\section{Related Works}

Fair machine learning is a significant field that has witnessed the advancement of various methods. In this section, we will primarily focus on the related work in fair classification.
Existing fair machine learning methods can be broadly classified into three stages of processing: pre-processing \citep{calmonOptimizedDataPreProcessing2017,xuFairganFairnessawareGenerative2018,sattigeriFairnessGANGenerating2019, ramaswamyFairAttributeClassification2021}, post-processing \citep{hardtEqualityOpportunitySupervised2016,bolukbasiManComputerProgrammer2016,pleissFairnessCalibration2017,
mehrabiAttributingFairDecisions2022,alghamdi2022beyond}, and in-processing \citep{zafarFairnessConstraintsMechanisms2017a,
zhaoFairMetaLearningFewShot2020,maryFairnessAwareLearningContinuous2019a, leeMaximalCorrelationFramework2022,moyerInvariantRepresentationsAdversarial2019, songLearningControllableFair2020a, creagerFlexiblyFairRepresentation2019,chuangFairMixupFairness2020,parkFairContrastiveLearning2022,
liuFairRepresentationLearning2022,guoLearningFairRepresentations2022,lowy2022stochastic}.

Our method belongs to the in-processing category, where the concept of correlation is employed as a surrogate loss for fairness criteria.
In addition to employing Pearson correlation and distance covariance, HGR maximum correlation or mutual information can also be utilized as a measure of the statistical independence between random variables, serving as a surrogate constraint for fairness. 
\cite{maryFairnessAwareLearningContinuous2019a} and  \cite{leeMaximalCorrelationFramework2022} utilize HGR maximal correlation to measure the dependence between the model output and sensitive attributes as a penalty term within their frameworks to enforce fairness.
However, it's important to note that performing an exact computation of the HGR maximal correlation is impossible. To mitigate this issue, both papers utilize alternative approaches such as kernel density estimation (KDE) or Soft-HGR as approximations to calculate the desired measure.

%3333333333333333333333
From the perspective of information theory, mutual information(MI) and its variants can also serve as proxy for fairness criteria, aiming to reduce MI to minimize the model's dependence on sensitive attributes \citep{moyerInvariantRepresentationsAdversarial2019, songLearningControllableFair2020a, creagerFlexiblyFairRepresentation2019}.
However, calculating MI can be challenging due to the lack of probability density function \citep{poole2019variational}. 
Consequently, instead of directly estimating mutual information, they approximate it by their empirical version, upper/lower bounds or some variational methods. Moyer et al. \cite{moyerInvariantRepresentationsAdversarial2019} introduced invariant representations without adversarial Training through MI, transplanting adversarial losses from generative literature to encoder/decoder settings. Song et al. \cite{songLearningControllableFair2020a} proposed an MI-motivated objective for learning maximally expressive representations subject to fairness constraints, which can optimize approximations to Lagrangian dual objective. In the FERMI framework \citep{lowy2022stochastic}, the authors propose an objective function that incorporates the exponential R\'enyi mutual information as a regularizer to balance fairness and accuracy. They employ an empirical version of the exponential R\'enyi mutual information to calculate the regularization term. Notably, they also introduce a first stochastic in-processing fairness algorithm and establish guaranteed convergence for it. 
Another fairness representation learning method based on MI is FCLR \citep{gupta2021controllable}. FCLR tackles parity control by bounding mutual information using contrastive information estimators, operating under the assumption of normality.

A prominent strategy for ensuring algorithmic fairness involves kernel-based methods. In this paradigm, fair regression and classification are achieved by incorporating the Hilbert-Schmidt Independence Criterion (HSIC) \citep{gretton2005kernel} as a fairness regularizer. This approach offers two significant advantages: it simplifies the optimization problem and readily accommodates multiple sensitive variables simultaneously. By leveraging kernel functions instead of linear formulations, these methods can effectively capture and mitigate complex, nonlinear dependencies, thereby enhancing their applicability to real-world data \citep{perez2017fair}. The kernel-based framework continues to evolve, with recent advancements including fair kernel clustering \citep{zhou2024fair} and fair kernel regression \citep{perez2023fair}.

In parallel, another line of research focuses on fair representation learning, which employs various fairness regularizers to learn data representations that are invariant to sensitive attributes. This category includes a diverse set of techniques, such as: 
Adaptive Weights \citep{chai2022fairness}, 
Contrastive Estimation \citep{shen2021contrastive}, and 
Adversarial Learning \citep{zhang2018mitigating,lahotifairnessdemographics2020}.

Meanwhile, fairness can also effectively improved by augmenting data and applying fairness regularization to its outputs. \cite{chuangFairMixupFairness2020} proposed FairMixup, which regularizes the estimated paths of mixup sample to improve both generalization and fairness of model.

While distance covariance has been introduced to machine learning in prior works, applied in areas such as  few-shot learning, interpretability, robust learning, and fairness, to the best of our knowledge, conditional distance covariance is being applied to machine learning for the first time. Next, we offer a brief overview of two relevant works involving distance covariance in fairness learning.

Both  \cite{liuFairRepresentationLearning2022} and \cite{guoLearningFairRepresentations2022} have utilized distance covariance in fairness representation tasks. In their work,  \cite{liuFairRepresentationLearning2022} introduced population distance covariance as an alternative to MI and demonstrated its asymptotic equivalence to MI. However, computing either population distance covariance or MI requires knowledge of the analytical form of the respective distribution functions, which necessitates prior knowledge of the distribution function or density function. To address this, they assumed that the random variables follow a multivariate normal distribution and employed the Variational Autoencoder technique \citep{kingmaAutoencodingVariationalBayes2013} to fit the normal distribution. Nevertheless, the assumption of a normal distribution can be overly restrictive and may not hold for real-world data, potentially leading to errors in the analysis. 
\cite{guoLearningFairRepresentations2022} propose a method for learning fair representations by incorporating graph Laplacian regularization. They explore the relationship between graph regularization and distance correlation, highlighting its significance in the context of fairness representation tasks. Similarly to the approach presented by \cite{liuFairRepresentationLearning2022},  \cite{guoLearningFairRepresentations2022} employ a Variational Autoencoder (VAE) to learn representations, which relies on assumptions regarding the underlying distribution of the data.

Finally, we summarize the differences between our method and counterparts regularization methods based on statistical independence in \cref{tab:ComparisionWithSota}.

\begin{table}[htbp]
	\caption{Comparison of state-of-the-art (SOTA) surrogate fairness constraint methods.}
	\label{tab:ComparisionWithSota}
	% \vskip -0.15in
        \centering
			\begin{sc}
				\begin{tabular}{ccccc} 
					\toprule
					Method                                                              & Statistic                    & Fairness             & \begin{tabular}[c]{@{}c@{}}Multiple\\Attributes\end{tabular} & \begin{tabular}[c]{@{}c@{}}Continuous\\Attributes\end{tabular}  \\ 
					\hline
					HGR \citep{maryFairnessAwareLearningContinuous2019a} & HGR maximal correlation      & DP \& EO 			 & Yes      & Yes               \\
                    Soft-HGR \citep{leeMaximalCorrelationFramework2022} & HGR maximal correlation      & DP \& EO 			 & Yes      & Yes               \\
					FERMI \citep{lowy2022stochastic}                     & Rényi Mutual Information     & DP      			 & Yes       & No            \\
					FCLR \citep{gupta2021controllable}                   & Mutual Information           & DP                   & Yes      & Yes                 \\
					FairDisCo \citep{liuFairRepresentationLearning2022}  & Population DC                & DP                   & Yes      & No            \\
					Dist-Fair \citep{guoLearningFairRepresentations2022} & Empirical DC                 & DP                   & Yes      & Yes         \\
					FairMixup \citep{chuangFairMixupFairness2020}        & Paths of Mixup               & DP \& EO		     & No       & No              \\
					Ours                                               & Empirical DC \& CDC 		  & DP \& EO 			 & Yes      & Yes                 \\
					\bottomrule
				\end{tabular}
			\end{sc}
\end{table}
\section{Distance Covariance and Conditional Distance Covariance}
In this section, we begin by introducing two statistics to characterize independence and conditional independence. In Subsection \ref{sec:dc}, we introduce the concepts of distance covariance, conditional distance covariance and their empirical versions. 
Then we will present the theoretical results in Subsection \ref{sec:convergence}, which is followed by matrix form in Subsection \ref{sec:matrix_form}. 

Throughout of this paper, we use $\mathbf Y\in\mathbb{R}^p$, $\Z\in\mathbb{R}^q$ to represent two continuous random vectors and $\mathbf U\in \mathbb{R}^r$ to be a random vector (whether continuous or discrete). Let $g_{\Y,\Z}(y,z)$ be the joint probability density function, and $g_{\Y}(y), g_{\Z}(z)$ be marginal probability density functions. 
Let  $f_{\Y,\Z}(t,s) = \int e^{i(t^\top y+s^\top z)}g_{\Y,\Z}(y,z)dydz$ be the joint characteristic function, and $f_{\Y}(t) = \int e^{it^\top y}g_{\Y}(y)dy$, $f_{\Z}(s) = \int e^{is^\top z}g_{\Z}(z)dz$ be marginal characteristic functions corresponding to $\Y,\Z$. Denote $g_{\Y, \Z\vert\U}(y,z,u)$, $g_{\Y \vert \U}(y,u)$ and $g_{\Z \vert\U }(z,u)$ be the conditional joint probability density function and conditional marginal probability density functions, respectively. 
For vectors $t \in \mathbb{R}^p$ and $s \in \mathbb{R}^q$, the conditional joint characteristic function of $\Y$, $\Z$ given $\U$ is defined as $f_{\Y, \Z\vert \U}(t, s,u) =  \int e^{i(t^\top y+s^\top z)}g_{\Z,\Y \vert \U}(y,z,u)dydz$. In addition, $f_{\Y \vert \U}(t,u) = \int e^{it^\top y}g_{\Y \vert \U}(y,u)dy$, $f_{\Z\vert \U }(s,u) = \int e^{is^\top z}g_{\Z\vert \U }(z,u)dz$ be the conditional marginal characteristic functions of $\Y$,  $\Z$ given $\U$, respectively.
%######################################################
\subsection{Definitions}
\label{sec:dc}
In the following subsection, we will give the definitions of distance covariance (DC), conditional distance covariance (CDC) and their empirical versions.

Distance covariance $\mathcal{V}^2(\Y,\Z)$ is defined as 
$\mathcal{V}^2(\Y,\Z) = \iint_{\mathbb{R}^{p+q}}{| f_{\Y,\Z}(t,s) - f_{\Y}(t)f_{\Z}(s) |^2 w(t,s)}\,dtds$, 
where the weight $w(t,s) = (c_pc_q|t|^{1+p}|s|^{1+q})^{-1}$ with $c_d= \frac{\pi^{(1+d)/2}}{\Gamma((1+d)/2)}$ and $\Gamma$ being the gamma function (Definition 2 in \cite{szekelyMeasuringTestingDependence2007}).

Similar to distance covariance, conditional distance covariance $\mathcal{D}^2(\Y,  \Z \vert\U)$ (Definition 1 in \cite{wang2015conditional}) is defined as
$\mathcal{D}^2(\Y,\Z \vert \U=u ) = \iint{| f_{\Y, \Z \vert\U}(t,s,u) - f_{\Y\vert \U }(t,u)f_{\Z \vert \U}(s,u) |^2 w(t,s)}\,dtds$, Where $w(t,s)$ refers to the same weight function as defined in the definition of distance covariance. 

To measure the conditional independence between random vectors $\Y$ and $\Z$ for $\U$ given, denoted as $\Y\perp  \Z \vert \U$, we can use the quantities $\Ss_a(\Y,\Z,\U) = \E[\mathcal{D}^2(\Y, \Z \vert \U)a(\U)]$ (Equation (9) in \cite{wang2015conditional}), where $a(\cdot)$ is a nonnegative function with the same support as the probability density function of $\U$. Consequently, $\Y\perp  \Z \vert \U$ if and only if $\Ss_a(\Y,\Z,\U)=0$ \citep{wang2015conditional}. 

In practical scenarios, obtaining the probability density functions (cumulative distribution functions) can be challenging or even impossible. Consequently, directly acquiring the distance covariance becomes unfeasible. To address this issue, we resort to utilizing the empirical DC instead, which is guaranteed to be non-negative by Theorem 1 in \cite{szekelyMeasuringTestingDependence2007}.

Consider the observed samples $ {(Y_i, Z_i), i=1,\dots,n}$, which are drawn from the joint distribution of random vectors $(\Y,\Z) \in \mathbb{R}^p\times \mathbb{R}^q$. Let $Y=[Y_1,\cdots,Y_n]$ and $Z=[Z_1,\cdots,Z_n]$ represent the two sample matrices. The empirical distance covariance \citep{szekelyMeasuringTestingDependence2007} is then defined as follows:

\begin{definition}[Definition 4 in \cite{szekelyMeasuringTestingDependence2007}]\label{def:dcov}
	The empirical distance covariance $\mathcal{V}^2_n(Y,Z)$ is the product of the corresponding double centered distance matrices of the samples: 
		$\mathcal{V}^2_n(Y,Z) = \frac{1}{n^2} \sum_{k,l=1}^{n}A_{kl}B_{kl}$,
	where 
		$A_{kl}  = a_{kl} - \bar{a}_{k\cdot} - \bar{a}_{\cdot l} + \bar{a}_{\cdot \cdot},\ B_{kl} = b_{kl} - \bar{b}_{k\cdot} - \bar{b}_{\cdot l} + \bar{b}_{\cdot \cdot}$, 
	with $a_{kl}  = \Vert Y_k-Y_l \Vert_2$, $\bar{a}_{k\cdot}  = \frac{1}{n}\sum_{l=1}^{n}a_{kl}$, $\bar{a}_{\cdot l}  = \frac{1}{n}\sum_{k=1}^{n}a_{kl}$, $\bar{a}_{\cdot \cdot}  = \frac{1}{n^2}\sum_{l,k=1}^{n}a_{kl}$ and $b_{kl} = \Vert Z_k-Z_l \Vert_2$, $\bar{b}_{k\cdot}  = \frac{1}{n}\sum_{l=1}^{n}b_{kl}$, $\bar{b}_{\cdot l}  = \frac{1}{n}\sum_{k=1}^{n}b_{kl}$, $\bar{b}_{\cdot \cdot}  = \frac{1}{n^2}\sum_{l,k=1}^{n}b_{kl}$.
\end{definition}

Similar to DC, directly computing CDC can be challenging. Instead, we can use the sample CDC approach. For computational considerations, we adopt the choice $a(\U) = 12f^4(\U)$, following the approach in \cite{wang2015conditional}. Here, $f(\cdot)$ represents the density function of $\U$. The empirical conditional distance covariance \citep{wang2015conditional} is defined as follows:

\begin{definition}[Definition of empirical CDC in \cite{wang2015conditional}]\label{def:pcdcov}
Consider the observed samples ${(Y_i, Z_i,U_i )}_{i=1}^n$, which are drawn from the joint distribution of random vectors $( \mathbf{Y}, \mathbf{Z},\mathbf{U}) \in  \mathbb{R}^p \times \mathbb{R}^q\times\mathbb{R}^r $. We represent the three sample matrices as $Y=[Y_1,\cdots,Y_n]$, and $Z=[Z_1,\cdots,Z_n]$, $U=[U_1,\cdots,U_n]$. Let $K$ be a Gaussian kernel function, and define the kernel estimate of $\U$ as $K_{ik} = K_h(U_i- U_k)$, where $h$ is the bandwidth of the kernel density estimation (KDE). Then the empirical statistics of $\Ss_a(\Y,\Z,\U)$ is 
$$\Ss_n(Y,Z,U) =\frac{12}{n}\sum_{u=1}^{n}\frac{\left(\sum^n_{i=1}K_{iu}\right)^4}{n^4}\mathcal{D}^2_n( Y,Z \vert U_u),$$
 where $\omega_{ku}=K_{ku}/\sum_k K_{ku}$, $d_{kl}^Y = \Vert Y_k-Y_l \Vert_2$ and $d_{kl}^Z = \Vert Z_k-Z_l \Vert_2$, $D_1=\sum^n_{k,\ell}d^Y_{k\ell}d^Z_{k\ell}\omega_{ku}\omega_{\ell u}$, $D_2=\sum^n_{k,\ell}d^Y_{k\ell}\omega_{ku}\omega_{\ell u}\sum^n_{k,\ell}d^Z_{k\ell}\omega_{ku}\omega_{\ell u}$, $D_3=\sum^n_{k,\ell,m}d^Y_{k\ell}d^Z_{km}\omega_{ku}\omega_{\ell u}\omega_{mu}$ 
	and the sample conditional distance covariance $\mathcal{D}^2_n(Y, Z \vert U)$ is 
$\mathcal{D}^2_n(Y,Z \vert U_u) = D_1+D_2-2D_3$ by Equation (1) in supplementary material of \cite{wang2015conditional}.
\end{definition}
From Lemma 1 in \cite{wang2015conditional}, we have $\mathcal{D}^2_n( Y,Z \vert U_u)$ is nonnegative, then $\Ss_n(Y,Z,U)$ is also nonnegative.

\subsection{Convergence in probability about sample size}
\label{sec:convergence}
	From laws of large number, it ensures that as the sample size increases, the empirical estimate consistently approaches the true population value. It is worth noting that the convergence properties are contingent on having a sufficiently large sample size. This implies that a substantial number of samples is required to guarantee the convergence of the estimators. 
	
	In deep learning, working with a large number of samples is indeed crucial for effective model training. However, practical limitations, such as graphics card memory constraints, often restrict the amount of data that can be loaded at once. To overcome this challenge, mini-batch gradient descent is commonly employed, where only a subset of the training dataset (referred to as a mini-batch) is used to update the neural network parameters at each iteration.
	
	When using empirical (conditional) distance covariance as a substitute for (conditional) distance covariance in such scenarios, there is no direct guidance on the above convergence. The reason is that the convergence properties of empirical estimators heavily rely on the sample size. In mini-batch training, the mini-batch size ($n_b$) represents the effective sample size used for parameter updates. As $n_b$ is small compared to the total dataset, the convergence behavior may differ from the traditional almost sure convergence or convergence in probability. 
	In the following, we will estimate the probability that the empirical (conditional) distance covariance and the population (conditional) distance covariance are sufficiently close in terms of sample size. Thus, it will be easy to control sample size while considering the error rate given.
	
	The following theorems establish the results of convergence in probability about the sample size, which indicate approximately what batch size is necessary to achieve the desired level of accuracy.
\begin{theorem}[Convergence of DC in probability]\label{theo:converge}
Let  $\mathbf Y\in\mathbb{R}^p$ and $\mathbf Z\in\mathbb{R}^q$ be two sub-Gaussian random vectors and $Y=[Y_1,\cdots,Y_n]$, $Z=[Z_1,\cdots,Z_n]$ be the sample matrices. For any $\epsilon>0$, we have
\begin{align*}
	 P(|\mathcal{V}^2_{n}(Y,Z)-\mathcal{V}^2(\mathbf Y,\mathbf Z)|>\epsilon) \le 
	 \frac {3}n+6e n\cdot\exp\left(-\frac{\epsilon}{576K^2_1}\sqrt{\frac n{\log n}}\right).
\end{align*}
\end{theorem}
\begin{proof}
    We defer the proof to Appendix \ref{p:converge}.
\end{proof}

\begin{theorem}[Convergence of CDC in probability]\label{p:cdc_convergence}
	Let $\mathbf Y\in\mathbb{R}^p$, $\mathbf Z\in\mathbb{R}^q$ be two sub-Gaussian random vectors with the second moments of $\Y$ and $\Z$ exist, $\mathbf U\in\mathbb{R}^r$ satisfy the regularity conditions
	\begin{itemize}
\item[1.] $\int_{\R^r}uK(u)du=0$, $\int_{\R^r}K(u)du=0$, $\int_{\R^r}|K(u)|du<\infty$, $\int_{\R^r}uK^2(u)du>0$, $\int_{\R^r}u^2K(u)du<\infty$,
\item[2.] $h^r\to 0$ and $nh^r\to \infty$, as $n\to\infty$. This requires $h$ to be chosen appropriately according to $n$,
\item[3.] The density function of $\U$ and the conditional density function $f_{\Y,\Z|\U}$ are twice differentiable and all of the derivatives are bounded (denoted as $C<+\infty$),
\end{itemize}
  and  $Y=[Y_1,\cdots,Y_n]$ ,$Z=[Z_1,\cdots,Z_n]$, $U=[U_1,\cdots,U_n]$ be the sample matrices. Let $\Ss_n(Y,Z,U) = \frac{12}{n}\sum_{u=1}^{n}\left( \frac{\omega(U_u)}{n} \right)^4\mathcal{D}^2_n(Y, Z \vert U_u)$ and $\Ss_a(\Y,\Z,\U) = \E[\mathcal{D}^2(\Y, \Z \vert \U)a(\U)]$. For any $\epsilon>0$, we have
	\begin{equation*}
|\mathcal{S}_{n}(Y, Z,U)-\mathcal{S}_{a}(\mathbf Y, \mathbf Z,\mathbf U)|\le\epsilon +Ch^2,
	\end{equation*}
with probability at least $1- \frac {8}{n^{120}}- 8en\cdot\exp\left(-\frac{\epsilon}{3840K_1}\sqrt{\frac n{\log n}}\right)$, where $h$ is the bandwidth of the Gaussian kernel.
\end{theorem}
\begin{proof}
    We defer the proof to Appendix \ref{appendix:cdc_converge}.
\end{proof}

By setting $h$ in a way that satisfies the second condition, the term $Ch^2$ tends to zero when $n$ goes to infinity. Consequently, we can select an appropriate sample size to ensure that $\mathcal{S}_n$ is sufficiently close to $\mathcal{S}_a$. In this paper, we employ Silverman's rule \citep{silverman1986density} to determine the bandwidth for the KDE. Specifically, we set $h = (n \frac{r+2}{4})^{\frac{-1}{r+4}}$, where $r$ denotes the dimension of $\mathbf{U}$.

Although our regular condition is based on continuous random vectors, it is easy to extend to discrete random vectors. That is,
\begin{itemize}
\item[1.] Let $K(u)$ be a probability of $\U$,  $\E(\U)=0$, $\sum_u|K(u)|<\infty$, $\sum_uuK^2(u)>0$, $\E(\U^2)<\infty$.
\item[2.] $h^r\to 0$ and $nh^r\to \infty$, as $n\to\infty$. This requires $h$ to be chosen appropriately according to $n$.
\item[3.] The second-order differences of the probability of $\U$ and the conditional probability $f_{\Y,\Z|\U}$ are bounded (denoted as $C<+\infty$).
\end{itemize}

Furthermore, in our fair classification model, we regard $\mathbf{Y}$ and $\mathbf{Z}$ as the random vectors associated with the predicted label (or feature map) and the sensitive attribute, respectively. They are all bounded, otherwise the neural network will be unstable or not convergent.  Note that bounded random vectors are all sub-Gaussian random vectors (see Remark \ref{remark:sub-gaussian}), so the assumptions are reasonable. 
\subsection{Matrix forms}
\label{sec:matrix_form}
To enhance computational efficiency, we employ the matrix form of empirical (conditional) distance covariance for parallel calculations. The matrix forms are outlined in the propositions provided below.
\begin{proposition}\label{pro:matrix}

	Let $\{(Y_i, Z_i), i=1,\dots,n\}$ be the observed samples. Define $\tilde A\in\mathbb{R}^{n\times n}$ with $\tilde A_{kl}=\Vert Y_k-Y_l \Vert_2$,  and $\tilde B\in\mathbb{R}^{n\times n}$ with $\tilde B_{kl}=\Vert Z_k-Z_l \Vert_2$. Then the matrix form of empirical distance covariance is
$$\frac{1}{n^2}\langle A,B\rangle, \hbox{ with } A = \tilde{A} - \frac{1}{n}\mathbf{1}\tilde{A} - \frac{1}{n}\tilde{A}\mathbf{1} + \frac{1}{n^2}\mathbf{1}\tilde{A}\mathbf{1},\ B = \tilde{B} - \frac{1}{n}\mathbf{1}\tilde{B} - \frac{1}{n}\tilde{B}\mathbf{1} + \frac{1}{n^2}\mathbf{1}\tilde{B}\mathbf{1},$$
where $\langle\cdot,\cdot\rangle$ is the inner product of two matrices, $\mathbf{1}\in\mathbb{R}^{n\times n}$ is a matrix with all entries equal to 1.
\end{proposition}
\begin{proof}
Let $\tilde{A} := (a_{kl})_{n\times n}$, and $\bar{a}_{k\cdot}  = \frac{1}{n}\sum_{l=1}^{n}a_{kl}$, $\bar{a}_{\cdot l}  = \frac{1}{n}\sum_{k=1}^{n}a_{kl}$, $\bar{a}_{\cdot \cdot}  = \frac{1}{n^2}\sum_{l,k=1}^{n}a_{kl}$. Then 
\begin{align*}
			\begin{bmatrix}
					\bar{a}_{1\cdot} & \cdots & \bar{a}_{n\cdot} \\
					\vdots&\ddots & \vdots  \\
					\bar{a}_{1\cdot} & \cdots & \bar{a}_{n\cdot}
				\end{bmatrix}  =  \frac{1}{n}\mathbf{1}\tilde{A},\ 
			\begin{bmatrix}
					\bar{a}_{\cdot1} &  \cdots &  \bar{a}_{\cdot 1} \\ 
					\vdots &  \ddots &  \vdots \\
					\bar{a}_{\cdot n} &  \cdots &  \bar{a}_{\cdot n}
				\end{bmatrix}  =  \frac{1}{n}\tilde{A}\mathbf{1},\ 
			\begin{bmatrix}
					\bar{a}_{\cdot \cdot} & \cdots  & \bar{a}_{\cdot \cdot} \\
					\vdots&\ddots& \vdots  \\
					\bar{a}_{\cdot \cdot} & \cdots  & \bar{a}_{\cdot \cdot}
				\end{bmatrix}  = \frac{1}{n^2}\mathbf{1}\tilde{A}\mathbf{1}.
		\end{align*}
Hence,
		$A = (A_{kl})_{n \times n}  = (a_{kl} - \bar{a}_{k\cdot} - \bar{a}_{\cdot l} + \bar{a}_{\cdot \cdot})_{n \times n} \notag  = \tilde{A} - \frac{1}{n}\mathbf{1}\tilde{A} - \frac{1}{n}\tilde{A}\mathbf{1} + \frac{1}{n^2}\mathbf{1}\tilde{A}\mathbf{1}$. Similarly, $B = \tilde{B} - \frac{1}{n}\mathbf{1}\tilde{B} - \frac{1}{n}\tilde{B}\mathbf{1} + \frac{1}{n^2}\mathbf{1}\tilde{B}\mathbf{1}$. We conclude the result.
\end{proof}

\begin{proposition}\label{pro:matrix1}
	Let $\{(Y_i, Z_i,U_i), i=1,\dots,n\}$ be the observed samples. Define $D^Y=(d^Y_{k\ell})\in\mathbb{R}^{n\times n}$ with $d^Y_{k\ell}=\Vert Y_k-Y_\ell \Vert_2$,  $D^Z=(d^Z_{k\ell})\in\mathbb{R}^{n\times n}$ with $d^Z_{k\ell}=\Vert Z_k-Z_\ell \Vert_2$. Let $K\in\R^{n\times n}$ with $K_{k\ell}=\frac{1}{\sqrt{2\pi}\sigma}e^{-\frac{\|U_k-U_\ell\|^2}{2\sigma^2}}$.  
	Then the matrix form of empirical distance covariance is 
$$\frac{12}{n^5}(E_1+E_2+E_3),$$
where $E_1=\langle (K\mathbbm{1})\odot(K\mathbbm{1}), \diag(K^{\top}(D^Y\odot D^Z)K)\rangle$,  $E_2=\langle\diag(K^{\top}D^YK), \diag(K^{\top}D^ZK)\rangle$, $E_3=\langle K\mathbbm{1},\diag(K^{\top}((D^YK)\odot (D^ZK)))\rangle$, $\langle\cdot,\cdot\rangle$ is the inner product of two vectors, $\mathbbm{1}\in\mathbb{R}^{n}$ is a matrix with all entries equal to 1.
\end{proposition}
\begin{proof}
    We defer the proof to Appendix \ref{sec:proof_matrix_form}. 
\end{proof}  
In the following, we present the pseudocode for DC and CDC.

\begin{algorithm}[ht]
	\begin{algorithmic}[1]
		\Require{$\{(Y_i, Z_i), i=1,\dots,n\}$ be the observed samples.}
   \State Compute $\tilde A\in\mathbb{R}^{n\times n}$ with $\tilde A_{kl}=\Vert Y_k-Y_l \Vert_2$ and $\tilde B\in\mathbb{R}^{n\times n}$ with $\tilde B_{kl}=\Vert Z_k-Z_l \Vert_2$,
   \State Compute $A=\tilde A-\frac{1}{n}\sum^n_{i=1}\tilde{A}_{i,:}-\frac{1}{n}\sum^n_{j=1}\tilde{A}_{:,j}+\frac{1}{n^2}\sum^n_{i,j=1}\tilde{A}_{i,j}$,
   \State Compute $B=\tilde B-\frac{1}{n}\sum^n_{i=1}\tilde{B}_{i,:}-\frac{1}{n}\sum^n_{j=1}\tilde{B}_{:,j}+\frac{1}{n^2}\sum^n_{i,j=1}\tilde{B}_{i,j}$,\\
\Return $\frac{1}{n^2}\langle A,B\rangle$.
\end{algorithmic}
\caption{Computation of distance covariance.}
   \label{algo:dc}
\end{algorithm}
It is worth to noting that $\sum^n_{i=1}\tilde{A}_{i,:}$ is a row vector, $\sum^n_{j=1}\tilde{A}_{:,j}$ is a column vector and $\sum^n_{i,j=1}\tilde{A}_{i,j}$ is a scalar. Using broadcasting in Python, the operation $A=\tilde A-\frac{1}{n}\sum^n_{i=1}\tilde{A}_{i,:}-\frac{1}{n}\sum^n_{j=1}\tilde{A}_{:,j}+\frac{1}{n^2}\sum^n_{i,j=1}\tilde{A}_{i,j}$ is equivalent to $\tilde{A} - \frac{1}{n}\mathbf{1}\tilde{A} - \frac{1}{n}\tilde{A}\mathbf{1} + \frac{1}{n^2}\mathbf{1}\tilde{A}\mathbf{1}$. A similar construction applies to the matrix $B$.
\begin{algorithm}[ht]
	\begin{algorithmic}[1]
		\Require{${(Y_i, Z_i,U_i), i=1,\dots,n}$ be the observed samples.}
   \State  Compute $D^Y=(d^Y_{k\ell})\in\mathbb{R}^{n\times n}$ with $d^Y_{k\ell}=\Vert Y_k-Y_\ell \Vert_2$,  $D^Z=(d^Z_{k\ell})\in\mathbb{R}^{n\times n}$ with $d^Z_{k\ell}=\Vert Z_k-Z_\ell \Vert_2$,
   \State Compute $K\in\R^{n\times n}$ with $K_{k\ell}=\frac{1}{\sqrt{2\pi}\sigma}e^{-\frac{\|U_k-U_\ell\|^2}{2\sigma^2}}$,
   \State $E_1=\langle (K\mathbbm{1})\odot(K\mathbbm{1}), \diag(K^{\top}(D^Y\odot D^Z)K)\rangle$,  \State $E_2=\langle\diag(K^{\top}D^YK), \diag(K^{\top}D^ZK)\rangle$, 
   \State $E_3=\langle K\mathbbm{1},\diag(K^{\top}((D^YK)\odot (D^ZK)))\rangle$,\\
\Return $\frac{12}{n^5}(E_1+E_2+E_3).$
\end{algorithmic}
\caption{Computation of conditional distance covariance.}
   \label{algo:cdc}
\end{algorithm}

\section{The Fair Classification Problem}
\label{sec:model}
In the following we consider a standard fair supervised learning scenario. Let $\X \in \mathcal{X}$ be the predictive variables, $ \Y \in \mathcal{Y}\subset\mathbb{R}^p$ be target variables or labels, $ \Z \in \mathcal{Z} =\{Z_1,\cdots,Z_S\}\subset\mathbb{R}^q$ be a sensitive attribute. The datasets is a ternary pair of these variables $D = \{ (x_i, y_i, z_i) , i=1,2,\dots,n\} $. Our aim is to find a fair model $\phi : \X \rightarrow \hat{\Y} $ with respect to the associated sensitive attribute. That is, $ \hat{y}_i = \phi(x_i),\ \forall i\in[n]$.

While the fight against bias and discrimination has a long history in philosophy, psychology, and more recently in machine learning, there is still no universally accepted criterion for defining fairness due to cultural differences and varying preferences. In this paper, we will focus on two common fairness criteria: demographic parity (DP) \citep{dworkFairnessAwareness2012} and Equalized Odds (EO) \citep{hardtEqualityOpportunitySupervised2016}. 
The DP criterion requires that predictions $\hat{\Y}$ and sensitive attributes $\Z$ are independent, represented as $\hat{\Y} \perp \Z$. 
On the other hand, the EO criterion states that predictions $\hat{\Y}$ and sensitive attributes $\Z$ should be conditionally independent given the labels $\Y$, denoted as $\hat{\Y} \perp \Z \vert \Y$. 
Noting that these fair criterion align respectively with distance covariance and conditional distance covariance.

But both the computation of population DC and CDC requires knowledge of the analytical form of the distribution function, which are usually unknown in practice. In previous studies, \cite{liuFairRepresentationLearning2022} assumed the random variables followed the normal distribution and used VAE for approximation to subsequently calculate the population DC. In contrast, we compute empirical DC and CDC statistics as defined in \cref{def:dcov} and \cref{def:pcdcov}. 
Empirical DC and CDC, as statistics, can be directly computed from the sample without requiring additional prior knowledge or assumptions.

Therefore, when the fairness objective is DP, we employ empirical DC $\mathcal{V}_n^2(\hat{Y}, Z)$ as a constraint term, and when the fairness objective is EO, we use empirical statistics $\Ss_n(\hat{Y}, Z, Y)$ as a constraint term. To maintain uniform notation for the respective constraint term used for fairness objectives, we consistently denote it as $\mathcal{D}(\hat{Y}, Z, Y)$. Then, the optimization model is: 
\begin{equation}
	\label{eq:model_label}
	\min\ \mathcal{L}_{CE}(\hat Y,Y)\quad
	\hbox{s.t. }\mathcal{D}(\hat{Y}, Z, Y)=0,
\end{equation}
and the corresponding Lagrangian function of (\ref{eq:model_label}) is $\mathcal{L}(\phi_\theta, \lambda)  = \mathcal{L}_{CE}(\hat{Y}, Y) + \lambda \mathcal{D}(\hat{Y}, Z, Y) $, where $\lambda>0$ is the Lagrangian multiplier serving as a hyperparameter for balancing the fitting term and the regularized term. Hyperparameters play a crucial role in optimization, influencing not only the trade-off between fairness and accuracy but also the final convergence of the model during training (whether it can achieve higher accuracy at the same level of fairness, and vice versa).

Building on this inspiration, we consider $\lambda$ as a dual variable in this paper. 
Let $
g(\lambda) = {\inf}_{\theta}\ \mathcal{L}(\phi_\theta, \lambda)$, the corresponding Lagrangian dual is
\begin{equation}
	\label{eq:model_dual}
	\underset{\lambda}{\max}\ g(\lambda) = \underset{\lambda}{\max}\ \underset{\theta}{\inf}\ \mathcal{L}(\phi_\theta, \lambda).
\end{equation}

In fact, the model (\ref{eq:model_dual}) provides an alternative if it is not easy to choose a specific $\lambda$ when we minimize $\mathcal L$ w.r.t. $\theta$.
But model (\ref{eq:model_dual}) becomes a max-min problem, which may be hard to solve directly. We update iteratively and the scheme is as follows: we give the value $\lambda_e$ and $\theta_e$, and find the optimal value of $\theta_{e+1} $ and $\lambda_{e+1}$ by
\begin{align*}\label{eq:min_theta}
	& \theta_{e+1} = \underset{\theta}{\arg\min}\ \mathcal{L}(\phi_\theta, \lambda_e),\  \lambda_{e+1} = \lambda_e + \beta\cdot\frac{\partial \mathcal{L}(\phi_{\theta_e}, \lambda)}{\partial\lambda}\bigg|_{\lambda=\lambda_e}
	 =\lambda_e+\beta\cdot\mathcal{D}(\hat{Y}, Z, Y),
\end{align*}
where $\beta$ is the learning rate. In the optimization algorithm, we utilize a batch-wise approach by dividing the entire dataset into smaller batches. For each batch, we update the network parameters, which enables incremental improvements in the model's performance. Subsequently, after updating the parameters for all batches, we update the dual variable based on the average of the results obtained from each batch. 

\section{Numerical Experiments}
\label{sec:numerical}
In this section, we present the results of numerical experiments using our proposed method on four real-world datasets, including UCI Adult dataset, ACSIncome dataset and two image datasets. 
We compare our proposed method with the following baselines:
\textbf{Unfair} (without fairness), 
\textbf{HGR} \citep{maryFairnessAwareLearningContinuous2019a}, 
\textbf{Soft-HGR} \citep{leeMaximalCorrelationFramework2022},
\textbf{FairMixup} \citep{chuangFairMixupFairness2020},
\textbf{FERMI} \citep{lowy2022stochastic},
\textbf{FairDisCo} \citep{liuFairRepresentationLearning2022},
\textbf{Dist-Fair} \citep{guoLearningFairRepresentations2022}.
The criteria used to assess the performance of fairness are $\Delta$DP and $\Delta$EO \citep{parkFairContrastiveLearning2022}:
\begin{align*}
	\Delta \text{DP} =& \frac{1}{\binom{S}{2}} \sum_{i<j}|P(\hat{\Y} | \Z=Z_i) - P(\hat{\Y} | \Z=Z_j)|, \\
	\Delta \text{EO} =& \frac{1}{2\binom{S}{2}} \sum_{y\in\{0,1\}}\sum_{i<j}|P(\hat{\Y}=y | \Z=Z_i, \Y=y) 
	 - P(\hat{\Y}=y | \Z=Z_j, \Y=y)|,
\end{align*}
where $S$ represents the number of categories or groups within the sensitive attribute $\Z$. In the subsequent subsections, we present the results of our numerical experiments. For additional details and supplementary information, please refer to Appendix \ref{appendix:exp}. In addition, we put the ablation study in Appendix \ref{appendix:ablation} and the complexity analysis in Appendix \ref{appendix:complexity}. 

\subsection{Tabular Dataset}

In our study on tabular datasets, we utilized the widely recognized UCI Adult dataset \citep{duaUCIMachineLearning2017} and ACSIncome dataset \citep{ding2021retiring}. We followed the pre-processing procedures outlined by \cite{han2023ffb}.

\begin{figure}[ht]
	\vskip 0.2in
	\begin{minipage}{0.495\linewidth}
		\centering
		\includegraphics[width=1\linewidth]{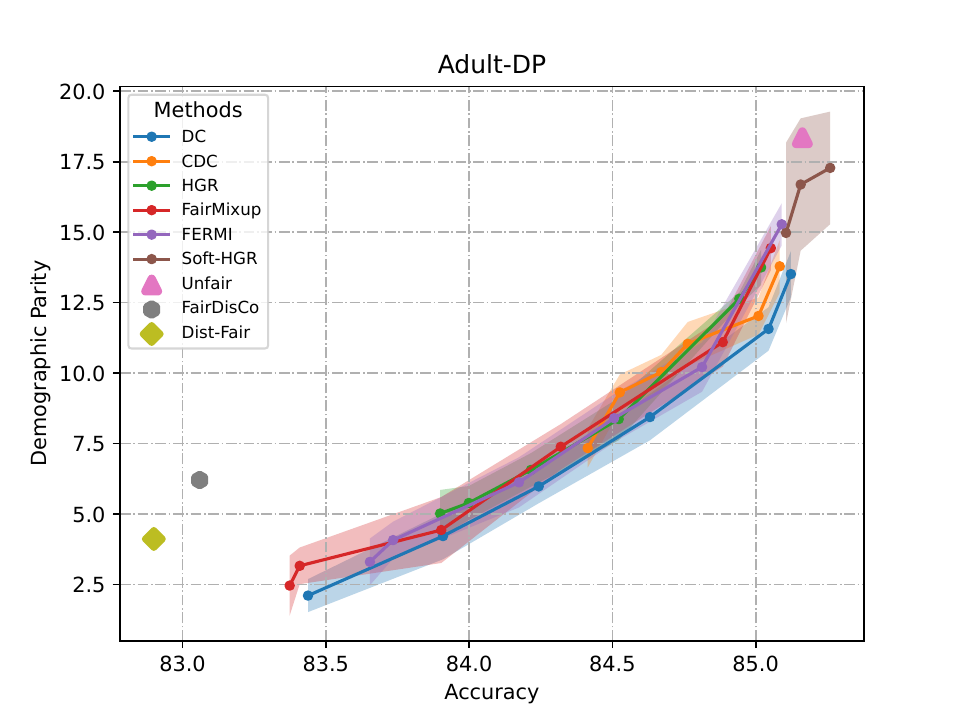}
		\label{fig:adult_dp}
	\end{minipage}
		\begin{minipage}{0.495\linewidth}
		\centering
		\includegraphics[width=1\linewidth]{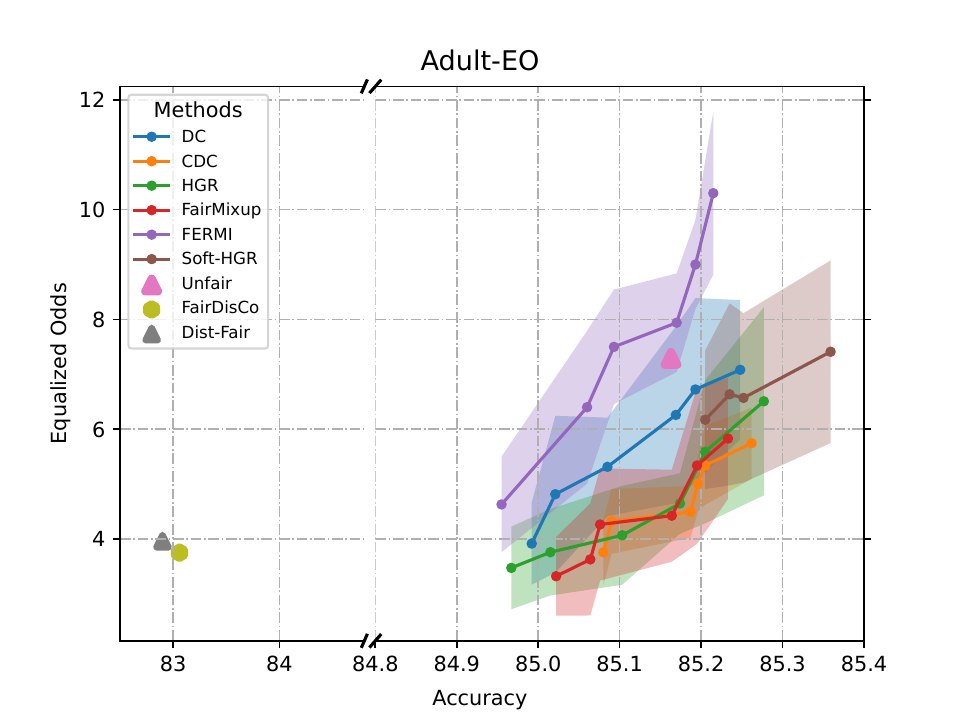}
		\label{fig:adult_eo}
	\end{minipage}
	\begin{minipage}{0.495\linewidth}
		\centering
		\includegraphics[width=1\linewidth]{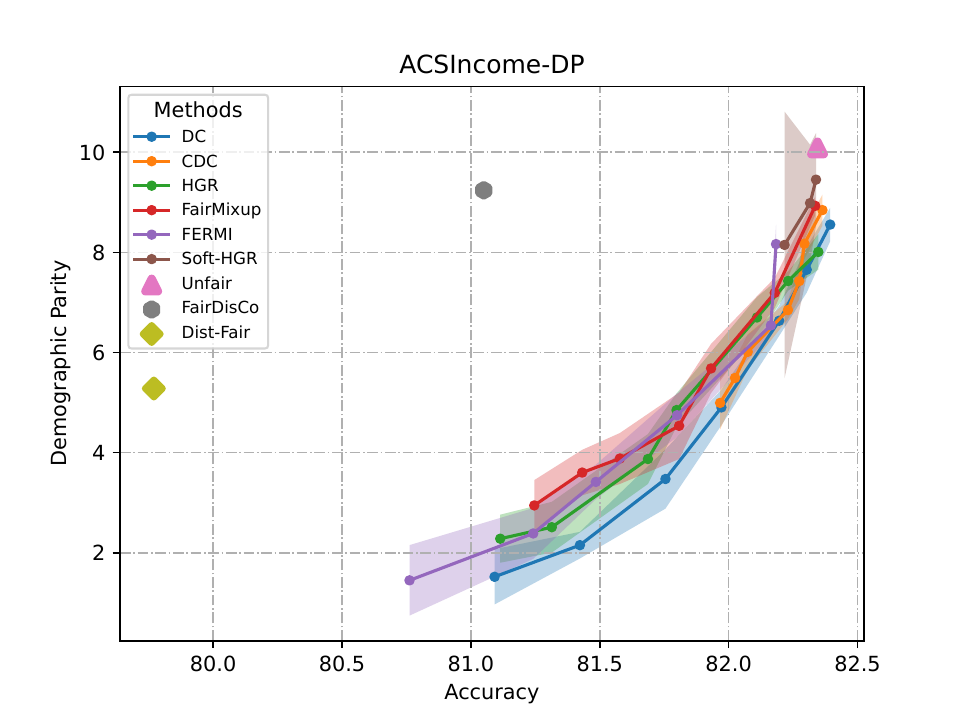}
		\label{fig:acs_dp}
	\end{minipage}
	\begin{minipage}{0.495\linewidth}
		\centering
		\includegraphics[width=1\linewidth]{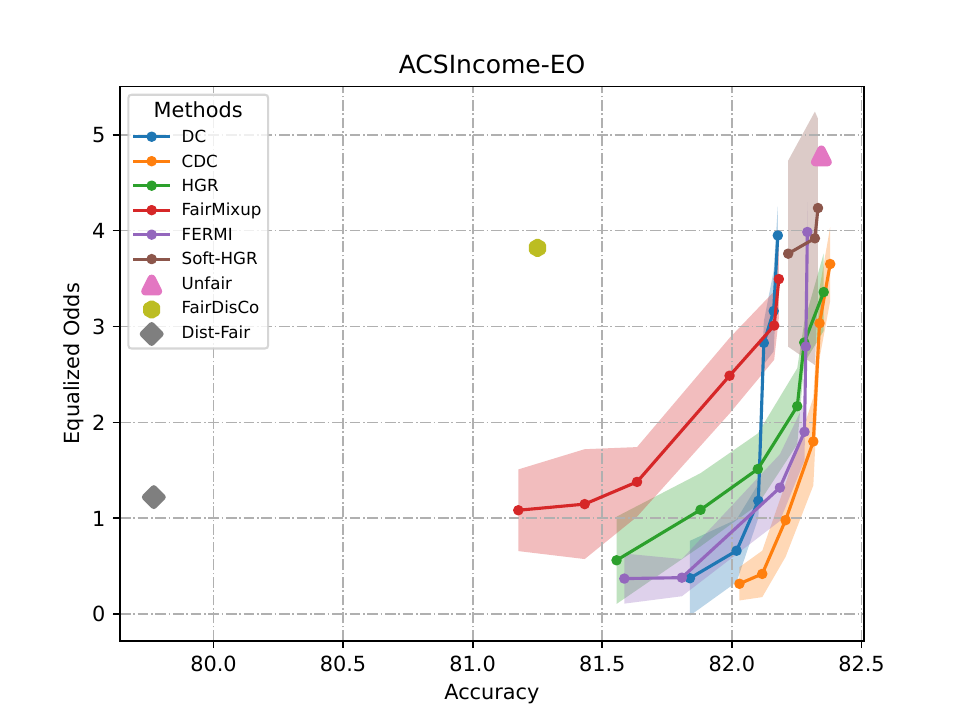}
		\label{fig:acs_eo}
	\end{minipage}
    \vskip -0.2in
	\caption{The results between accuracy and fairness for the UCI and ACSIncome datasets.}
	\label{fig:tabular}
	
\end{figure}

For experiments, we use a four-layer Multilayer Perceptron (MLP) model with ReLU activation function and Cross Entropy Loss, referred to as the \textbf{Unfair} model. All baselines use the same classifier network structure to ensure a fair comparison.    
The four subfigures in Figure \ref{fig:tabular} depict the trade-off curves between accuracy and either $\Delta$DP or $\Delta$EO for the UCI and ACSIncome datasets with different initial guesses of $\lambda$. In our experiments, the initial guesses we choose are in the interval $[0.5,20]$. The displayed results represent the averages and standard deviations of both accuracy and fairness criteria over 10 runs. In the second subfigure, non-uniform tick marks have been applied on the horizontal axis at both ends of the double slash. This adjustment aims to improve the visualization of the trade-off curve for better clarity and understanding.

In the figure, the position of the curve closer to the right indicates higher accuracy, while the curve positioned closer to the bottom signifies fairer outcomes according to the specified fairness metric. DC and CDC are considered among the most effective methods for achieving the trade-off between accuracy and  either DP or EO, respectively. The results highlight the competitive performance of our methods on the Tabular dataset. This suggests that utilizing (conditional) distance covariance as a regularizer yields superior performance in capturing (conditional) independence, as our model achieves higher accuracy while maintaining the same level of $\Delta$DP or $\Delta$EO.
\begin{table*}[ht]
	\centering
	\caption{Classification Results on CelebA.}
	\label{tab:celeba}
	\resizebox{1\columnwidth}{!}{
	\renewcommand\arraystretch{1.1}
	\begin{tabular}{lccccccccccccccc} 
		\hline
		\multirow{2}{*}{Methods} & \multicolumn{3}{c}{Attractive/gender}                                                      &                      & \multicolumn{3}{c}{Wavy hair/gender}                                                                           &                      & \multicolumn{3}{c}{Attractive/young}                                                                   &                      & \multicolumn{3}{c}{Attractive/gender\ young}                                         \\ 
		\cmidrule(r){2-4}\cmidrule(r){6-8}\cmidrule(r){10-12}\cmidrule(r){14-16}
		& Acc                        & $\Delta$EO                         & $\Delta$DP               &                      & Acc                                 & $\Delta$EO                         & $\Delta$DP                          &                      & Acc                                 & $\Delta$EO                          & $\Delta$DP                 &                      & Acc                        & $\Delta$EO                & $\Delta$DP                \\ 
		\hline
		Unfair                   & $\mathbf{82.188_{\pm0.4}}$ & 20.514$_{\pm0.7}$                    & 39.722$_{\pm3}$            &                      & $\mathbf{83.147_{\pm0.4}}$          & 19.229$_{\pm1.6}$                    & 30.96$_{\pm1.7}$                      &                      & $ \mathbf{82.129_{\pm0.2}}$         & 22.912$_{\pm0.4}$                     & 46.072$_{\pm0.5}$            &                      & $\mathbf{82.149_{\pm0.2}}$ & 24.418$_{\pm0.6}$           & 34.649$_{\pm0.3}$           \\
		HGR                      & 80.423$_{\pm0.2}$            & 1.175$_{\pm0.2}$                     & $\mathbf{24.720_{\pm1.5}}$          &                      & 80.139$_{\pm0.7}$                     & $\mathbf{2.052_{\pm0.4}}$                     & $\mathbf{17.156_{\pm1.2}}$ &                      & 78.796$_{\pm0.7}$                     & 8.74$_{\pm0.8}$                       & 31.91$_{\pm0.9}$             &                      & 78.320$_{\pm0.2}$            & 9.571$_{\pm0.4}$            & 26.149$_{\pm0.2}$           \\
		Fair Mixup               & 79.146$_{\pm0.2}$            & 1.303$_{\pm0.4}$ & $\mathbf{25.243_{\pm1}}$            &  & 81.118$_{\pm0.1}$ & $\mathbf{1.643_{\pm0.5}}$ & 24.063$_{\pm3.2}$ &  & 79.321$_{\pm0.7}$ & 11.715$_{\pm1.5}$ & 47.313$_{\pm1.2}$            & 	 & --                          & --                         & --                        \\
		FERMI                    &     80.580$_{\pm0.2}$          &      $\mathbf{0.479_{\pm0.1}}$               &      $\mathbf{24.809_{\pm0.2}}$   &                      &  80.725$_{\pm0.8}$           &    $\mathbf{1.826_{\pm0.3}}$            &    $\mathbf{16.722_{\pm1.0}}$                 &                      & 78.678$_{\pm0.4}$         &       7.583$_{\pm0.9}$       &    $\mathbf{28.335_{\pm2.3}}$      &                      &  76.904$_{\pm0.3}$       & 9.430$_{\pm0.4}$         & $\mathbf{22.631_{\pm0.6}}$            \\
		Dist-Fair                & 70.738$_{\pm1.2}$            & 13.108$_{\pm3}$                      & $\mathbf{26.169_{\pm3.8}}$          &                      & 67.471$_\pm1$                       & 13.764$_{\pm5.3}$                    & $\mathbf{14.655_{\pm5.5}}$ &                      & 71.553$_{\pm0.4}$                     & 8.926$_{\pm0.7}$                      & $\mathbf{26.215_{\pm0.7}}$ &                      & 71.072$_{\pm1.1}$           & 18.92$_{\pm3.3}$            & $\mathbf{24.111_{\pm2.3}}$           \\
        Soft-HGR &$81.295_{\pm0.1}$&$5.48_{\pm0.3}$&$29.975_{\pm0.2}$&&$82.230_{\pm0.6}$&$18.747_{\pm2.2}$&$28.433_{\pm1.9}$&&$81.63_{\pm0.2}$&$45.377_{\pm1.5}$&$21.927_{\pm0.6}$&&$82.530_{\pm0.1}$&$30.323_{\pm1.9}$&$28.847_{\pm 2.0}$\\
		DC                       & 80.568$_{\pm0.3}$            & $\mathbf{0.563_{\pm0.3}}$        & $\mathbf{24.121_{\pm1}}$ &                      & 81.785$_{\pm0.2}$                     & $\mathbf{1.122_{\pm0.6}}$          & $\mathbf{17.665_{\pm1.2}}$ &                      & 78.845$_{\pm0.5}$                     & $7.511_{\pm0.9}$           & 34.371$_{\pm0.6}$            &                      & 77.532$_{\pm1.0}$            & $\mathbf{8.107_{\pm1.0}}$ & $\mathbf{23.314_{\pm1}}$  \\
		CDC                      & 80.346$_{\pm0.1}$           &  $ \mathbf{0.334_{\pm0.1}}$                &   $\mathbf{24.413_{\pm0.2}}$         &                      &    80.480$_{\pm0.3}$              &     $\mathbf{0.865_{\pm0.3}}$               &  $\mathbf{16.044_{\pm0.2}}$                &                      &    78.666$_{\pm0.3}$                &    $\mathbf{4.150_{\pm0.9}}$                   &    $28.352_{\pm1.0}$          &                      &    77.684$_{\pm0.1}$          &    $\mathbf{7.788_{\pm0.7}}$       &     $23.518_{\pm0.2}$           \\
		\hline
	\end{tabular}}
	% \vskip -0.2in
\end{table*}
\begin{table}
	\centering
	\caption{Classification Results on UTKFace.}
	\label{tab:utkface}
	\renewcommand\arraystretch{1.1}
	\begin{tabular}{lccccccc} 
		\toprule
		\multirow{2}{*}{Methods} & \multicolumn{3}{c}{Age/gender}                      &  & \multicolumn{3}{c}{Age/race}                             \\ 
		\cmidrule(r){2-4}\cmidrule(r){6-8}
		& Acc              & $\Delta$EO      & $\Delta$DP         &  & Acc               & $\Delta$EO      & $\Delta$DP         \\ 
		\hline
		Unfair                   &   80.80$_{\pm1.0}$     &  9.19$_{\pm1.7}$   &   18.93$_{\pm2.2}$     &  &    $\mathbf{82.16_{\pm0.4}}$ & 14.68$_{\pm0.7}$  &   25.69$_{\pm0.3}$    \\
		HGR                      & $\mathbf{81.96_{\pm0.7}}$ & 4.89$_{\pm1.6}$ & 16.73$_{\pm1.7}$  &  & 80.53$_{\pm0.3}$ & 9.58$_{\pm0.6}$ &  20.07$_{\pm0.6}$ \\
		Fair Mixup               & 79.44$_{\pm1.0}$ & 8.90$_{\pm0.7}$ & 17.70$_{\pm1.1}$  &  & 80.67$_{\pm1.6}$ & 13.91$_{\pm1.2}$&  24.02$_{\pm0.6}$ \\
		FERMI                    & $\mathbf{80.99_{\pm0.6}}$ & 2.56$_{\pm0.7}$ & 13.61$_{\pm0.6}$  &  & 80.10$_{\pm0.2}$ & $\mathbf{2.34_{\pm0.9}}$ &  9.61$_{\pm0.8}$ \\
		Dist-Fair                & 58.52$_{\pm0.8}$  & $1.39_{\pm0.5}$ &  $\mathbf{1.19_{\pm0.5}}$    &  & 59.10$_{\pm1.3}$  &$\mathbf{1.84_{\pm0.5}}$   &  $\mathbf{1.53_{\pm0.5}}$  \\
        Soft-HGR &80.11$_{\pm2.2}$&7.80$_{\pm2.2}$&17.20$_{\pm 3.1}$&&80.42$_{\pm2.2}$&13.80$_{\pm1.4}$&23.60$_{\pm2.2}$\\
		DC                       & $\mathbf{81.42_{\pm0.3}}$ & $\mathbf{0.82_{\pm0.2}}$ & 12.13$_{\pm0.7}$   &  & 78.07$_{\pm0.7}$ & $\mathbf{1.34_{\pm0.3}}$ & 9.94$_{\pm0.1}$    \\
		CDC                      & 80.75$_{\pm0.5}$ &$\mathbf{ 0.69_{\pm0.1}}$ & 11.456$_{\pm0.1}$ &  & 79.82$_{\pm1.5}$ & $\mathbf{1.18_{\pm0.2}}$ & 11.83$_{\pm0.8}$  \\
		\bottomrule
	\end{tabular}
\vskip -0.2in
\end{table}

\subsection{Image Dataset}

In our experiments on image datasets, we conducted evaluations on both the CelebA and UTKFace datasets. For each classification task, we utilize the ResNet-18 architecture \citep{heDeepResidualLearning2016a} as the encoder $\psi_\theta$ to process the input facial images into representations. These representations are then passed through a two-layer Multilayer Perceptron to make the final predictions. The \textbf{Unfair} model is trained with the Cross Entropy loss function without any regularization term. All baselines share the same architecture as ``Unfair''. The only exception is the Dist-fair model, which employs a VAE-ResNet18 architecture instead.

Due to the requirement of calculating the probability density function of the multivariate normal distribution in \textbf{FairDisCo} \citep{liuFairRepresentationLearning2022}, its utilization becomes impractical when dealing with high-dimensional cases. Therefore, we are unable to offer comparison results for the image datasets as they typically possess extremely large dimensions.

\subsubsection{CelebA}
The CelebA dataset \citep{liuDeepLearningFace2015} is a widely used image dataset for various computer vision tasks, particularly those related to faces. 
In our experiments, we selected one or two sensitive attributes from the CelebA dataset to evaluate the performance of our methods on mitigating potential ethical risks. For the classification tasks, we specifically focused on the attributes `attractiveness',   `wavy hair', `gender' and `young'. These attributes were chosen based on two main criteria. First, we considered attributes that have a high Pearson correlation coefficient with the sensitive attributes. Second, we considered attributes where the subgroups based on the sensitive attributes are imbalanced. 

In Table \ref{tab:celeba}, the first three groups present the classification accuracy and fairness (measured by $\Delta$DP or $\Delta$EO) for various combinations using one sensitive attribute. Due to the imbalanced nature of the CelebA dataset, it can be challenging to achieve both high accuracy and satisfy DP simultaneously. Our method demonstrates strong competitiveness in achieving both high accuracy and fairness across these attribute combinations. The results indicate that our approach performs favorably compared to other methods in terms of balancing accuracy and fairness in classification tasks.  
The last group in Table \ref{tab:celeba} demonstrates that our method maintains fairness in classification even in scenarios involving multiple sensitive attributes (Attractive/male \& young). This ability is facilitated by the application of (conditional) distance covariance, which allows for handling fair classification with arbitrary dimensional random vectors.
\subsubsection{UTKFace} \label{sec:utkface}

The UTKFace dataset \citep{zhangAgeProgressionRegression2017} is a dataset consisting of around 20,000 facial images. Each image is associated with attribute labels such as gender, age, and ethnicity. Following from \cite{han2023ffb}, we consider  `gender' and `race' as the sensitive attribute and  `age' as the target attribute. During our evaluation, we partitioned the UTKFace dataset into training (70\%), validation (15\%), and testing (15\%) sets. Table \ref{tab:utkface} displays the results of our evaluation. Specifically, when considering the Age/gender pair, HGR achieves slightly higher accuracy compared to our methods. However, it is worth noting that our methods demonstrate significantly better fairness criteria when compared to other existing methods. For the Age/race pair, our methods are also competitive when balancing the accuracy and fairness.

Additionally, we present a comparison of the training times between different methods in Table \ref{tab:time}. 
Due to the matrix computations, our methods also have advantages in terms of computational efficiency. 
It is important to note that while we have provided results for one group per dataset in this table, the same conclusion holds for other groups within each dataset in our experiments. 
\begin{table}
	\centering
	\caption{Comparison of Training Times.}
	\label{tab:time}
	\renewcommand\arraystretch{1.1}
	\begin{tabular}{lcccc} 
		\hline
		\multicolumn{1}{c}{\diagbox{Method}{Dataset}} & Adult  & ACSIncome & CelebA     & UTKFace  \\ 
		\hline
		Unfair                                        & 56s    & 45s       & 44m 38s    & 24m 18s  \\
		HGR                                           & 1m 6s  & 46s       & 47m 24s    & 25m 28s  \\
		Fair Mixup                                    & 58s    & 47s       & 1h 6m 31s  & 30m 59s  \\
		FERMI                                         & 6m 52s & 10m19s    & 6h 14m 49s & 45m 45s  \\
		FairDisCo                                     & 6m 22s  & 13m 32s   & --          & --        \\
		Dist-Fair                                     & 6m 28s & 13m 46s   & 2h 17m 7s  & 33m 27s   \\
        Soft-HGR &1m 6s&3m 16s&4h 30m 26s& 35m 44s\\
		DC                                            & 59s    & 47s       & 45m 38s    & 25m 36s  \\
		CDC                                           & 1m 6s  & 49s       & 54m 59s    & 26m 41s  \\
		\hline
	\end{tabular}
\vskip -0.2in
\end{table}

\section{Conclusion}
In this paper, we introduce (conditional) distance covariance as powerful tools to depict fairness in machine learning.   
We analyze the properties of (conditional) distance covariance and provide convergence analysis of the empirical (conditional) distance covariance in probability. To address the challenge of setting a balanced balance parameter, we treat it as a dual variable and update it along with model parameters.  
Finally, we demonstrate the effectiveness and wide applicability through numerical experiments.

Future work in the field of fair machine learning encompasses several directions. While our method primarily focuses on classification tasks, it is essential to explore its applicability to regression problems with discrete/continuous sensitive attribute(s) as well. Furthermore, it is crucial to delve into the integration of statistical and optimization techniques with fairness in machine learning. 

\bibliographystyle{alpha}
\bibliography{references_bak}

@article{alghamdi2022beyond,
  title={Beyond Adult and COMPAS: Fair multi-class prediction via information projection},
  author={Alghamdi, Wael and Hsu, Hsiang and Jeong, Haewon and Wang, Hao and Michalak, Peter and Asoodeh, Shahab and Calmon, Flavio},
  journal={Advances in Neural Information Processing Systems},
  volume={35},
  pages={38747--38760},
  year={2022}
}

@article{ding2021retiring,
  title={Retiring adult: New datasets for fair machine learning},
  author={Ding, Frances and Hardt, Moritz and Miller, John and Schmidt, Ludwig},
  journal={Advances in neural information processing systems},
  volume={34},
  pages={6478--6490},
  year={2021}
}

@article{bellamyAIFairness3602018,
  title = {{{AI Fairness}} 360: {{An Extensible Toolkit}} for {{Detecting}}, {{Understanding}}, and {{Mitigating Unwanted Algorithmic Bias}}},
  shorttitle = {{{AI Fairness}} 360},
  author = {Bellamy, Rachel K. E. and Dey, Kuntal and Hind, Michael and Hoffman, Samuel C. and Houde, Stephanie and Kannan, Kalapriya and Lohia, Pranay and Martino, Jacquelyn and Mehta, Sameep and Mojsilovic, Aleksandra and Nagar, Seema and Ramamurthy, Karthikeyan Natesan and Richards, John and Saha, Diptikalyan and Sattigeri, Prasanna and Singh, Moninder and Varshney, Kush R. and Zhang, Yunfeng},
  year = {2018},
  journal = {arXiv preprint arXiv:1810.01943},
  //number = {arXiv:1810.01943},
  eprint = {1810.01943},
  primaryclass = {cs},
  publisher = {{arXiv}},
  //doi = {10.48550/arXiv.1810.01943},
  urldate = {2023-02-23},
  archiveprefix = {arxiv},
  keywords = {Computer Science - Artificial Intelligence}
}

@inproceedings{bolukbasiManComputerProgrammer2016,
  title = {Man Is to Computer Programmer as Woman Is to Homemaker? Debiasing Word Embeddings},
  shorttitle = {Man Is to Computer Programmer as Woman Is to {Homemaker}?},
  booktitle = {Advances in Neural Information Processing Systems},
  author = {Bolukbasi, Tolga and Chang, Kai-Wei and Zou, James Y. and Saligrama, Venkatesh and Kalai, Adam T.},
  year = {2016},
  volume = {29}
}

@article{calmonOptimizedDataPreProcessing2017,
  title = {Optimized {{Data Pre-Processing}} for {{Discrimination Prevention}}},
  author={Calmon, Fl{\'a}vio P. and Wei, Dennis and Ramamurthy, Karthikeyan Natesan and Varshney, Kush R.},
  journal={CoRR, abs/1704.03354},
  year={2017}
}

@inproceedings{chuangFairMixupFairness2020,
  title = {Fair Mixup: Fairness via Interpolation},
  shorttitle = {Fair Mixup},
  booktitle = {International Conference on Learning Representations},
  author = {Chuang, Ching-Yao and Mroueh, Youssef},
  year = {2020},
  urldate = {2023-09-19},
  langid = {english}
}

@inproceedings{creagerFlexiblyFairRepresentation2019,
  title = {Flexibly {{Fair Representation Learning}} by {{Disentanglement}}},
  author = {Creager, Elliot and Madras, David and Jacobsen, J{\"o}rn-Henrik and Weis, Marissa A. and Swersky, Kevin and Pitassi, Toniann and Zemel, Richard},
  booktitle={International conference on machine learning},
  pages={1436--1445},
  year={2019},
  organization={PMLR}
}

@article{duaUCIMachineLearning2017,
  title = {{{UCI Machine Learning Repository}}, {{Adult Dataset}}},
  author = {Dua, Dheeru and Graff, Casey},
  year = {2017},
  journal = {URL https://archive. ics. uci. edu/ml/datasets/adult}
}

@inproceedings{dworkFairnessAwareness2012,
  title = {Fairness through {{Awareness}}},
  booktitle = {Proceedings of the 3rd Innovations in Theoretical Computer Science Conference},
  author = {Dwork, Cynthia and Hardt, Moritz and Pitassi, Toniann and Reingold, Omer and Zemel, Richard},
  year = {2012},
  pages = {214--226}
}

@article{guoLearningFairRepresentations2022,
  title = {Learning {{Fair Representations}} via {{Distance Correlation Minimization}}},
  author = {Guo, Dandan and Wang, Chaojie and Wang, Baoxiang and Zha, Hongyuan},
  year = {2022},
  journal = {IEEE Transactions on Neural Networks and Learning Systems},
  pages = {1--14},
  //issn = {2162-2388},
  //doi = {10.1109/TNNLS.2022.3187165},
  //keywords = {Correlation,Distance correlation,Encoding,fair representation learning,Feature extraction,graph Laplacian regularization,Laplace equations,Learning systems,Representation learning,sensitive attributes,Task analysis}
}

@inproceedings{hardtEqualityOpportunitySupervised2016,
  title = {Equality of {{Opportunity}} in {{Supervised Learning}}},
  booktitle = {Advances in Neural Information Processing Systems},
  author = {Hardt, Moritz and Price, Eric and Price, Eric and Srebro, Nati},
  year = {2016},
  volume = {29}
}

@inproceedings{heDeepResidualLearning2016a,
  title = {Deep {{Residual Learning}} for {{Image Recognition}}},
  booktitle = {Proceedings of the {{IEEE}} Conference on Computer Vision and Pattern Recognition},
  author = {He, Kaiming and Zhang, Xiangyu and Ren, Shaoqing and Sun, Jian},
  year = {2016},
  pages = {770--778}
}

@inproceedings{kimLearningNotLearn2019,
  title = {Learning {{Not}} to {{Learn}}: {{Training Deep Neural Networks With Biased Data}}},
  shorttitle = {Learning {{Not}} to {{Learn}}},
  booktitle = {Proceedings of the {{IEEE}} Conference on Computer Vision and Pattern Recognition},
  author = {Kim, Byungju and Kim, Hyunwoo and Kim, Kyungsu and Kim, Sungjin and Kim, Junmo},
  year = {2019},
  pages = {9004--9012},
}

@article{kingmaAutoencodingVariationalBayes2013,
  title = {{{Auto-Encoding Variational Bayes}}},
  author = {Kingma, Diederik P. and Welling, Max},
  year = {2013},
  journal = {arXiv preprint arXiv:1312.6114},
  eprint = {1312.6114},
  archiveprefix = {arxiv}
}

@article{leeMaximalCorrelationFramework2022,
  title = {A {{Maximal Correlation Framework}} for {{Fair Machine Learning}}},
  author = {Lee, Joshua and Bu, Yuheng and Sattigeri, Prasanna and Panda, Rameswar and Wornell, Gregory W. and Karlinsky, Leonid and Schmidt Feris, Rogerio},
  year = {2022},
  journal = {Entropy},
  volume = {24},
  number = {4},
  pages = {461},
  //issn = {1099-4300},
  //doi = {10.3390/e24040461},
  urldate = {2022-12-22},
  langid = {english},
  //keywords = {ObsCite}
}

@inproceedings{liuDeepLearningFace2015,
  title = {Deep {{Learning Face Attributes}} in the {{Wild}}},
  booktitle = {Proceedings of the IEEE International Conference on Computer Vision},
  author = {Liu, Ziwei and Luo, Ping and Wang, Xiaogang and Tang, Xiaoou},
  year = {2015},
  pages = {3730--3738},
  urldate = {2023-04-02}
}

@inproceedings{liuFairRepresentationLearning2022,
  title = {Fair Representation Learning: An Alternative to Mutual Information},
  shorttitle = {Fair {{Representation Learning}}},
  booktitle = {Proceedings of the 28th ACM SIGKDD Conference on Knowledge Discovery and Data Mining},
  author = {Liu, Ji and Li, Zenan and Yao, Yuan and Xu, Feng and Ma, Xiaoxing and Xu, Miao and Tong, Hanghang},
  year = {2022},
  //series = {{{KDD}} '22},
  pages = {1088--1097},
  urldate = {2023-04-02},
  keywords = {distance covariance,fair representation learning,mutual information}
}

@inproceedings{maryFairnessAwareLearningContinuous2019a,
  title = {Fairness-Aware Learning for Continuous Attributes and Treatments},
  booktitle = {Proceedings of the 36th International Conference on Machine Learning},
  author = {Mary, Jeremie and Calauz{\`e}nes, Cl{\'e}ment and Karoui, Noureddine El},
  year = {2019},
  pages = {4382--4391},
  publisher = {{PMLR}},
  address = {CA},
  issn = {2640-3498},
  urldate = {2023-04-11},
  langid = {english}
}

@inproceedings{mehrabiAttributingFairDecisions2022,
  title = {Attributing Fair Decisions with Attention Interventions},
  booktitle = {Proceedings of the 2nd Workshop on Trustworthy Natural Language Processing (TrustNLP 2022)},
  author = {Mehrabi, Ninareh and Gupta, Umang and Morstatter, Fred and Steeg, Greg Ver and Galstyan, Aram},
  year = {2022},
  pages = {12--25}
}

@article{mehrabiSurveyBiasFairness2022,
  title = {A {{Survey}} on {{Bias}} and {{Fairness}} in {{Machine Learning}}},
  author={Mehrabi, Ninareh and Morstatter, Fred and Saxena, Nripsuta and Lerman, Kristina and Galstyan, Aram},
  journal={ACM computing surveys (CSUR)},
  volume={54},
  number={6},
  pages={1--35},
  year={2021},
  publisher={ACM New York, NY, USA}
}

@article{moyerInvariantRepresentationsAdversarial2019,
  title = {Invariant {{Representations}} without {{Adversarial Training}}},
  author = {Moyer, Daniel and Gao, Shuyang and Brekelmans, Rob and Steeg, Greg Ver and Galstyan, Aram},
  journal={Advances in Neural Information Processing Systems},
  volume={31},
  year={2018}
}

@inproceedings{parkFairContrastiveLearning2022,
  title = {Fair Contrastive Learning for Facial Attribute Classification},
  booktitle = {Proceedings of the IEEE Conference on Computer Vision and Pattern Recognition},
  author = {Park, Sungho and Lee, Jewook and Lee, Pilhyeon and Hwang, Sunhee and Kim, Dohyung and Byun, Hyeran},
  year = {2022},
  pages = {10389--10398},
  urldate = {2022-10-11},
  langid = {english}
}

@article{parkLearningDisentangledRepresentation2021,
  title = {Learning Disentangled Representation for Fair Facial Attribute Classification via Fairness-aware Information Alignment},
  author = {Park, Sungho and Hwang, Sunhee and Kim, Dohyung and Byun, Hyeran},
  year = {2021},
  journal = {Proceedings of the AAAI Conference on Artificial Intelligence},
  volume = {35},
  number = {3},
  pages = {2403--2411}
}

@inproceedings{pleissFairnessCalibration2017,
  title = {On Fairness and Calibration},
  booktitle = {Advances in Neural Information Processing Systems},
  author = {Pleiss, Geoff and Raghavan, Manish and Wu, Felix and Kleinberg, Jon and Weinberger, Kilian Q.},
  year = {2017},
  volume = {30},
  //publisher = {{Curran Associates, Inc.}},
  urldate = {2023-02-25}
}

@inproceedings{ramaswamyFairAttributeClassification2021,
  title = {Fair Attribute Classification through Latent Space De-biasing},
  booktitle = {2021 IEEE/CVF Conference on Computer Vision and Pattern Recognition (CVPR)},
  author = {Ramaswamy, Vikram V. and Kim, Sunnie S. Y. and Russakovsky, Olga},
  year = {2021},
  pages = {9297--9306},
  //publisher = {{IEEE}},
  //address = {{Nashville, TN, USA}},
  //doi = {10.1109/CVPR46437.2021.00918},
  urldate = {2023-02-25},
  //isbn = {978-1-66544-509-2},
  langid = {english}
}

@article{sattigeriFairnessGANGenerating2019,
  title = {Fairness {{GAN}}: {{Generating}} Datasets with Fairness Properties Using a Generative Adversarial Network},
  shorttitle = {Fairness {{GAN}}},
  author = {Sattigeri, Prasanna and Hoffman, Samuel C. and Chenthamarakshan, Vijil and Varshney, Kush R.},
  year = {2019},
  journal = {IBM Journal of Research and Development},
  volume = {63},
  number = {4/5},
  pages = {3--1},
  publisher = {{IBM}}
}

@book{serflingApproximationTheoremsMathematical2009,
  title = {Approximation Theorems of Mathematical Statistics},
  author = {Serfling, Robert J.},
  year = {2009},
  address = {New York},
  publisher = {{John Wiley \& Sons}}
}

@inproceedings{songLearningControllableFair2020a,
  title = {Learning {{Controllable Fair Representations}}},
  author = {Song, Jiaming and Kalluri, Pratyusha and Grover, Aditya and Zhao, Shengjia and Ermon, Stefano},
  booktitle={The 22nd International Conference on Artificial Intelligence and Statistics},
  pages={2164--2173},
  year={2019},
  organization={PMLR}
}

@article{szekelyMeasuringTestingDependence2007,
  title = {Measuring and Testing Dependence by Correlation of Distances},
  author = {Sz{\'e}kely, G{\'a}bor J. and Rizzo, Maria L. and Bakirov, Nail K.},
  year = {2007},
  journal = {The Annals of Statistics},
  volume = {35},
  number = {6},
  eprint = {0803.4101},
  primaryclass = {math, stat},
  //issn = {0090-5364},
  //doi = {10.1214/009053607000000505},
  urldate = {2023-01-13},
  archiveprefix = {arxiv},
  keywords = {62G10 (Primary) 62H20 (Secondary),Mathematics - Statistics Theory}
}

@inproceedings{xuFairganFairnessawareGenerative2018,
  title = {Fairgan: Fairness-aware Generative Adversarial Networks},
  shorttitle = {Fairgan},
  booktitle = {2018 IEEE International Conference on Big Data (Big Data)},
  author = {Xu, Depeng and Yuan, Shuhan and Zhang, Lu and Wu, Xintao},
  year = {2018},
  pages = {570--575},
  address = {WA},
  publisher = {{IEEE}}
}

@inproceedings{zafarFairnessConstraintsMechanisms2017a,
  title = {Fairness {{Constraints}}: {{Mechanisms}} for {{Fair Classification}}},
  shorttitle = {Fairness Constraints},
  booktitle = {Artificial Intelligence and Statistics},
  author = {Zafar, Muhammad Bilal and Valera, Isabel and Rogriguez, Manuel Gomez and Gummadi, Krishna P.},
  year = {2017},
  pages = {962--970},
  address = {Fort Lauderdale},
  publisher = {{PMLR}}
}

@inproceedings{zhangAgeProgressionRegression2017,
  title = {Age Progression/Regression by Conditional Adversarial Autoencoder},
  booktitle = {Proceedings of the IEEE Conference on Computer Vision and Pattern Recognition},
  author = {Zhang, Zhifei and Song, Yang and Qi, Hairong},
  year = {2017},
  pages = {5810--5818},
  urldate = {2023-04-19}
}

@article{vershynin2010introduction,
  title={Introduction to the non-asymptotic analysis of random matrices},
  author={Vershynin, Roman},
  journal={arXiv preprint arXiv:1011.3027},
  year={2010}
}

@inproceedings{zhaoFairMetaLearningFewShot2020,
  title = {Fair {{Meta-Learning For Few-Shot Classification}}},
  author={Zhao, Chen and Li, Changbin and Li, Jincheng and Chen, Feng},
  booktitle={2020 IEEE international conference on knowledge graph (ICKG)},
  pages={275--282},
  year={2020},
  organization={IEEE}
}

@inproceedings{gupta2021controllable,
  title={Controllable guarantees for fair outcomes via contrastive information estimation},
  author={Gupta, Umang and Ferber, Aaron M. and Dilkina, Bistra and Ver Steeg, Greg},
  booktitle={Proceedings of the AAAI Conference on Artificial Intelligence},
  volume={35},
  pages={7610--7619},
  year={2021}
}

@article{wang2015conditional,
  title={Conditional distance correlation},
  author={Wang, Xueqin and Pan, Wenliang and Hu, Wenhao and Tian, Yuan and Zhang, Heping},
  journal={Journal of the American Statistical Association},
  volume={110},
  number={512},
  pages={1726--1734},
  year={2015},
  publisher={Taylor \& Francis}
}

@inproceedings{lowy2022stochastic,
  title={A Stochastic Optimization Framework for Fair Risk Minimization},
  author={Lowy, Andrew and Baharlouei, Sina and Pavan, Rakesh and Razaviyayn, Meisam and Beirami, Ahmad},
  booktitle={Workshop on Trustworthy and Socially Responsible Machine Learning, NeurIPS 2022},
  year={2022}
}

@inproceedings{poole2019variational,
  title={On variational bounds of mutual information},
  author={Poole, Ben and Ozair, Sherjil and Van Den Oord, Aaron and Alemi, Alex and Tucker, George},
  booktitle={International Conference on Machine Learning},
  pages={5171--5180},
  year={2019},
  organization={PMLR}
}

@misc{silverman1986density,
  title={Density estimation for statistics and data analysis},
  author={Silverman, Bernard W.},
  journal={Monographs on Statistics and Applied Probability},
  year={1986},
  publisher={CRC Press}
}

@article{han2023ffb,
  title={FFB: A Fair Fairness Benchmark for In-Processing Group Fairness Methods},
  author={Han, Xiaotian and Chi, Jianfeng and Chen, Yu and Wang, Qifan and Zhao, Han and Zou, Na and Hu, Xia},
  journal={arXiv preprint arXiv:2306.09468},
  year={2023}
}

@inproceedings{perez2017fair,
  title={Fair kernel learning},
  author={P{\'e}rez-Suay, Adri{\'a}n and Laparra, Valero and Mateo-Garc{\'\i}a, Gonzalo and Mu{\~n}oz-Mar{\'\i}, Jordi and G{\'o}mez-Chova, Luis and Camps-Valls, Gustau},
  booktitle={Joint European Conference on Machine Learning and Knowledge Discovery in Databases},
  pages={339--355},
  year={2017},
  organization={Springer}
}

@article{zhou2024fair,
  title={Fair Kernel K-Means: from Single Kernel to Multiple Kernel},
  author={Zhou, Peng and Li, Rongwen and Du, Liang},
  journal={Advances in Neural Information Processing Systems},
  volume={37},
  pages={99686--99714},
  year={2024}
}

@article{perez2023fair,
  title={Fair kernel regression through cross-covariance operators},
  author={Perez-Suay, Adrian and Gordaliza, Paula and Loubes, Jean-Michel and Sejdinovic, Dino and Camps-Valls, Gustau},
  journal={Transactions on Machine Learning Research},
  year={2023}
}

@inproceedings{lahotifairnessdemographics2020,
  title = {Fairness without demographics through adversarially reweighted learning},
  booktitle = {Advances in Neural Information Processing Systems},
  author = {Lahoti, Preethi and Beutel, Alex and Chen, Jilin and Lee, Kang and Prost, Flavien and Thain, Nithum and Wang, Xuezhi and Chi, Ed },
  year = {2020},
  volume = {33},
  pages = {728-740}
}

@inproceedings{chai2022fairness,
  title={Fairness with adaptive weights},
  author={Chai, Junyi and Wang, Xiaoqian},
  booktitle={International conference on machine learning},
  pages={2853--2866},
  year={2022},
  organization={PMLR}
}

@article{shen2021contrastive,
  title={Contrastive learning for fair representations},
  author={Shen, Aili and Han, Xudong and Cohn, Trevor and Baldwin, Timothy and Frermann, Lea},
  journal={arXiv preprint arXiv:2109.10645},
  year={2021}
}

@inproceedings{zhang2018mitigating,
  title={Mitigating unwanted biases with adversarial learning},
  author={Zhang, Brian Hu and Lemoine, Blake and Mitchell, Margaret},
  booktitle={Proceedings of the 2018 AAAI/ACM Conference on AI, Ethics, and Society},
  pages={335--340},
  year={2018}
}

@article{gretton2005kernel,
  title={Kernel methods for measuring independence},
  author={Gretton, Arthur and Herbrich, Ralf and Smola, Alexander and Bousquet, Olivier and Sch{\"o}lkopf, Bernhard},
  year={2005},
  journal={Journal of Machine Learning Research},
  volumn={6},
  pages={2075--2129},
  publisher={MIT Press}
}

%%%%%%%%%%%%%%%%%%%%%%%%%%%%%%%%%%%%%%%%%%%%%%%%%%%%%%%%%%%%%%%%%%%%%%%%%%%%%%%
%%%%%%%%%%%%%%%%%%%%%%%%%%%%%%%%%%%%%%%%%%%%%%%%%%%%%%%%%%%%%%%%%%%%%%%%%%%%%%%
% APPENDIX
%%%%%%%%%%%%%%%%%%%%%%%%%%%%%%%%%%%%%%%%%%%%%%%%%%%%%%%%%%%%%%%%%%%%%%%%%%%%%%%
%%%%%%%%%%%%%%%%%%%%%%%%%%%%%%%%%%%%%%%%%%%%%%%%%%%%%%%%%%%%%%%%%%%%%%%%%%%%%%%
\appendix

\section{Some useful lemmas and definitions}
\label{appendix:proof}
\begin{lemma}[Theorem 3 in \cite{szekelyMeasuringTestingDependence2007}] \label{lemma:covariance}
	For any random variables $\mathbf{Y}\in\mathbb{R}^p$ and $\mathbf Z\in\mathbb{R}^q$ with $\mathbb{E}(\|\mathbf{Y}\|_2+\|\mathbf{Z}\|_2)<+\infty$, we have $\mathcal{V}^2(\mathbf{Y},\mathbf{Z}) = 0$ if and only if  $\mathbf{Y}$ and $\mathbf{Z}$ are independent.
\end{lemma}

Next, we present the proofs of convergence in probability with respect to the sample size. To begin, we introduce a definition and a lemma concerning the probability inequality of the $U$-statistic.
\begin{definition}[$U$-statistic, Subsection 5.1.1 in \cite{serflingApproximationTheoremsMathematical2009}]
    Let $Y_1,Y_2,\cdots$ be independent observations on a distribution $F$. For some function $h=h(y_1,\cdots,y_m)$, called a kernel, and $\theta=\theta(F)=\E(h(Y_1,\cdots,Y_m))$, the $U$-statistic for estimation of $\theta$ on the basis of a sample $Y_1,\cdots,Y_n$ of size $n\ge m$ is obtained by averaging the kernel $h$ symmetrically over the observations:
    \[U_n=U(Y_1,\cdots,Y_n)=\frac{1}{\tbinom{n}{m}}\sum_ch(Y_{i_1},\cdots,Y_{i_m}),\]
    where $\sum_c$ denotes summation over the $\tbinom n m$ combinations of $m$ distinct elements $\{i_1,\cdots,i_m\}$ from $\{1,\cdots,n\}$. Clearly, $U_n$ is an unbiased estimate of $\theta$.
\end{definition}
\begin{lemma}[Theorem 5.6.1.A of \cite{serflingApproximationTheoremsMathematical2009}]\label{u-statistic:covergence}
	Let $h(Y_1, \dots, Y_m)$ be a kernel function of the $U$-statistic $U_n$, and $\theta = E\{h(Y_1, \dots, Y_m)\}$. Let $a$ and $b$ be the upper and lower bounds of $h(Y_1, \dots, Y_m)$. That is, $a \le h(Y_1, \dots, Y_m) \le b$. For any $t>0$ and $n>m$,  we have 
	\begin{equation*}
		P(U_n - \theta \ge t) \le \exp\{-2[n/m]t^2/(b-a)^2\}, 
	\end{equation*}
	where $[n/m]$ denotes the greatest integer function, i.e. the integer part of $n/m$.
\end{lemma}
From Lemma \ref{u-statistic:covergence} and the symmetry of $U$-statistic, we can get the following result:
\begin{equation}\label{u-statistic:covergence:abs}
	P(|U_n - \theta | \ge t) \le 2\exp\{-2[n/m]t^2/(b-a)^2\},
\end{equation}
which is the key point of our proofs.

\begin{definition}[Sub-Gaussian random variables, Definition 5.7 in \cite{vershynin2010introduction}]
	A random variable $ v $ is called a sub-Gaussian random variable if one of the following 4 conditions holds:
	\begin{itemize}
		\item Tails: the probability $P(|v|>t)\le\exp(1-t^2/K^2_1)$ for all $t\ge 0$;
		\item Moments: $(\E|v|^p)^{1/p}\le K_2\sqrt{p}$ for all $p\ge 1;$
		\item Super-exponential moment: $\E\left(\exp\frac{v^2}{K^2_3}\right)\le e$;
	\end{itemize}
	Moreover, if $\E(v)=0$, the above 3 properties are equivalent to 
	\begin{itemize}
		\item Moment generating function:
		$ \E\exp\left(t v\right)\le \exp\left(\frac{K_4 t^2}{2}\right), \quad \forall\ t\in \mathbb{R}$.
	\end{itemize}
	where $K_i>0,i=1,2,3,4$ are constants differing from each other by at most an absolute constant factor. The sub-Gaussian norm of $v$, denoted $\|v\|_{\psi_2}$, is defined to be
	\[\|v\|_{\psi_2}=\sup_{p\ge 1}p^{-1/2}(\E|v-E v|^p)^{1/p}.\]
\end{definition}
In fact, the above 4 conditions are equivalent, see Lemma 5.5 of \cite{vershynin2010introduction}.

\begin{remark}\label{remark:sub-gaussian}
Actually, sub-Gaussian random variables are an extension of standard normal random variable. 
In addition, Classical examples of sub-gaussian random variables are Gaussian, Bernoulli
 and all bounded random variables. (see Example 5.8 in \cite{vershynin2010introduction}).
 \end{remark}
 
		\begin{definition}[Sub-Gaussian random vectors, Definition 5.22 of \cite{vershynin2010introduction}]
			We say that a random vector $\vv\in\F^d$ is sub-Gaussian if the one-dimensional marginals $\langle \vv,\vu\rangle$ are sub-Gaussian random variables for all $\vu\in\F^d$. The sub-Gaussian norm of $\vv$ is defined as
			\[\|\vv\|_{\psi_2}=\sup_{\|\vu\|=1}\|\langle \vv,\vu\rangle\|_{\psi_2}.\]
		\end{definition}
%#########################################################################################

		\begin{lemma}
			\label{lem:subgaussian}
			Let $\Y,\Z$ be two sub-Gaussian random vectors. For any $t>0$, there exists a constant $K_1>0$ such that 
			\begin{equation*}
				P(\|\Y\|_2 > t) \le e\cdot\exp(-t^2/K_1),\ P(\|\Z\|_2 > t) \le e\cdot\exp(-t^2/K_1).
			\end{equation*}
		\end{lemma}
		\begin{proof}
		By the definition of sub-Gaussian random vector, we have $\|\Y\|_1=\langle \Y,\sign(\Y)\rangle$ is a sub-Gaussian random variable if $\Y$ is a sub-Gaussian random vector. Therefore, there exits a positive constant $K_1$ s.t.
		\begin{equation*}\label{sub gauss}
			P(\|\Y\|_1 > t) \le e\cdot\exp(-t^2/K_1).
		\end{equation*}
		Note that $\|\Y\|_2\le\|\Y\|_1$, so $P(\|\Y\|_2 > t)\le P(\|\Y\|_1 > t) \le e\cdot\exp(-t^2/K_1)$. Similarly, we have $P(\|\Z\|_2 > t)\le P(\|\Z\|_1 > t) \le e\cdot\exp(-t^2/K_1)$ since $\Z$ is a sub-Gaussian random vector.
		\end{proof}	
%&&&&&&&&&&&&&&&&&&&&&&&&&&&&&&&&&&&&&&&&&&&&&&&&&&&&&&&&&&&&&&&&&&&&&&&&&&
\section{Proof of Proposition \ref{pro:matrix1}} \label{sec:proof_matrix_form}	
\begin{proof}
By Definition \ref{def:pcdcov}, we have
$$\Ss_n(Y,Z,U) =\frac{12}{n}\sum_{u=1}^{n}\frac{\left(\sum^n_{i=1}K_{iu}\right)^4}{n^4}(D_1+D_2-2D_3),$$
 where $\omega_{ku}=K_{ku}/\sum_k K_{ku}$, $d_{kl}^Y = \Vert Y_k-Y_l \Vert_2$ and $d_{kl}^Z = \Vert Z_k-Z_l \Vert_2$, $D_1=\sum^n_{k,\ell}d^Y_{k\ell}d^Z_{k\ell}\omega_{ku}\omega_{\ell u}$, $D_2=\sum^n_{k,\ell}d^Y_{k\ell}\omega_{ku}\omega_{\ell u}\sum^n_{k,\ell}d^Z_{k\ell}\omega_{ku}\omega_{\ell u}$, $D_3=\sum^n_{k,\ell,m}d^Y_{k\ell}d^Z_{km}\omega_{ku}\omega_{\ell u}\omega_{mu}$. Therefore,
 \begin{align*}
 \Ss_n(Y,Z,U) =\frac{12}{n^5}(E_1+E_2+E_3),
 \end{align*}
 where $E_1=\sum^n_{u=1}\left(\sum^n_{i=1}K_{iu}\right)^2\sum^n_{k,\ell=1}d^Y_{k\ell}d^Z_{k\ell}K_{ku}K_{\ell u}$, $E_2=\sum^n_{u=1}\sum^n_{k,\ell}d^Y_{k\ell}K_{ku}K_{\ell u}\sum^n_{k,\ell}d^Z_{k\ell}K_{ku}K_{\ell u}$, $E_3=\sum^n_{u=1}\left(\sum^n_{i=1}K_{iu}\right)\sum^n_{k,\ell,m}d^Y_{k\ell}d^Z_{km}K_{ku}K_{\ell u}K_{mu}$. Note that $K\in\R^{n\times n}$ with $K_{k\ell}=\frac{1}{\sqrt{2\pi}\sigma}e^{-\frac{\|U_k-U_\ell\|^2}{2\sigma^2}}$. Define $K_u=\begin{bmatrix}K_{1u}&\cdots&K_{nu}
 \end{bmatrix}^{\top}$. Next, we will calculate $E_1$, $E_2$ and $E_3$.
 \begin{align*}
 E_1\xlongequal{D=D^Y\odot D^Z=(d_{k\ell})}{}&\sum^n_{u=1}\left(\sum^n_{i=1}K_{iu}\right)^2\sum^n_{k,\ell=1}d_{k\ell}K_{ku}K_{\ell u}=\sum^n_{u=1}\left(\sum^n_{i=1}K_{iu}\right)^2K^{\top}_uDK_u\\
 =&\langle (K\mathbbm{1})\odot(K\mathbbm{1}), \diag(K^{\top}DK)\rangle=\langle (K\mathbbm{1})\odot(K\mathbbm{1}), \diag(K^{\top}(D^Y\odot D^Z)K)\rangle.
 \end{align*}
 \begin{align*}
 E_2=&\sum^n_{u=1}(K^{\top}_uD^YK_u)(K^{\top}_uD^ZK_u)=\langle\diag(K^{\top}D^YK), \diag(K^{\top}D^ZK)\rangle.
 \end{align*}
 \begin{align*}
 E_3=&\sum^n_{u=1}\left(\sum^n_{i=1}K_{iu}\right)\sum^n_{k}K_{ku}\left(\sum^n_{\ell}d^Y_{k\ell}K_{\ell u}\right)\left(\sum^n_{m}d^Z_{km}K_{mu}\right)\\
 \xlongequal[E^Z=D^ZK]{E^Y=D^YK}&\sum^n_{u=1}\left(\sum^n_{i=1}K_{iu}\right)\sum^n_{k=1}E^Y_{ku}E^Z_{ku}K_{ku}\\
 \xlongequal{E=E^Y\odot E^Z}&\sum^n_{u=1}\left(\sum^n_{i=1}K_{iu}\right)\sum^n_{k=1}E_{ku}K_{ku}=\sum^n_{u=1}\left(\sum^n_{i=1}K_{iu}\right)K^{\top}_uE_u=\langle K\mathbbm{1},\diag(K^{\top}E)\rangle\\
 =&\langle K\mathbbm{1},\diag(K^{\top}((D^YK)\odot (D^ZK)))\rangle,
 \end{align*}
 where $E_u$ is the $u$-th column of $E$. In summary, we conclude the result.
 \end{proof}

%&&&&&&&&&&&&&&&&&&&&&&&&&&&&&&&&&&&&&&&&&&&&&&&&&&&&&&&&&&&&&&&&&&&&&&&&&&
\section{Proof of Theorem \ref{theo:converge}}	\label{p:converge}	
\begin{proof}[Proof of Theorem \ref{theo:converge}]
		Let $(\tilde{\Y},\tilde{\Z})$ be an i.i.d copy of $(\Y, \Z)$. From Remark 3 of \cite{szekelyMeasuringTestingDependence2007}, we have	
		\begin{equation*}
			\mathcal{V}^2(\Y,\Z) = S_1 + S_2 - 2S_3,
		\end{equation*}
		where $S_i$, $i=1,2,3$ are defined as follows:
		\begin{align*}
			S_1  = \E \Vert \Y - \tilde{\Y} \Vert_2 \Vert \Z - \tilde{\Z} \Vert_2, \ 
			S_2  = \E \Vert \Y - \tilde{\Y} \Vert_2 \E \Vert \Z - \tilde{\Z} \Vert_2, \ 
			S_3  = \E \left[ \E \Vert \Y - \tilde{\Y} \Vert_2 \big| \Y \right]  \E \left[ \E  \Vert \Z - \tilde{\Z} \Vert_2 \big| \Z \right].
		\end{align*}
		From Equation (2.18) in \cite{szekelyMeasuringTestingDependence2007}, we have the empirical distance covariance $\mathcal{V}^2_n(Y,Z) = \hat{S}_1 + \hat{S}_2 - 2\hat{S}_3$, where
		\begin{align*}
			& \hat{S}_1 = \frac{1}{n^2}\sum_{i,j=1}^{n} \Vert Y_i - Y_j \Vert_2 \Vert Z_i - Z_j \Vert_2, \\
			& \hat{S}_2 = \frac{1}{n^2}\sum_{i,j=1}^{n} \Vert Y_i - Y_j \Vert_2 \: \frac{1}{n^2}\sum_{i,j=1}^{n} \Vert  Z_i - Z_j  \Vert_2, \\
			& \hat{S}_3 = \frac{1}{n^3} \sum_{i,j,\ell=1}^{n} \Vert Y_i - Y_j \Vert_2 \Vert Z_i - Z_\ell \Vert_2.
		\end{align*}
Then
		\begin{equation*}
			\begin{split}
				&P(|\mathcal{V}^2_{n}(Y,Z)-\mathcal{V}^2(\Y,\Z)|>\epsilon)= P(\{ | (\hat{S}_1 + \hat{S}_2 - 2\hat{S}_3) - S_1 + S_2 - 2 S_3 | > \epsilon \})  		\\
				\le& P(\{ | \hat{S}_1 - S_1  | > \epsilon/4 \}) + P(\{ | \hat{S}_2 - S_2  | > \epsilon/4 \}) + P(\{ | \hat{S}_3 - S_3 | > \epsilon/4 \}) 	
				:= E_1+E_2+E_3,
			\end{split}
		\end{equation*}
		where $E_1=P(\{ | \hat{S}_1 - S_1 | > \epsilon/4 \})$, $E_2=P(\{ | \hat{S}_2 - S_2  | > \epsilon/4 \})$, $E_3= P(\{ | \hat{S}_3 - S_3  | > \epsilon/4 \}) $. We only need to estimate the upper bounds of $E_1$, $E_2$ and $E_3$, respectively. Combining Lemmas \ref{lem:s1_estimate}, \ref{lem:s2_estimate} and \ref{lem:s3_estimate}, we conclude the result.
		\end{proof}
\begin{lemma}
	\label{lem:s1_estimate}
Let $S_1  = \E \Vert \Y - \tilde{\Y} \Vert_2 \Vert \Z - \tilde{\Z} \Vert_2$, $ \hat{S}_1 = \frac{1}{n^2}\sum_{i,j=1}^{n} \Vert Y_i - Y_j \Vert_2 \Vert Z_i - Z_j \Vert_2$. For any $\epsilon>0$, we have $P(\{ | \hat{S}_1 - S_1 | > \epsilon/4 \}) \le  \frac {2}n+2e n\cdot\exp\left(-\frac{\epsilon}{32K^2_1}\sqrt{\frac n{\log n}}\right)$.
	
\end{lemma}
\begin{proof} 
Choose the kernel functions 
$h_1(Y_i, Y_j,Z_i,Z_j) = \Vert Y_i - Y_j \Vert_2\Vert Z_i - Z_j \Vert_2$, and construct corresponding $U$-statistics $\hat U_1  = \frac{1}{n(n-1)} \sum_{i \neq j} h_1(Y_i, Y_j,Z_i,Z_j)$ of $S_1$, then $S_1=\E(\hat U_1)$. Denote $$\tilde{S}_1=\frac{1}{n^2}\sum_{i,j=1}^{n} h_1(Y_i, Y_j,Z_i,Z_j)\mathbb{I}(h_1(Y_i, Y_j,Z_i,Z_j)\le M_1)$$ and $\tilde{U}_1=\frac{1}{n(n-1)}\sum_{i\ne j} h_1(Y_i, Y_j,Z_i,Z_j)\mathbb{I}(h_1(Y_i, Y_j,Z_i,Z_j)\le M_1)$, where $\mathbb{I}(\cdot)$ is an indicator function and $M_1=\sqrt{\frac n{\log n}}\frac{\epsilon}{8}$ is a constant.	
		Define sets 
		\begin{equation*}
			\begin{aligned}
				&	G_1 = \{ | \tilde{S}_1 - \E(\hat U_1) | \le \epsilon/4 \},	\		G_2 = \{ \tilde{S}_1 = \hat{S}_1,\tilde U_1=\hat U_1 \},		\	G_3 = \{ \Vert Y_i - Y_j \Vert_2 \Vert Z_i -Z_j \Vert_2 \le M_1, \; \forall i,j \},		\\
				&	G_4 = \{ \Vert Y_i \Vert_2^2 + \Vert Z_i \Vert_2^2 \le M_1/2, \; \forall i \}	,	\	G   = \{ | \hat{S}_1 - \E(\hat U_1)  | \le \epsilon/4 \},
			\end{aligned}
		\end{equation*}
		we have
		\begin{equation*}
			G_1 \cap G_2 \subset G,\ 		
			G_3 \subset G_2,\ 		
			G_4 \subseteq G_3,	
		\end{equation*}
		where $G_4 \subseteq G_3$ can obtain since
		\begin{equation*}
			\begin{split}
				& \Vert Y_i - Y_j \Vert_2 \Vert Z_i -Z_j \Vert_2 \le \frac{\Vert Y_i - Y_j \Vert_2^2 + \Vert Z_i - Z_j \Vert_2^2}{2}		\\
				=& \frac{1}{2} \left( \Vert Y_i \Vert_2^2 + \Vert Y_j \Vert_2^2  - 2Y^{\top}_iY_j + \Vert Z_i \Vert_2^2 + \Vert Z_j \Vert_2^2  - 2Z^{\top}_iZ_j \right)		\\
				\le&  \left(\Vert Y_i \Vert_2^2 + \Vert Y_j \Vert_2^2 + \Vert Z_i \Vert_2^2 + \Vert Z_j \Vert_2^2  \right)\le 2\ \underset{i}{\mathrm{max}} \: (\Vert Y_i \Vert_2^2 +\Vert Z_i \Vert_2^2).
			\end{split}
		\end{equation*}
		Thus, $G_3^c \subseteq G_4^c$  ($ S^c$ denote the complementary set of the set $S$).

		By the inclusion relation of the sets, we establish the following probability inequalities:
		
		\begin{equation}\label{ps1}
			\begin{aligned}
				&P(G^c) \le P(G_1^c \cup G_2^c)  = P\left(G_1^c \cap (G_2 \cup G_2^c) \cup G_2^c \right)	 = P\left( (G_1^c \cap G_2) \cup (G_1^c \cap G_2^c ) \cup G_2^c \right)		\\
				 \le &P\left( (G_1^c \cap G_2) \cup G_2^c \right)	 \le P (G_1^c \cap G_2) + P(G_2^c) 		 \le P (G_1^c \cap G_2) + P(G_3^c)	 \le P (G_1^c \cap G_2) + P(G_4^c).
			\end{aligned}
		\end{equation}
Note that the kernel size $m=2$, by Lemma \ref{u-statistic:covergence}, we obtain % there exists a constant $C>0$ 
\begin{align}
	P(|\tilde U_1 - \E(\tilde U_1)|\ge \epsilon/8) & \le 2\exp\left\{-2[n/m]\frac{\epsilon^2}{64}/\left(\frac{n\epsilon^2}{64\log n}\right)\right\} = \frac {2}n \label{u-stat}.
\end{align}
Since
\begin{align*}
	0<\tilde U_1 - \tilde{S}_1 & = \frac{1}{n(n-1)} \sum_{i \neq j} h_1(Y_i, Y_j, Z_i, Z_j)\mathbb{I}(h_1(Y_i, Y_j,Z_i,Z_j) - \frac{1}{n^2} \sum_{i,j}^{n} h_1(Y_i, Y_j, Z_i, Z_j)\mathbb{I}(h_1(Y_i, Y_j,Z_i,Z_j)		\\
	& = \left( \frac{1}{n(n-1)} - \frac{1}{n^2} \right) \sum_{i \neq j} h_1(Y_i, Y_j, Z_i, Z_j) \mathbb{I}(h_1(Y_i, Y_j,Z_i,Z_j)\\
	& \le \left( \frac{1}{n(n-1)} - \frac{1}{n^2} \right) (n(n-1))M=\frac Mn=\frac{\epsilon}{8\sqrt{n\log n}}<\frac\epsilon 8,
\end{align*}
when $n\ge 2$, so $P(\{ | \tilde{S}_1 -\tilde U_1 | > \frac{\epsilon}{8} \} = 0$. Thus, from (\ref{u-stat}), get
\begin{equation}\label{pg1cg2}
	\begin{aligned}
		P (G_1^c \cap G_2)&=P(\{ | \tilde{S}_1 - \E(\tilde U_1) | >\frac \epsilon 4\})  = P(\{ | \tilde{S}_1 -\tilde U_1 + \tilde U_1- \E(\tilde U_1)| > \frac \epsilon 4\})		\\
		& \le P(\{ | \tilde{S}_1 -\tilde U_1 | > \frac{\epsilon}{8} \} ) + P(\{ | \tilde U_1- \E(\tilde U_1) | > \frac{\epsilon}{8} \}) \le \frac 2n.
	\end{aligned}
\end{equation}
		Next, we estimate $P(G_4^c)$. We have
		\begin{equation}\label{pg4}
			\begin{aligned}
				&P(G_4^c)  = P(\{ \Vert Y_i \Vert_2^2 + \Vert Z_i \Vert_2^2 > M_1/2, \; \forall i \} )		\\
				\le&  P(\{ \Vert Y_i \Vert_2^2 > \frac{M_1}{4}, \; \forall i \} ) + P(\{ \Vert Z_i \Vert_2^2 > \frac{M_1}{4}, \; \forall i \} )		\\
				=& P(\{ \Vert Y_i \Vert_2 > \frac{\sqrt{M_1}}{2}, \; \forall i \} ) + P(\{ \Vert Z_i \Vert_2 > \frac{\sqrt{M_1}}{2}, \; \forall i \} )\\
				\le& 2e\cdot n\exp\left(-\frac{M_1}{4K^2_1}\right)\le 2e\cdot n\exp\left(-\frac{\epsilon}{32K^2_1}\sqrt{\frac n{\log n}}\right),
			\end{aligned}
		\end{equation}
		where the last inequality is based on Lemma \ref{lem:subgaussian}, since $Y_i$, $Z_i$ are the Sub-Gaussian vectors.
		
		Thus, combining (\ref{ps1}), (\ref{pg1cg2}), (\ref{pg4}), we have
		\begin{equation*}
			\begin{aligned}
				P(\{ | \hat{S}_1 - S_1) | > \epsilon \}) = P(G^c)\le P (G_1^c \cap G_2) + P(G_4^c)	
				\le  \frac {2}n+2e n\cdot\exp\left(-\frac{\epsilon}{32K^2_1}\sqrt{\frac n{\log n}}\right).
			\end{aligned}
		\end{equation*}
\end{proof}

\begin{lemma}
\label{lem:s2_estimate}
Let $S_2  = \E \Vert \Y - \tilde{\Y} \Vert_2 \E \Vert \Z - \tilde{\Z} \Vert_2$, $\hat{S}_2 = \frac{1}{n^2}\sum_{i,j=1}^{n} \Vert Y_i - Y_j \Vert_2 \: \frac{1}{n^2}\sum_{i,j=1}^{n} \Vert  Z_i - Z_j  \Vert_2$. For any $\epsilon>0$, we have $P(\{ | \hat{S}_2 - S_2 | > \epsilon \}) 		
			\le  \frac{2}{n^{18^2}}+2en\cdot\exp\left(-\frac{\epsilon}{576K^2_1}\sqrt{\frac n{\log n}}\right)$.
\end{lemma}
\begin{proof}
Next, we consider $\hat{S}_2$. The $U$-statistics related to $S_2$ with the kernel function $\hat h_2$, 
		\begin{align*}
			\hat{h}_2(Y_i, Z_i, Y_j, Z_j,Y_k, Z_k, Y_\ell, Z_\ell)  =& \frac{1}{6}(\Vert Y_i - Y_j \Vert_2 \Vert Z_k -Z_\ell \Vert_2 + \Vert Y_k - Y_j \Vert_2 \Vert Z_i -Z_\ell \Vert_2 	\\
			& + \Vert Y_i - Y_k \Vert_2 \Vert Z_j -Z_\ell \Vert_2 + \Vert Y_j - Y_\ell \Vert_2 \Vert Z_i -Z_k \Vert_2 	\\
			& + \Vert Y_\ell - Y_i \Vert_2 \Vert Z_k -Z_j \Vert_2 + \Vert Y_\ell - Y_k \Vert_2 \Vert Z_i -Z_j \Vert_2),   	\\
\hat U_2 &= \frac{1}{\binom{n}{4}}\sum_{i<j<k<\ell}\hat h_2 (Y_i, Z_i, Y_j, Z_j,Y_k, Z_k, Y_\ell, Z_\ell).
		\end{align*}
Let $M_2=\sqrt{\frac n{\log n}}\frac{\epsilon}{144}$ and $\mathbb{I}(\cdot)$ is an indicator function. Define $$\tilde h_2 (Y_i, Z_i, Y_j, Z_j,Y_k, Z_k, Y_l, Z_l)  = 	\hat{h}_2(Y_i, Z_i, Y_j, Z_j,Y_k, Z_k, Y_\ell, Z_\ell) \mathbb{I}(	\hat{h}_2(Y_i, Z_i, Y_j, Z_j,Y_k, Z_k, Y_\ell, Z_\ell) \le M_3),$$ and
		\begin{equation*}
			\tilde{S}_2=\frac{1}{n^4}\sum^n_{i,j,k,\ell=1}\tilde h_2 (Y_i, Z_i, Y_j, Z_j,Y_k, Z_k, Y_\ell, Z_\ell),\ \tilde U_2 = \frac{1}{\binom{n}{4}}\sum_{i<j<k<\ell}\tilde h_2 (Y_i, Z_i, Y_j, Z_j,Y_k, Z_k, Y_\ell, Z_\ell).
		\end{equation*}
Define sets:
\begin{equation*}
	\begin{aligned}
		&	G_1 = \{ | \tilde{S}_2 - \E(\hat{U}_2) | \le \epsilon/4 \},		\	G_2 = \{ \tilde{S}_2 = \hat{S}_2,\tilde{U}_2 = \hat{U}_2 \},		\	G_2 = \{ \hat{h}_2(Y_i, Z_i, Y_j, Z_j,Y_k, Z_k, Y_\ell, Z_\ell) \le M_2, \; \forall i,j,k,\ell \},		\\
		&	G_4 = \{ \Vert Y_i \Vert_2^2 + \Vert Z_i \Vert_2^2 \le M_2/2, \; \forall i \}	,	\	G   = \{ | \hat{S}_2 - \E(\hat{U}_2) | \le \epsilon/4 \}.
	\end{aligned}
\end{equation*}
Similar with (\ref{ps1}), we have $P(G^c)\le P(G^c_1\cap G_2)+P(G^c_4)$. Note that
\begin{align}\label{eq:est_us_tilde-2}
|\tilde U_2-\tilde S_2|=&\Bigg|4!\left(\frac{1}{n(n-1)(n-2)(n-3)}-\frac{1}{n^4}\right)\sum_{i<j<k<\ell}\tilde h_2 (Y_i, Z_i, Y_j, Z_j,Y_k, Z_k, Y_\ell, Z_\ell)\nonumber\\
&-\frac{4!}{n^4}\sum_{i=j<k<\ell}\tilde h_2 (Y_i, Z_i, Y_j, Z_j,Y_k, Z_k, Y_\ell, Z_\ell)\mathbb{I}(\hat{h}_2(Y_i, Z_i, Y_j, Z_j,Y_k, Z_k, Y_\ell, Z_\ell) \le M_2)\nonumber\\
&-\frac{4!}{n^4}\sum_{<j=k<\ell}\tilde h_2 (Y_i, Z_i, Y_j, Z_j,Y_k, Z_k, Y_\ell, Z_\ell)\mathbb{I}(\hat{h}_2(Y_i, Z_i, Y_j, Z_j,Y_k, Z_k, Y_\ell, Z_\ell) \le M_2)\nonumber\\
&-\frac{4!}{n^4}\sum_{i<j<k=\ell}\tilde h_2 (Y_i, Z_i, Y_j, Z_j,Y_k, Z_k, Y_\ell, Z_\ell)\mathbb{I}(\hat{h}_2(Y_i, Z_i, Y_j, Z_j,Y_k, Z_k, Y_\ell, Z_\ell) \le M_2)\Bigg|\nonumber\\
\le&\frac{(6n^2-11n+6)M_2}{n^3}+\frac{12 M_2(n-1)(n-2)}{n^3}\le\frac\epsilon 8.
\end{align}	
Combining the above inequality \eqref{eq:est_us_tilde-2} and Lemma  \ref{u-statistic:covergence}, we have 
		\begin{equation}\label{U3S3-2}
			\begin{aligned}
				P(G^c_1\cap G_2)=& P\left({\left| \tilde{S}_2 - \E(\tilde{U}_2) \right| > \epsilon/4}\right) =P\left({\left| \tilde{S}_2 -
\tilde U_2 + \tilde U_2 - \E(\tilde{U}_2) \right| > \epsilon/4}\right) 		\\
				\le & P\left(\left|\tilde{S}_2 - \tilde U_2 \right| > \epsilon/8 \right)  + P\left(\left| \tilde U_2 - \E(\tilde{U}_2) \right| > \epsilon/8\right)=P\left(\left| \tilde U_2 - \E(\tilde{U}_2) \right| > \epsilon/8\right)\le \frac{2}{n^{18^2}},
			\end{aligned}
		\end{equation}
Next, we estimate $P(G_4^c)$ by Lemma \ref{lem:subgaussian}. We have
\begin{equation}
\label{eq:h4c-2}
\begin{aligned}
P(G_4^c)=&P\left( \Vert Y_i \Vert_2^2 + \Vert Z_i \Vert_2^2 > M_2/2, \; \forall i \right)		\\
\le	& P\left( \Vert Y_i \Vert_2 > \frac{\sqrt{M_2}}{2}, \; \forall i  \right)  + P\left( \Vert Z_i \Vert_2 > \frac{\sqrt{M_2}}{2}, \; \forall i  \right) 
\le  2en\cdot\exp\left(-\frac{\epsilon}{576K^2_1}\sqrt{\frac n{\log n}}\right).	
\end{aligned}
\end{equation}		
Combining \eqref{U3S3-2} and \eqref{eq:h4c-2}, we can conclude the result.
\end{proof}

\begin{lemma}
\label{lem:s3_estimate}
Let $S_3  = \E \left[ \E \Vert \Y - \tilde{\Y} \Vert_2 \big| \Y \right]  \E \left[ \E \Vert_2 \Vert \Z - \tilde{\Z} \Vert_2 \big| \Z \right]$ and $\hat{S}_3 = \frac{1}{n^3} \sum_{i,j,\ell=1}^{n} \Vert Y_i - Y_j \Vert_2 \Vert Z_i - Z_\ell \Vert_2$. For any $\epsilon>0$, we have $P(\{ | \hat{S}_3 - S_3) | > \epsilon \}) 		
			\le  \frac{2}{n^{9^2}}+2en\cdot\exp\left(-\frac{\epsilon}{324K^2_1}\sqrt{\frac n{\log n}}\right)$.
\end{lemma}
\begin{proof}
Next, we construct the $U$-statistics related to $S_3$ with the kernel function $\hat h_3$, 
		\begin{align*}
			\hat{h}_3(Y_i, Z_i, Y_j, Z_j, Y_\ell, Z_\ell)  = &\frac{1}{6}(\Vert Y_i - Y_j \Vert_2 \Vert Z_i -Z_\ell \Vert_2 + \Vert Y_\ell - Y_j \Vert_2 \Vert Z_i -Z_\ell \Vert_2 	\\
			& + \Vert Y_i - Y_j \Vert_2 \Vert Z_j -Z_\ell \Vert_2 + \Vert Y_i - Y_\ell \Vert_2 \Vert Z_j -Z_\ell \Vert_2 	\\
			& + \Vert Y_\ell - Y_j \Vert_2 \Vert Z_i -Z_j \Vert_2 + \Vert Y_\ell - Y_i \Vert_2 \Vert Z_i -Z_j \Vert_2),   	\\
\hat U_3 &= \frac{1}{\binom{n}{3}}\sum_{i<j<\ell}\hat h_3 (Y_i, Z_i, Y_j, Z_j, Y_\ell, Z_\ell).
		\end{align*}
Let $M_3=\sqrt{\frac n{\log n}}\frac{\epsilon}{72}$ and $\mathbb{I}(\cdot)$ is an indicator function. Define $$\tilde h_3 (Y_i, Z_i, Y_j, Z_j, Y_l, Z_l)  = 	\hat{h}_3(Y_i, Z_i, Y_j, Z_j, Y_\ell, Z_\ell) \mathbb{I}(	\hat{h}_3(Y_i, Z_i, Y_j, Z_j, Y_\ell, Z_\ell) \le M_3),$$ and
		\begin{equation*}
			\tilde{S}_3=\frac{1}{n^3}\sum^n_{i,j,\ell=1}\tilde h_3 (Y_i, Z_i, Y_j, Z_j, Y_\ell, Z_\ell),\ \tilde U_3 = \frac{1}{\binom{n}{3}}\sum_{i<j<\ell}\tilde h_3 (Y_i, Z_i, Y_j, Z_j, Y_\ell, Z_\ell).
		\end{equation*}
Define sets:
\begin{equation*}
	\begin{aligned}
		&	H_1 = \{ | \tilde{S}_3 - \E(\hat{U}_3) | \le \epsilon/4 \},		\	H_2 = \{ \tilde{S}_3 = \hat{S}_3,\tilde{U}_3 = \hat{U}_3 \},		\	H_3 = \{ \hat{h}_3(Y_i, Z_i, Y_j, Z_j, Y_\ell, Z_\ell) \le M_3, \; \forall i,j,\ell \},		\\
		&	H_4 = \{ \Vert Y_i \Vert_2^2 + \Vert Z_i \Vert_2^2 \le M_3/2, \; \forall i \}	,	\	H   = \{ | \hat{S}_3 - \E(\hat{U}_3) | \le \epsilon/4 \}.
	\end{aligned}
\end{equation*}
Similar with (\ref{ps1}), we have $P(H^c)\le P(H^c_1\cap H_2)+P(H^c_4)$. Note that
\begin{align}\label{eq:est_us_tilde}
|\tilde U_3-\tilde S_3|=&\Bigg|3!\left(\frac{1}{n(n-1)(n-2)}-\frac{1}{n^3}\right)\sum_{i<j<\ell}\tilde h_3 (Y_i, Z_i, Y_j, Z_j, Y_\ell, Z_\ell)\nonumber\\
&-\frac{3!}{n^3}\sum_{i= j<\ell}\tilde h_3 (Y_i, Z_i, Y_j, Z_j, Y_\ell, Z_\ell)\mathbb{I}(\hat{h}_3(Y_i, Z_i, Y_j, Z_j, Y_\ell, Z_\ell) \le M_3)\\
&-\frac{3!}{n^3}\sum_{i<j=\ell}\tilde h_3 (Y_i, Z_i, Y_j, Z_j, Y_\ell, Z_\ell)\mathbb{I}(\hat{h}_3(Y_i, Z_i, Y_j, Z_j, Y_\ell, Z_\ell) \le M_3)\Bigg|\nonumber\\
\le&\frac{(3n-2)M_3}{n^2}+\frac{6 M_3}{n}\le\frac\epsilon 8.
\end{align}	
since $\hat{h}_3(Y_i, Z_i, Y_j, Z_j, Y_\ell, Z_\ell)=2\Vert Y_i - Y_\ell \Vert_2 \Vert Z_i -Z_j \Vert_2,\ j=\ell$. Combining the above inequality \eqref{eq:est_us_tilde} and Lemma  \ref{u-statistic:covergence}, we have 
		\begin{equation}\label{U3S3}
			\begin{aligned}
				P(H^c_1\cap H_2)=& P\left({\left| \tilde{S}_3 - \E(\tilde{U}_3) \right| > \epsilon/4}\right) =P\left({\left| \tilde{S}_3 -
\tilde U_3 + \tilde U_3 - \E(\tilde{U}_3) \right| > \epsilon/4}\right) 		\\
				\le & P\left(\left|\tilde{S}_3 - \tilde U_3 \right| > \epsilon/8 \right)  + P\left(\left| \tilde U_3 - \E(\tilde{U}_3) \right| > \epsilon/8\right)=P\left(\left| \tilde U_3 - \E(\tilde{U}_3) \right| > \epsilon/8\right)\le\frac{2}{n^{81}},
			\end{aligned}
		\end{equation}
Next, we estimate $P(H_4^c)$ by Lemma \ref{lem:subgaussian}. There exists a constant $C_\epsilon>0$, we have
\begin{equation}
\label{eq:h4c}
\begin{aligned}
P(H_4^c)=&P\left( \Vert Y_i \Vert_2^2 + \Vert Z_i \Vert_2^2 > M_3/2, \; \forall i \right)		\\
\le	& P\left( \Vert Y_i \Vert_2 > \frac{\sqrt{M_3}}{4}, \; \forall i  \right)  + P\left( \Vert Z_i \Vert_2 > \frac{\sqrt{M_3}}{4}, \; \forall i  \right) 
\le  2en\cdot\exp\left(-\frac{\epsilon}{324K^2_1}\sqrt{\frac n{\log n}}\right).	
\end{aligned}
\end{equation}		
Combining \eqref{U3S3} and \eqref{eq:h4c}, we can conclude the result.
\end{proof}

%###########################################################################

\section{Proof of Theorem \ref{p:cdc_convergence}}\label{appendix:cdc_converge}
\begin{proof}
Form Lemma \ref{lem:theo6_cdc}, we have
\begin{equation}
\label{eq:approximate}
|\E(\T_{n}(Y,  Z, U))-\mathcal{S}_{a}(\Y,  \Z, \U)|\le Ch^2
\end{equation}
 holds. 
We further use the result in Theorem \ref{theo:concentration_cdc}, for any $\epsilon>0$, we have 
\begin{equation}
\label{eq:concentrate}
|\mathcal{S}_{n}(Y,  Z, U)-\E(\T_{n}(Y,  Z, U))|\le \epsilon,
\end{equation}
with probability at least $1- \frac {8}{n^{120}}- 8en\cdot\exp\left(-\frac{\epsilon}{3840K_1}\sqrt{\frac n{\log n}}\right)$. Combining \eqref{eq:approximate} and \eqref{eq:concentrate}, we conclude the result.
\end{proof}

%#####################################################################################################
\begin{lemma}[The proof of Theorem 6 in supplementary material of paper \cite{wang2015conditional}]\label{lem:theo6_cdc}
    Let $\T_n(Y,  Z, U)$ be the corresponding $U$-statistic of $\Ss_n(Y,  Z, U)$. Then
\begin{equation}
%\label{eq:approximate_lem}
|\E(\T_{n}(Y,  Z, U))-\mathcal{S}_{a}(\Y,  \Z, \U)|\le Ch^2
\end{equation}
 holds, where $h$ is the bandwidth of the Gaussian kernel and $C$ is the upper bound of the second derivatives of both the density function of $\U$ and the conditional density function $f_{\Y,\Z|\U}$. 
\end{lemma}
\begin{theorem}
\label{theo:concentration_cdc}
Let $\mathbf Y\in\mathbb{R}^p$, $\mathbf Z\in\mathbb{R}^q$ be sub-Gaussian random vectors and $\mathbf U\in\mathbb{R}^r$ obey the regularity condition in Theorem \ref{p:cdc_convergence}.  Define the corresponding sample matrices  $Y=[Y_1,\cdots,Y_n]$, $Z=[Z_1,\cdots,Z_n]$, $U=[U_1,\cdots,U_n]$. Let $\Ss_n(Y,Z,U) = \frac{12}{n}\sum_{u=1}^{n}\left( \frac{\omega(U_u)}{n} \right)^4\mathcal{D}^2_n(Y,  Z \vert U)$ be the sample estimate and $\T_n(Y,  Z, U)$ be the corresponding $U$-statistic of $\Ss_n(Y,  Z, U)$. For any $\epsilon>0$, there exist positive constants $c$ and $c_\epsilon$, we have
	\begin{equation*}
P(|\mathcal{S}_{n}(Y,  Z, U)-\E[\T_{n}(Y,  Z, U)]|>\epsilon ) \le  \frac {8}{n^{120}}+ 8en\cdot\exp\left(-\frac{\epsilon}{3840K_1}\sqrt{\frac n{\log n}}\right).
	\end{equation*}
\end{theorem}
\begin{proof}
%%%%%
Define $K_{iu}=\frac 1hK\left(\frac{U_i-U_u}h\right)$. By Definition \ref{def:pcdcov}, we have $\Ss_n(Y,  Z, U) = \hat{S}_1 + \hat{S}_2 - 2\hat{S}_3 $, where
\begin{align*}
	 \hat{S}_1 =& \frac{12}{n^5} \sum_{i,j,k,\ell,u=1}^{n} \Vert Y_i - Y_j \Vert_2 \Vert Z_i - Z_j \Vert_2 K_{iu}K_{ju}K_{ku}K_{\ell u},\\
	\hat{S}_2 =& \frac{12}{n^5} \sum_{i,j,k,\ell,u=1}^{n} \Vert Y_i - Y_j \Vert_2 \Vert Z_k - Z_\ell \Vert_2 K_{iu}K_{ju}K_{ku}K_{\ell u},\\
	\hat{S}_3 =& \frac{12}{n^5} \sum_{i,j,k,\ell,u=1}^{n} \Vert Y_i - Y_j \Vert_2 \Vert Z_i - Z_k \Vert_2 K_{iu}K_{ju}K_{ku}K_{\ell u}.
\end{align*}
Define 
\begin{align*}
p_{ijk\ell u}=&\Vert Y_i - Y_j \Vert_2 \Vert Z_i - Z_j \Vert_2 K_{iu}K_{ju}K_{ku}K_{\ell u},\\
q_{ijk\ell u}=&\Vert Y_i - Y_j \Vert_2 \Vert Z_k - Z_\ell \Vert_2 K_{iu}K_{ju}K_{ku}K_{\ell u},\\
r_{ijk\ell u}=&\Vert Y_i - Y_j \Vert_2 \Vert Z_i - Z_k \Vert_2 K_{iu}K_{ju}K_{ku}K_{\ell u},
\end{align*} and 
\begin{align*}
\hat p_{ijk\ell u}=&\frac15(p_{ijk\ell u}+p_{ujk\ell i}+p_{iuk\ell j}+p_{iju\ell k}+p_{ijku\ell}),\\
\hat q_{ijk\ell u}=&\frac15(q_{ijk\ell u}+q_{ujk\ell i}+q_{iuk\ell j}+q_{iju\ell k}+q_{ijku\ell}),\\
\hat r_{ijk\ell u}=&\frac15(r_{ijk\ell u}+r_{ujk\ell i}+r_{iuk\ell j}+r_{iju\ell k}+r_{ijku\ell}).
\end{align*}  
Then the corresponding $U$-statistic of $\hat S_1$, $\hat S_2$ and $\hat S_3$ are 
\[\hat T_1=\frac{12}{\tbinom{n}{5}}\sum_{i<j<k<\ell<u}\hat p_{ijk\ell u},\ \hat T_2=\frac{12}{\tbinom{n}{5}}\sum_{i<j<k<\ell<u}\hat q_{ijk\ell u},\ \hat T_3=\frac{12}{\tbinom{n}{5}}\sum_{i<j<k<\ell<u}\hat r_{ijk\ell u},\]
and $\T_{n}(Y,  Z, U)=\hat T_1+\hat T_2-2\hat T_3$. Therefore, 
\begin{equation*}
	\begin{split}
		&P(|\mathcal{S}_{n}(Y,  Z, U)-\E(\T_{n}(Y,  Z, U))|>\epsilon)= P(\{ | (\hat{S}_1 + \hat{S}_2 - 2\hat{S}_3) - (\E(\hat T_1) +\E( \hat T_2) - 2\E(\hat T_3)) | > \epsilon \})  		\\
		\le& P(\{ | \hat{S}_1 - \E(\hat T_1)  | > \epsilon/4 \}) + P(\{ | \hat{S}_2 - \E(\hat T_2)  | > \epsilon/4 \}) + P(\{ | \hat{S}_3 - \E(\hat T_3) | > \epsilon/4 \}) 	\\
		\coloneqq& E_1+E_2+E_3,
	\end{split}
\end{equation*}
where $E_1=P(\{ | \hat{S}_1 - \E(\hat T_1) | > \epsilon/4 \})$, $E_2=P(\{ | \hat{S}_2 - \E(\hat T_2)  | > \epsilon/4 \})$, $E_3= P(\{ | \hat{S}_3 - \E(\hat T_3)  | > \epsilon/4 \}) $. In the following we will estimate the upper bounds of $E_1$, $E_2$ and $E_3$, respectively.

Let $\mathbb{I}(\cdot)$ be an indicator function. Denote $\tilde S_1=\frac{12}{n^5} \sum_{i,j,k,\ell,u=1}^{n} \Vert Y_i - Y_j \Vert_2 \Vert Z_i - Z_j \Vert_2\mathbb{I}(\Vert Y_i - Y_j \Vert_2 \Vert Z_i - Z_j \Vert_2\le M) K_{iu}K_{ju}K_{ku}K_{\ell u}$, $\tilde S_2=\frac{12}{n^5} \sum_{i,j,k,\ell,u=1}^{n} \Vert Y_i - Y_j \Vert_2 \Vert Z_k - Z_\ell \Vert_2\mathbb{I}(\Vert Y_i - Y_j \Vert_2 \Vert Z_k - Z_\ell \Vert_2\le M) K_{iu}K_{ju}K_{ku}K_{\ell u}$ and $\tilde S_3=\frac{12}{n^5} \sum_{i,j,k,\ell,u=1}^{n} \Vert Y_i - Y_j \Vert_2 \Vert Z_i - Z_k \Vert_2\mathbb{I}(\Vert Y_i - Y_j \Vert_2 \Vert Z_i - Z_k \Vert_2\le M) K_{iu}K_{ju}K_{ku}K_{\ell u}$.

Define 
\begin{align*}
\alpha_{ijk\ell u}=& \Vert Y_i - Y_j \Vert_2 \Vert Z_i - Z_j \Vert_2\mathbb{I}(\Vert Y_i - Y_j \Vert_2 \Vert Z_i - Z_j \Vert_2\le M) K_{iu}K_{ju}K_{ku}K_{\ell u},\\
\beta_{ijk\ell u}=&\Vert Y_i - Y_j \Vert_2 \Vert Z_k - Z_\ell \Vert_2\mathbb{I}(\Vert Y_i - Y_j \Vert_2 \Vert Z_k - Z_\ell \Vert_2\le M) K_{iu}K_{ju}K_{ku}K_{\ell u},\\
\gamma_{ijk\ell u}=&\Vert Y_i - Y_j \Vert_2 \Vert Z_i - Z_k \Vert_2\mathbb{I}(\Vert Y_i - Y_j \Vert_2 \Vert Z_i - Z_k \Vert_2\le M) K_{iu}K_{ju}K_{ku}K_{\ell u},
\end{align*} and 
\begin{align*}
\hat \alpha_{ijk\ell u}=&\frac15(\alpha_{ijk\ell u}+\alpha_{ujk\ell i}+\alpha_{iuk\ell j}+\alpha_{iju\ell k}+\alpha_{ijku\ell}),\\
\hat \beta_{ijk\ell u}=&\frac15(\beta_{ijk\ell u}+\beta_{ujk\ell i}+\beta_{iuk\ell j}+\beta_{iju\ell k}+\beta_{ijku\ell}),\\
\hat \gamma_{ijk\ell u}=&\frac15(\gamma_{ijk\ell u}+\gamma_{ujk\ell i}+\gamma_{iuk\ell j}+\gamma_{iju\ell k}+\gamma_{ijku\ell}).
\end{align*}  
Then the corresponding $U$-statistic of $\tilde S_1$, $\tilde S_2$ and $\tilde S_3$ are 
\[ \tilde T_1=\frac{12}{\tbinom{n}{5}}\sum_{i<j<k<\ell<u}\hat \alpha_{ijk\ell u},\ \tilde T_2=\frac{12}{\tbinom{n}{5}}\sum_{i<j<k<\ell<u}\hat \beta_{ijk\ell u},\  \tilde T_3=\frac{12}{\tbinom{n}{5}}\sum_{i<j<k<\ell<u}\hat \gamma_{ijk\ell u}.\]

Next we show the following statement. \emph{For any $\epsilon>0$, there exist constants $C_1,C_{1,\epsilon}>0$ such that: $E_1 
	\le \frac{C_1}n+4nC_{1,\epsilon}\exp\left(-\sqrt{\frac n{\log n}}\right) $.
}\\
Denote $M=\sqrt{\frac n{\log n}}\frac{\epsilon}{960}$ as a constant.	
Define sets 
\begin{equation*}\label{sets}
	\begin{aligned}
		&	G_1 = \{ | \tilde{S}_1 - \E(\hat T_1) | \le \epsilon/4 \},		\	G_2 = \{ \tilde{S}_1 = \hat{S}_1,\tilde{T}_1 = \hat{T}_1 \},		\	G_3 = \{ \Vert Y_i -Y_j \Vert_2\Vert Z_k - Z_\ell \Vert_2  \le M, \; \forall i,j,k,\ell \},		\\
		&	G_4 = \{ \Vert Y_i \Vert_2^2 + \Vert Z_j \Vert_2^2  \le M_1/2, \; \forall i,j \}	,	\	G   = \{ | \hat{S}_1 - \E(\hat T_1) | \le \epsilon/4 \},
	\end{aligned}
\end{equation*}
we have
\begin{equation*}
	G_1 \cap G_2 \subset G,\ 		
	G_3 \subset G_2,\ 		
	G_4 \subseteq G_3,	
\end{equation*}
where $G_4 \subseteq G_3$ can obtain since 
\begin{equation*}
	\begin{split}
		& \Vert Y_i -Y_j \Vert_2 \Vert Z_k - Z_\ell \Vert_2 \le \frac{ \Vert Y_i - Y_j \Vert_2^2+\Vert Z_k - Z_\ell \Vert_2^2 }{2}		\\
		=& \frac{1}{2} \left( \Vert Y_i \Vert_2^2 + \Vert Y_j \Vert_2^2  - 2Y^{\top}_iY_j + \Vert Z_k \Vert_2^2 + \Vert Z_\ell \Vert_2^2  - 2Z^{\top}_kZ_\ell \right)		\\
		\le&  \left(\Vert Y_i \Vert_2^2 + \Vert Y_j \Vert_2^2+\Vert Z_k \Vert_2^2 + \Vert Z_\ell \Vert_2^2   \right)\le 2\ \underset{i,j}{\mathrm{max}} \: (\Vert Y_i \Vert_2^2+\Vert Z_j \Vert_2^2 ), \ \forall i,j,k,\ell.
	\end{split}
\end{equation*}
Thus, $G_3^c \subseteq G_4^c$  ($ \I^c$ denote the complementary set of the set $\I$).

By the inclusion relation of the sets, we establish the following probability inequalities:

\begin{equation}\label{ps1_cdc}
	\begin{aligned}
		P(G^c) &\le P(G_1^c \cup G_2^c)  = P\left(G_1^c \cap (G_2 \cup G_2^c) \cup G_2^c \right)		
		= P\left( (G_1^c \cap G_2) \cup (G_1^c \cap G_2^c ) \cup G_2^c \right)		\\
		& \le P\left( (G_1^c \cap G_2) \cup G_2^c \right)		
		 \le P (G_1^c \cap G_2) + P(G_2^c) 	
		 \le P (G_1^c \cap G_2) + P(G_3^c)		
		 \le P (G_1^c \cap G_2) + P(G_4^c).
	\end{aligned}
\end{equation}

Note that the kernel size $m=2$, by Lemma \ref{u-statistic:covergence}, there exists a constant $C_1>0$ we obtain
\begin{align}
	P(|\tilde T_1 - \E(\tilde{T}_1)|\ge \epsilon/8) & \le 2\exp\left\{-2[n/m]\frac{\epsilon^2}{64}/\left(\frac{n\epsilon^2}{(960)^2\log n}\right)\right\} = \frac {2}{n^{120}} \label{u-stat-cdc}.
\end{align}
Define $\I_1=\{(i,j,k,\ell,u)|1\le i<j<k<\ell<u\le n\}$ and $\I_2=\{(i,j,k,\ell,u)|1\le i\le j\le k\le \ell\le u\le n\}$, then 
\begin{align*}
	0&<\tilde T_1 - \tilde{S}_1  \le \frac{5!\times 12}{n(n-1)(n-2)(n-3)(n-4)} \sum_{(i,j,k,l,u)\in \I_1} \hat \alpha_{ijk\ell u} - \frac{5!\times 12}{n^5} \sum_{(i,j,k,l,u)\in \I_2} \hat \alpha_{ijk\ell u}		\\
	& = 12\times\left( \frac{5!}{n(n-1)(n-2)(n-3)(n-4)} - \frac{5!}{n^5} \right) \sum_{(i,j,k,l,u)\in \I_1} \hat \alpha_{ijk\ell u} - \frac{5!\times 12}{n^5}\sum_{(i,j,k,l,u)\in \I_2/\I_1}\hat \alpha_{ijk\ell u}		\\
	& \le  \frac{5!\times12(10n^3-35n^2+50n-24)}{n^5(n-1)(n-2)(n-3)(n-4)}  \sum_{(i,j,k,l,u)\in \I} \hat \alpha_{ijk\ell u}		\\
	& \le12 \frac{10n^3-35n^2+50n-24}{n^4} M \le \frac{120}{n} \cdot \frac{\epsilon\sqrt n} {960\sqrt{\log n}} <\frac\epsilon 8,
\end{align*}
where the last second inequality holds since $35n^2-50n+24>0,\ \forall n$. Therefore, $P(\{ | \tilde{S}_1 -\tilde T_1 | > \frac{\epsilon}{8} \} = 0$. Thus, from (\ref{u-stat-cdc}), get
\begin{equation}\label{pu1_cdc}
	\begin{aligned}
		P(G_1^c \cap G_2) =  P(\{ | \tilde{S}_1 - \E(\tilde{T}_1) | >\frac \epsilon 4\}) & = P(\{ | \tilde{S}_1 -\tilde T_1 + \tilde T_1- \E(\tilde{T}_1) | > \frac \epsilon 4\})		\\
		& \le P(\{ | \tilde{S}_1 -\tilde T_1 | > \frac{\epsilon}{8} \} ) + P(\{ | \tilde T_1- \E(\tilde{S}_1) | > \frac{\epsilon}{8} \})			
		 \le \frac {2}{n^{120}}.
	\end{aligned}
\end{equation}

Next, we estimate $P(G_4^c)$. There exists a constant $C_{1,\epsilon}>0$, we have
\begin{equation}\label{pg4_cdc}
	\begin{aligned}
		&P(G_4^c)  = P(\{\Vert Y_i \Vert_2^2+\Vert Z_j \Vert_2^2 > M_1/2 , \; \forall i,j \} )		\\
		\le& P\left(\left\{ \Vert Y_i \Vert_2^2 > \frac{M}{4}, \; \forall i \right\}\right) + P\left(\left\{ \Vert Z_j \Vert_2^2 > \frac{M}{4}, \; \forall j \right\}\right)		\\
		=& P\left(\left\{ \Vert Y_i \Vert_2 > \frac{\sqrt{M}}{2}\; \forall i \right\} \right) + P\left(\left\{ \Vert Z_j \Vert_2 > \frac{\sqrt{M}}{2}, \; \forall j \right\} \right), \\
		\le& 2en\exp\left(-\frac{M}{4K_1}\right)\le 2en\cdot\exp\left(-\frac{\epsilon}{3840K_1}\sqrt{\frac n{\log n}}\right),
	\end{aligned}
\end{equation}
where the last inequality is based on Lemma \ref{lem:subgaussian}, since $Y_i$, $Z_j$  are the Sub-Gaussian vectors. Thus, combining (\ref{ps1_cdc}), (\ref{pu1_cdc}), (\ref{pg4_cdc}), we have
\begin{equation*}
	\begin{aligned}
		P(\{ | \hat{S}_1 - \E(\hat T_1) | > \epsilon \}) = P(G^c)\le P (G_1^c \cap G_2) + P(G_4^c)		
		\le  \frac {2}{n^{120}}+ 2en\cdot\exp\left(-\frac{\epsilon}{3840K_1}\sqrt{\frac n{\log n}}\right).
	\end{aligned}
\end{equation*}
%############################
Similarly, we can show: \emph{For any $\epsilon>0$, we have $E_2 \le \frac {2}{n^{120}}+ 2en\cdot\exp\left(-\frac{\epsilon}{3840K_1}\sqrt{\frac n{\log n}}\right) $ and $E_3 \le \frac {2}{n^{120}}+ 2en\cdot\exp\left(-\frac{\epsilon}{3840K_1}\sqrt{\frac n{\log n}}\right) $}. In summary, we conclude the result.
\end{proof}
%######################################################################################################

\section{Additional Details about Numerical Results}
\label{appendix:exp}
The experiments are conducted in a Linux environment using the PyTorch library, utilizing the computational capabilities of a NVIDIA A100 Tensor Core GPU. In the classification task, the cross-entropy loss function is employed. The batch size for tabular datasets is set to 1024, while for image datasets, it is set to 256. The initial learning rate is 0.1 and the bandwidth is $h = (n \frac{r+2}{4})^{\frac{-1}{r+4}}$ for both cases.

\subsection{Adult}
We followed the preprocessing procedures outlined in \cite{han2023ffb}, and the sensitive attribute chosen is `gender'. 
During the training process, Stochastic Gradient Descent (SGD) with momentum which is set to $0.9$ is utilized as the optimization algorithm. The model is trained for a total of 40 epochs, with the learning rate decayed by a factor of 10 at the 15th and 30th epochs.

In our experiments, we employ a dataset splitting strategy that involved dividing the data into training, validation, and testing sets. The proportions used for the splits are 70\% for training, 15\% for validation, and 15\% for testing. To address the potential variability introduced by different dataset splits, we apply 10 different random seeds for the splitting process. For each split, we conduct the experiments and record the results. The reported results are the average over multiple experiments.

\subsection{ACSIncome}

ACSIncome is a dataset to predict whether an individual’s income is above \$50,000, after filtering the 2018 US-wide American Community Survey (ACS) Public Use Microdata Sample (PUMS) data sample to only include individuals above the age of 16, who reported usual working hours of at
least 1 hour per week in the past year, and an income of at least \$100. We follow the preprocessing procedures outlined as \cite{han2023ffb} and the sensitive attribute chosen is `race'.

\subsection{CelebA}
The CelebA dataset consists of over 200,000 celebrity images, with annotations for various attributes such as gender, age, and presence of facial hair. The CelebA dataset is known for its imbalanced class distribution, particularly with attributes such as gender, where the number of male and female samples is significantly different. This dataset presents a challenge for fairness evaluation due to the inherent bias in the data. 

In our experiments on the CelebA dataset, the dataset was divided into training, validation, and testing sets, with sizes of $162k$, $18k$, and $19k$ samples, respectively. To encode the input data and extract meaningful representations, we utilized the ResNet-18 architecture. Following the encoding step, we employed a two-layer neural network for prediction. The neural network consisted of a hidden layer with a size of 128 neurons, and the Rectified Linear Unit (ReLU) activation function was applied. 

During the training process, we utilized Stochastic Gradient Descent (SGD) with momentum which is set to $0.9$ as the optimization algorithm. The model was trained for a total of 90 epochs. The learning rate was decayed by a factor of 10 at the 20th, 40th, and 60th epochs. This learning rate decay strategy helps facilitate convergence and improve the model's performance over the course of training.

In this paper, we focus on analyzing four groups of training and test datasets that contain binary sensitive attributes. Each group consists of a total of 162,770 training samples and 19,962 test samples.

To provide detailed insights into the dataset characteristics, we present  the sample numbers and proportions for different target/sensitive attribute settings in Table \ref{dataset:celeba}. Additionally, Table \ref{dataset:celeba2} provides the sample numbers and proportions for different target/sensitive attribute settings, specifically focusing on scenarios involving two sensitive attributes. 

\begin{table*}[htbp]
\centering
\caption{Compositions of the CelebA datasets with binary sensitive attribute.}\label{dataset:celeba}	
        \resizebox{1\columnwidth}{!}{
	\renewcommand\arraystretch{1.3}
\begin{tabular}{c|cc|cc|cc} 
	\toprule
	Training          & \multicolumn{2}{c|}{162770}            & \multicolumn{2}{c|}{162770}           & \multicolumn{2}{c}{162770}            \\
	& \multicolumn{2}{c|}{Attractive/gender} & \multicolumn{2}{c|}{Wavy hair/gender} & \multicolumn{2}{c}{Attractive/young}  \\ 
	\cmidrule(r){2-7}
	& Target c0       & Target c1            & Target c0       & Target c1           & Target c0       & Target c1           \\ 
	\hline
	Sensitive Class 0 & 29920 (18.38\%) & 49247 (30.26\%)      & 52289 (32.12\%) & 58499 (35.94\%)     & 30618 (18.81\%) & 48549 (29.83\%)     \\
	Sensitive Class 1 & 64589 (39.68\%) & 19014 (11.68\%)      & 42220 (25.94\%) & 9762 (6.00\%)       & 5364 (3.30\%)   & 78239 (48.07\%)     \\ 
	\hline\hline
	Testing           & \multicolumn{2}{c|}{19962}             & \multicolumn{2}{c|}{19962}            & \multicolumn{2}{c}{19962}             \\
	& \multicolumn{2}{c|}{Attractive/gender} & \multicolumn{2}{c|}{Wavy hair/gender} & \multicolumn{2}{c}{Attractive/young}  \\ 
	\cmidrule(r){2-7}
	& Target c0       & Target c1            & Target c0       & Target c1           & Target c0       & Target c1           \\ 
	\hline
	Sensitive Class 0 & 4263 (21.36\%)  & 5801 (29.06\%)       & 6178(30.95\%)   & 6517 (32.65\%)      & 4066 (20.37\%)  & 5998 (30.05\%)      \\
	Sensitive Class 1 & 7984(40.00\%)   & 1914 (9.59\%)        & 6069(30.40\%)   & 1198 (6.00\%)       & 782 (3.92\%)    & 9116 (45.67\%)      \\
	\bottomrule
\end{tabular} }
	\end{table*}

\begin{table}[htbp]
\caption{Compositions of the CelebA datasets with multiple sensitive attributes.}\label{dataset:celeba2}
		\centering
		\renewcommand\arraystretch{1.2}
		\begin{tabular}{ll|cccc}
			\toprule
			Training &                  & \multicolumn{4}{c}{Sensitive Attribute}                             \\ \hline
			& Target Attribute & Female and Old & Female and Young & Male and Old   & Male and Young \\ 
            \cmidrule(r){3-6} 
			& Attractive       & 7522 (4.62\%)   & 23096 (14.19\%)   & 22398 (13.76\%) & 26151 (16.07\%) \\
			& Unattractive     & 3645 (2.24\%)   & 1719 (1.06\%)     & 60944 (37.44\%) & 17295 (10.63\%) \\ \hline
			Testing & Attractive       & 1299 (6.51\%)   & 2767 (13.86\%)    & 2964 (14.85\%)  & 3034 (15.20\%)  \\
			& Unattractive     & 617 (3.09\%)    & 165 (0.83\%)      & 7367 (36.91\%)  & 1749 (8.76\%)  \\
			\bottomrule
	\end{tabular}
\end{table}

\begin{table}[htbp]
\caption{Compositions of the UTKFace datasets with binary sensitive attributes.}
	\label{dataset:utkface}
	\centering
	\renewcommand\arraystretch{1}
\begin{tabular}{c|cc|cc} 
	\toprule
	Training          & \multicolumn{2}{c|}{17119}             & \multicolumn{2}{c}{17119}             \\
	& \multicolumn{2}{c|}{Age/gender}        & \multicolumn{2}{c}{Age/race}          \\ 
	\cmidrule(r){2-5}
	& Target c0      & Target c1             & Target c0      & Target c1            \\ 
	\hline
	Sensitive Class 0 & 2514 (14.69\%) & 4302 (25.13\%)        & 3077(17.97\%)  & 3739 (21.84\%)       \\
	Sensitive Class 1 & 5695 (33.27\%) & 4608 (26.92\%)        & 6743 (39.39\%) & 3560 (20.80\%)       \\ 
	\hline\hline
	Testing           & \multicolumn{2}{c|}{3556}              & \multicolumn{2}{c}{3556}              \\
	& \multicolumn{2}{c|}{Age/gender} & \multicolumn{2}{c}{Age/race}  \\ 
	\cmidrule(r){2-5}
	& Target c0      & Target c1             & Target c0      & Target c1            \\ 
	\hline
	Sensitive Class 0 & 486 (13.67\%)  & 958 (26.94\%)         & 655(18.42\%)   & 789 (22.19\%)        \\
	Sensitive Class 1 & 1173(32.99\%)  & 939 (26.41\%)         & 1433(40.30\%)  & 679 (19.09\%)        \\
	\bottomrule
	\end{tabular}
\end{table}

\subsection{UTKFace}
\label{appendix:utk}
In our approach, we utilize the ResNet-18 architecture to encode the input data into a representation of dimension 100. After encoding the input data, we employ a two-layer neural network for prediction. This neural network consists of a hidden layer with a size of 128 neurons and applies the Rectified Linear Unit (ReLU) activation function. The training model we use is the cross-entropy loss.

During the training process, we employ Stochastic Gradient Descent (SGD) with momentum as the optimization algorithm. We train the model for a total of 90 epochs. To facilitate effective training, we implement a learning rate decay strategy. Specifically, we reduce the learning rate by a factor of ten at the 20th, 40th, and 60th epochs.

\section{Ablation Study}
\label{appendix:ablation}
\begin{figure}[htp]
	\vskip -0.2in
	\centering
	\caption{Comparison results on the tabular datasets by Lagrangian dual method (Lag dual) and using a fixed balancing parameter (w/o lag dual), where `w/o' stands for `without'.}
	\label{ablation:tabular}
	\begin{minipage}{0.4\linewidth}
		\centering
		\includegraphics[width=1\linewidth]{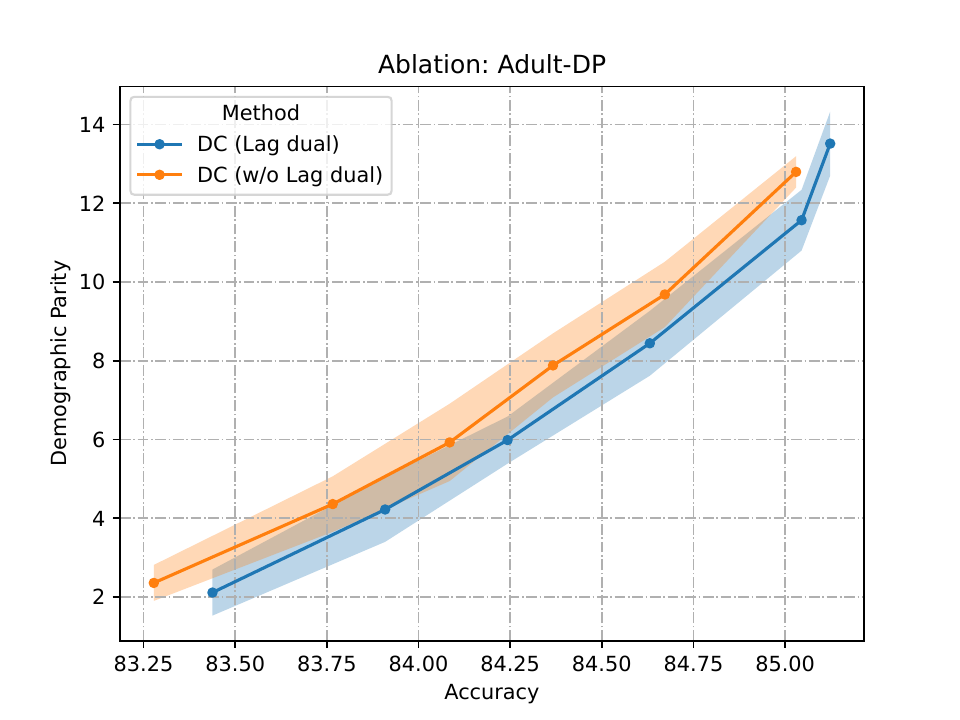}
	\end{minipage}
	\begin{minipage}{0.4\linewidth}
		\centering
		\includegraphics[width=1\linewidth]{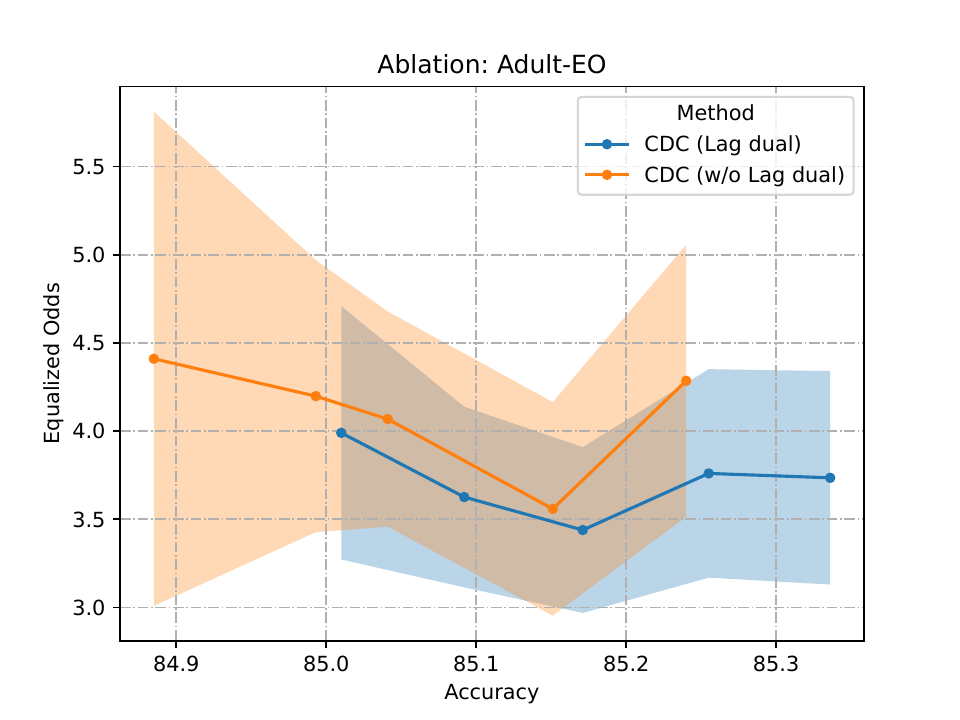}
	\end{minipage}
	\begin{minipage}{0.4\linewidth}
		\centering
		\includegraphics[width=1\linewidth]{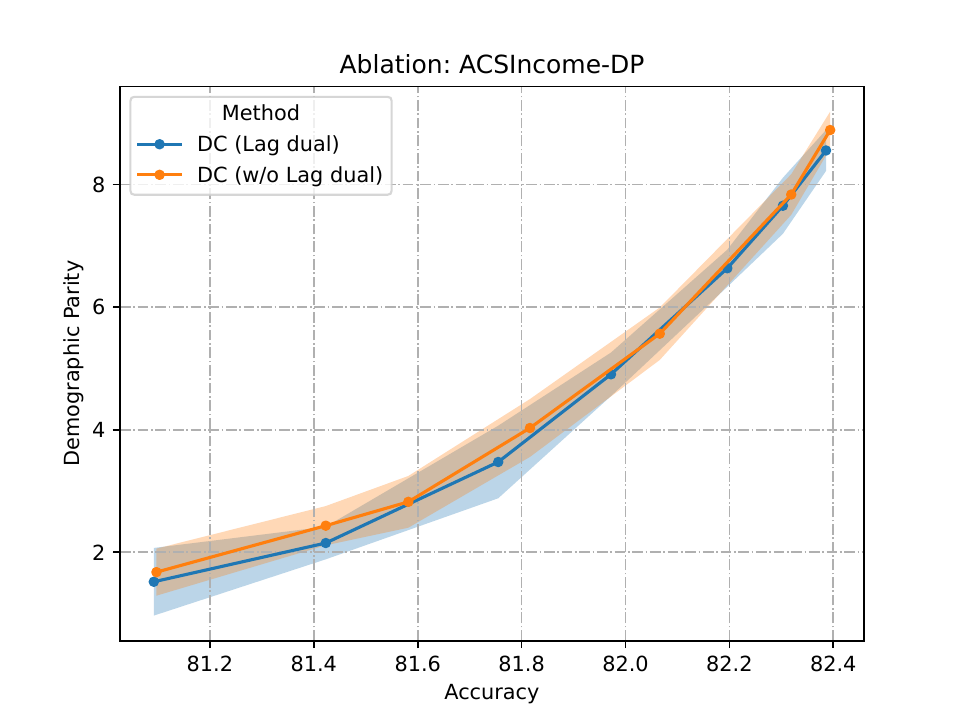}
	\end{minipage}
	\begin{minipage}{0.4\linewidth}
		\centering
		\includegraphics[width=1\linewidth]{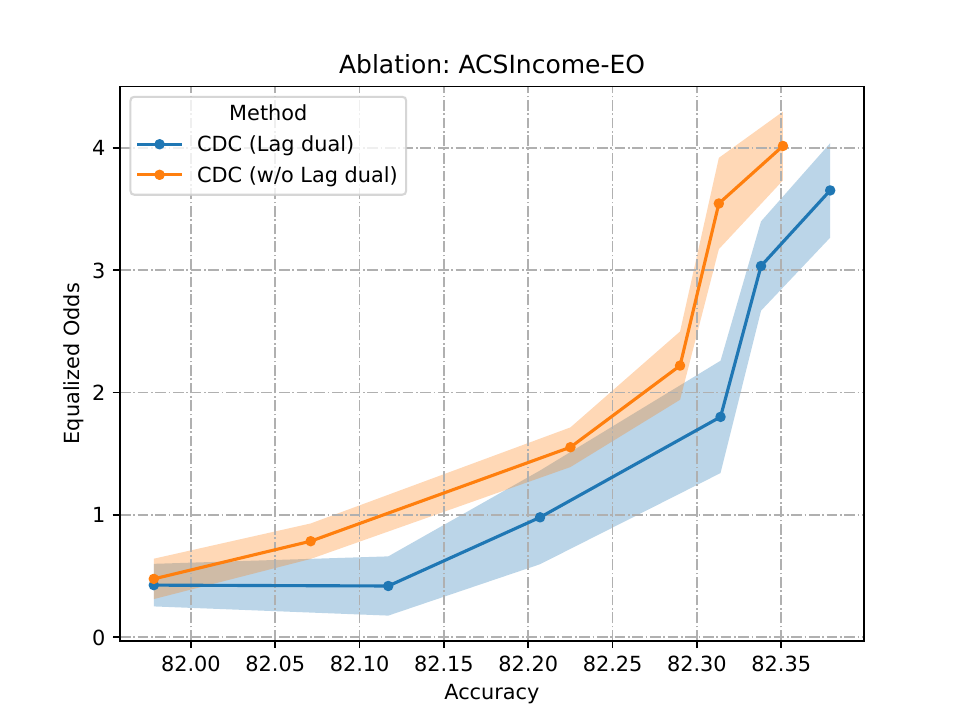}
	\end{minipage}
\end{figure}
In the ablation study, we aim to analyze the impact of dynamically adjusting the balancing parameter on the performance of our proposed Lagrangian dual algorithm. To compare the effectiveness of dynamic balancing parameter adjustment with using a fixed balancing parameter, we conduct experiments on Tabular datasets and two image datasets.

\subsection{Tabular Datasets}
In \cref{ablation:tabular}, we provide the comparison results on the tabular datasets (Adult and ACSIncome) for our proposed DC and CDC-based methods by Lagrangian dual method and using a fixed balancing parameter. These results show that incorporating the Lagrangian dual method consistently leads to improved fairness-accuracy trade-off curves under different fairness criteria. This indicates that the Lagrangian dual algorithm is effective in achieving a balance between fairness and accuracy on the tabular datasets. 

\begin{figure}[htp]
	\vskip -0.2in
	\centering
	\caption{Trends in accuracy and fairness metrics over 90 epochs for distance covariance (DC) on the CelebA dataset with specified hyperparameter settings.}
	\label{ablation:dc}
	\begin{minipage}{0.4\linewidth}
		\centering
		\includegraphics[width=1\linewidth]{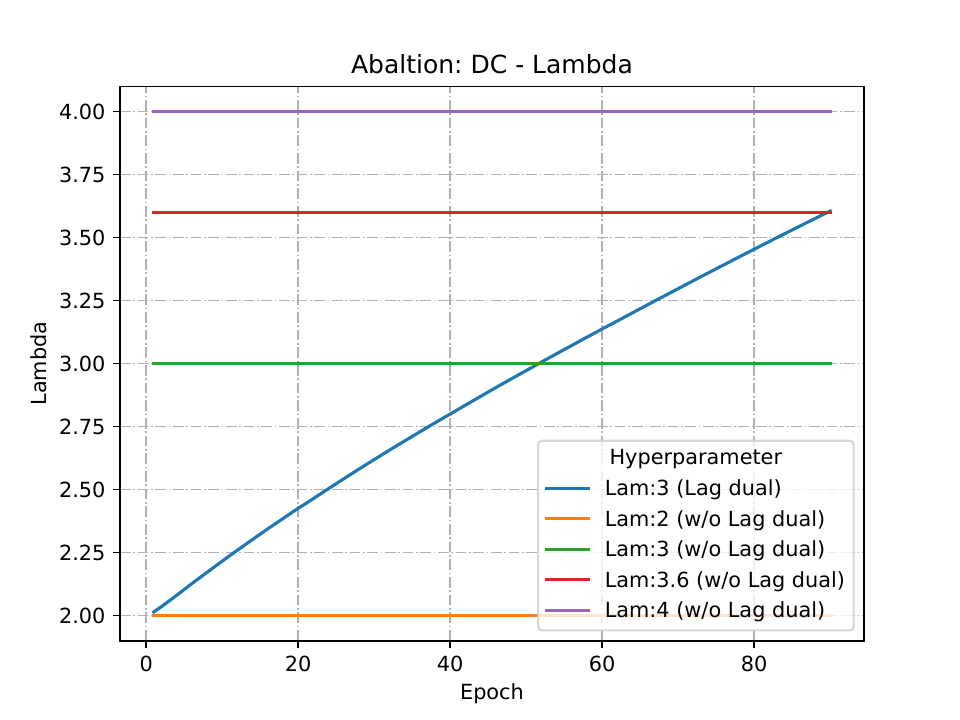}
	\end{minipage}
	\begin{minipage}{0.4\linewidth}
		\centering
		\includegraphics[width=1\linewidth]{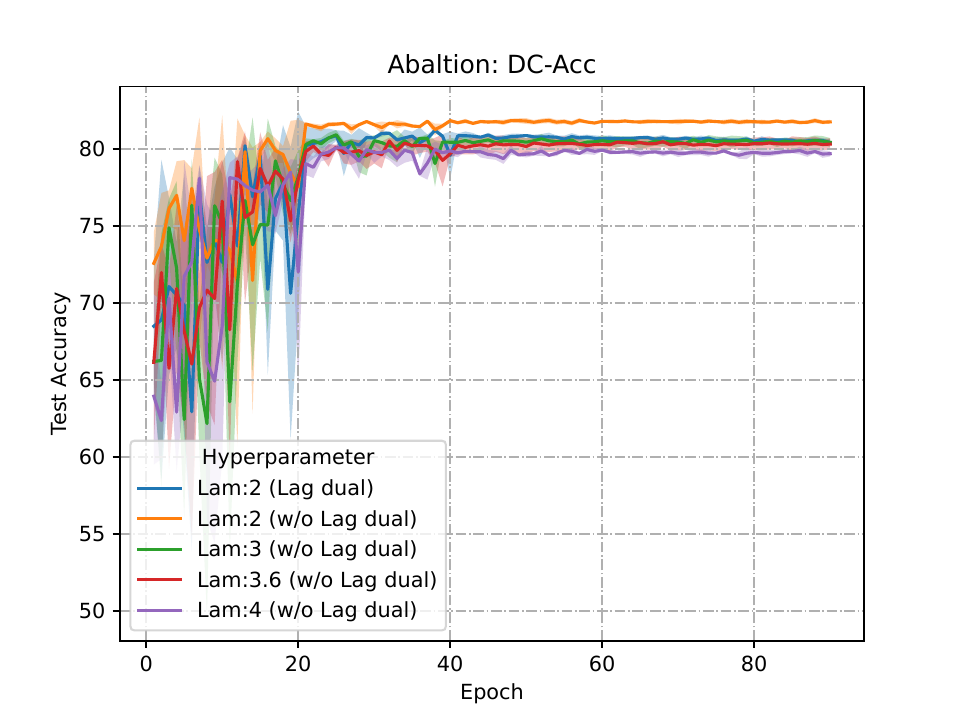}
	\end{minipage}
	\begin{minipage}{0.4\linewidth}
		\centering
		\includegraphics[width=1\linewidth]{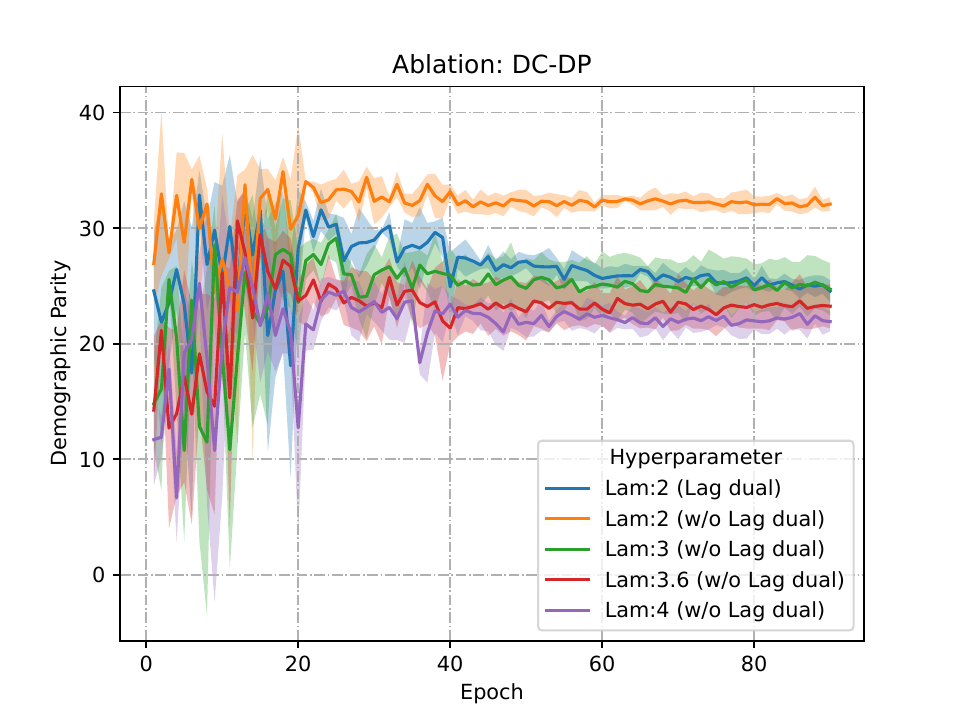}
	\end{minipage}
	\begin{minipage}{0.4\linewidth}
		\centering
		\includegraphics[width=1\linewidth]{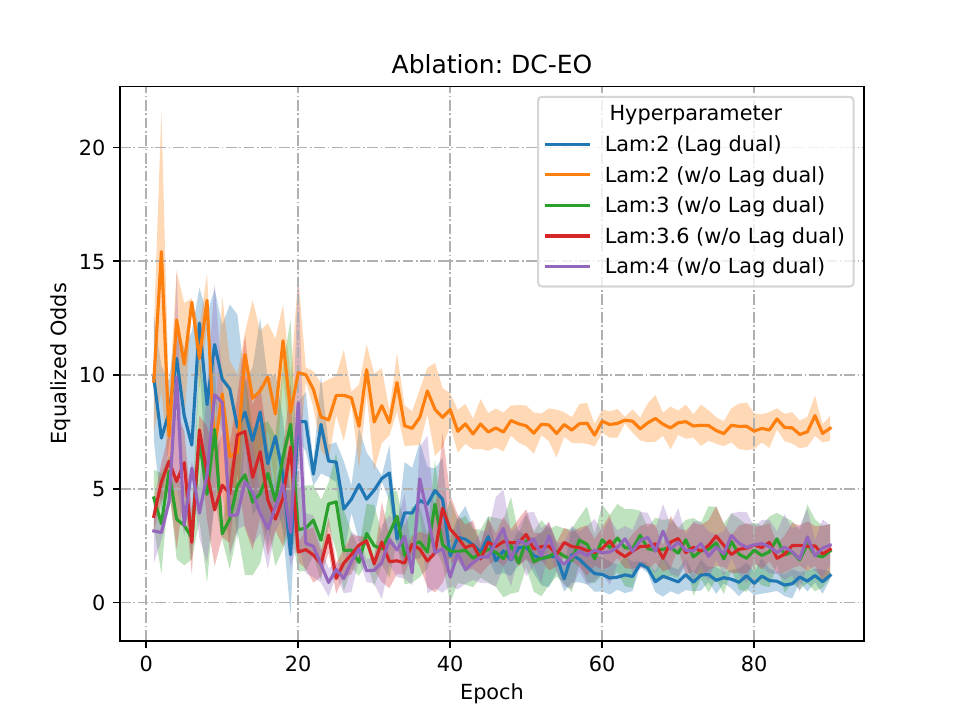}
	\end{minipage}
\end{figure}
\begin{figure}[htp]
	\vskip -0.2in
	\centering
	\caption{Trends in accuracy and fairness metrics over 90 epochs for conditional distance covariance (CDC) on the CelebA dataset with specified hyperparameter settings.}
	\label{ablation:cdc}
	\begin{minipage}{0.4\linewidth}
		\centering
		\includegraphics[width=1\linewidth]{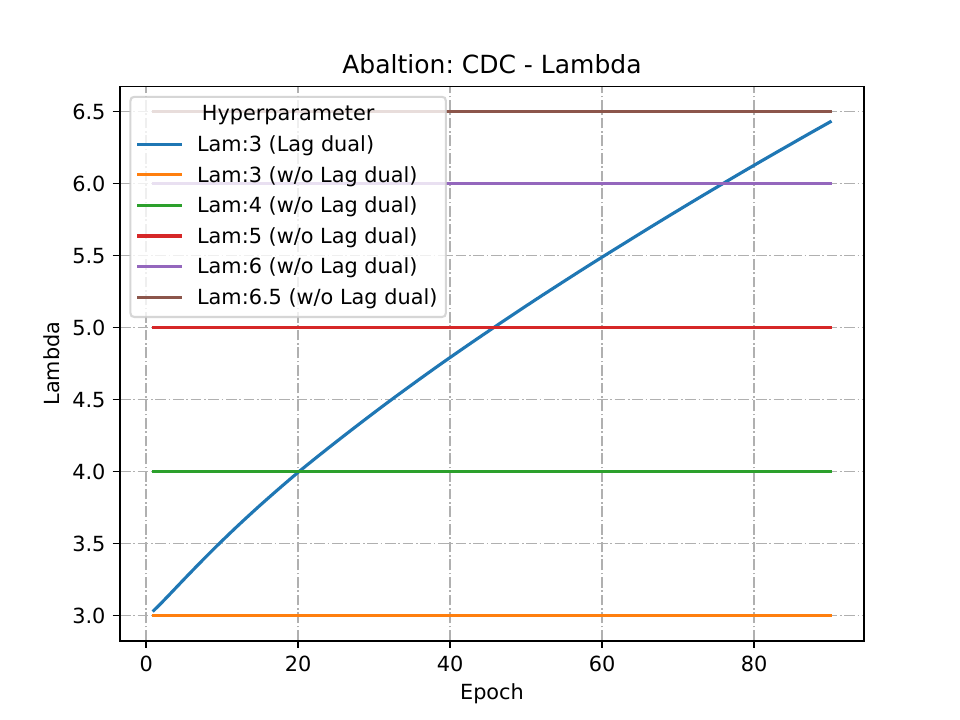}
	\end{minipage}
	\begin{minipage}{0.4\linewidth}
		\centering
		\includegraphics[width=1\linewidth]{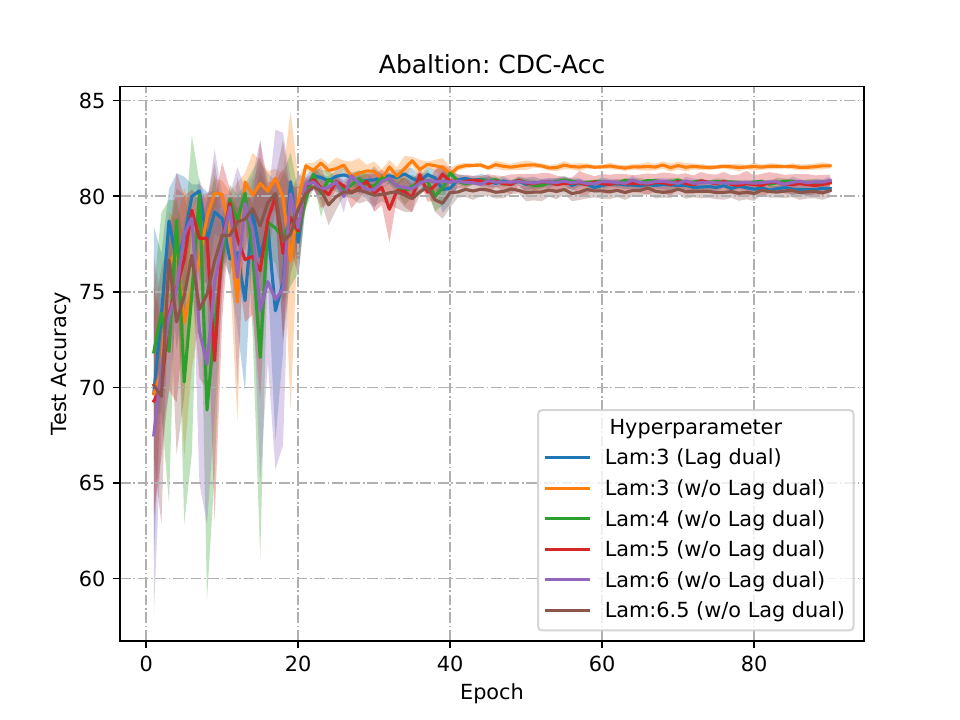}
	\end{minipage}
	\begin{minipage}{0.4\linewidth}
		\centering
		\includegraphics[width=1\linewidth]{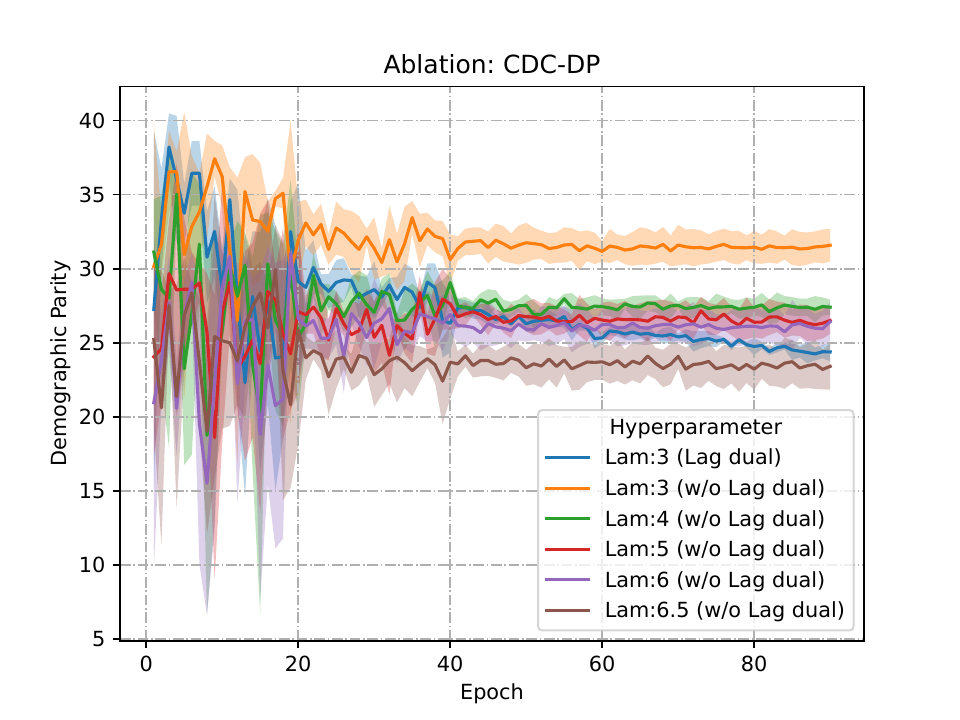}
	\end{minipage}
	\begin{minipage}{0.4\linewidth}
		\centering
		\includegraphics[width=1\linewidth]{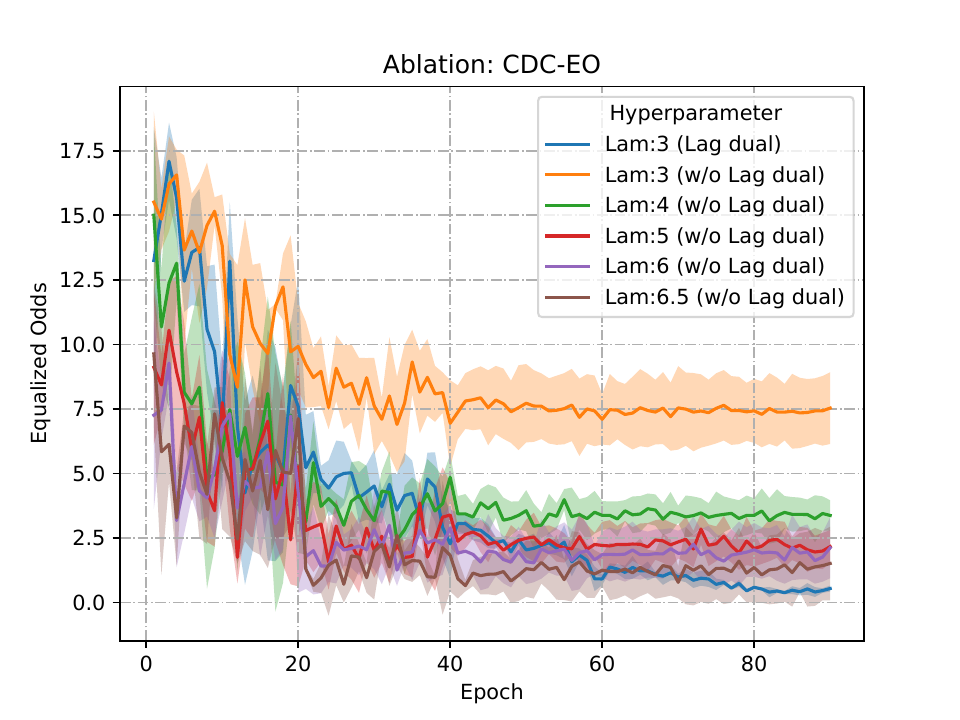}
	\end{minipage}
\end{figure}

\subsection{Image Datasets}
For the image datasets, we evaluate the performance by considering the test accuracy, DP, and EO metrics based on the training model at each epoch. The results presented reflect the evolution of test accuracy, DP, and EO throughout the training process.

For the CelebA dataset, our experiments focus on the `attractive' attribute as the target and the `gender' attribute as the sensitive attribute. When using distance covariance (DC) as a constraint term, we set the initial hyperparameters to $\lambda=2$ and $\beta=1$ for the Lagrangian dual method. We observe that $\lambda$ gradually increases from $2$ to $3.6$ during training (see the first subfigure in \cref{ablation:dc}). For comparison, we fix $\lambda$ to $2$, $3$, and $4$, respectively. In \cref{ablation:dc}, we illustrate the evolution of test DP, EO, and accuracy during training for these scenarios. %The results show that adaptively adjusting the balancing parameter leads to a better balance between accuracy and fairness.

When using conditional distance covariance (CDC) as a constraint term, we set initial hyperparameters to $\lambda=3$ and $\beta=5$ during training with the Lagrangian dual method. We find that $\lambda$ gradually increases from $3$ to $6.5$ during the whole training precedure (see the first subfigure in \cref{ablation:cdc}). For comparison, we fixed $\lambda$ at $3$, $4$, $5$, $6$, and $6.5$, respectively. The evolution of test DP, EO and accuracy  is depicted in \cref{ablation:cdc}. %during training for these scenarios.

\begin{figure}[htp]
	\vskip -0.2in
	\centering
	\caption{Trends in accuracy and fairness metrics over 90 epochs for distance covariance (DC) on the UTKFace dataset with specified hyperparameter settings.}
	\label{ablation:udc}
	\begin{minipage}{0.4\linewidth}
		\centering
		\includegraphics[width=1\linewidth]{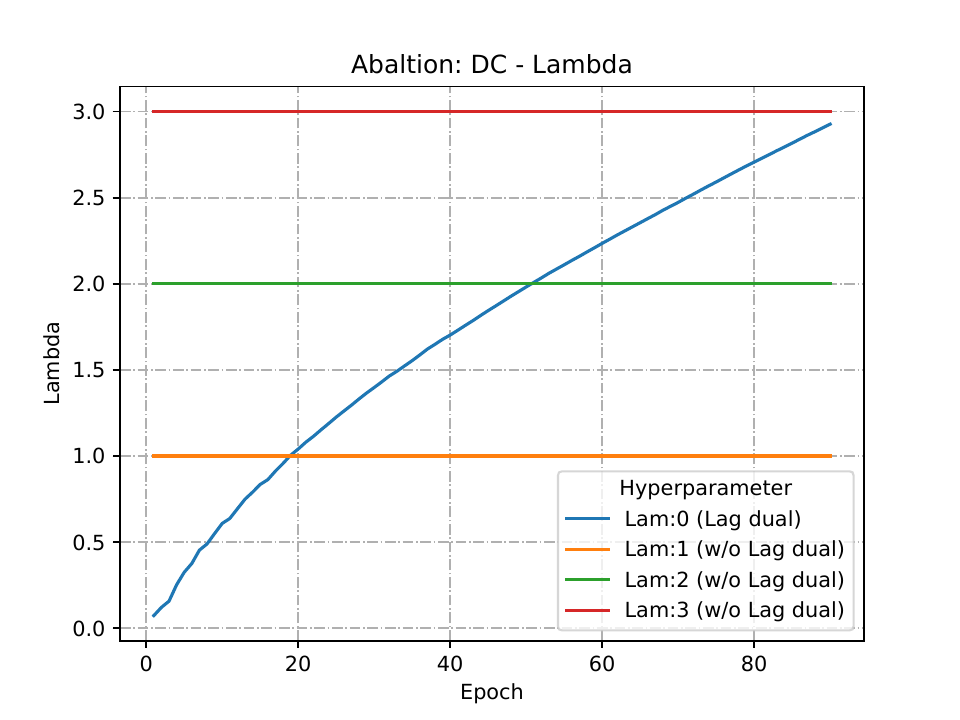}
	\end{minipage}
	\begin{minipage}{0.4\linewidth}
		\centering
		\includegraphics[width=1\linewidth]{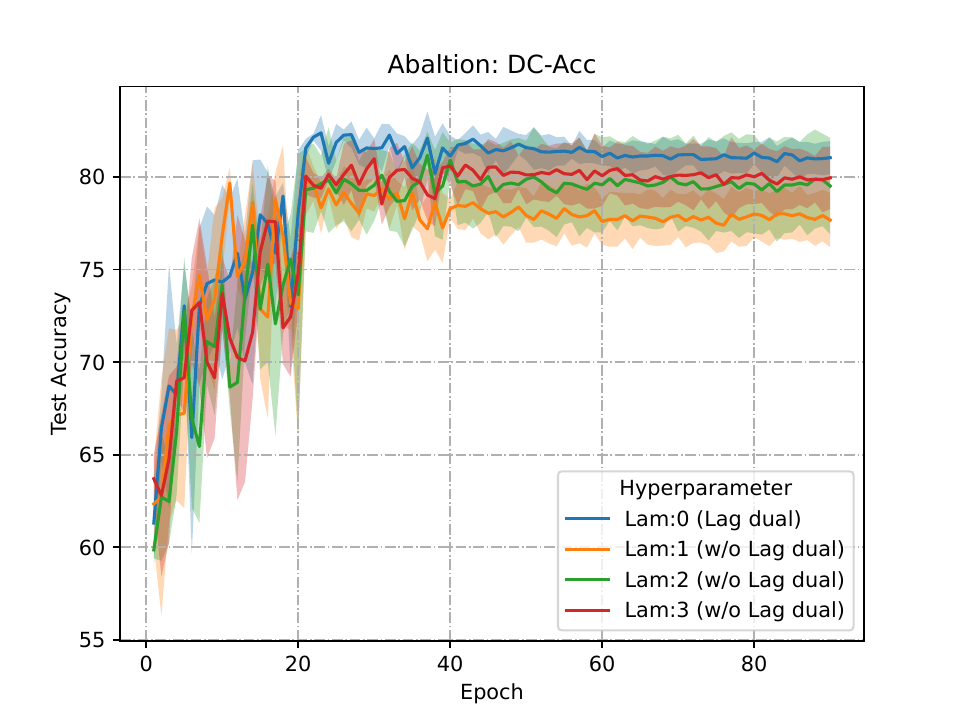}
	\end{minipage}
	\begin{minipage}{0.4\linewidth}
		\centering
		\includegraphics[width=1\linewidth]{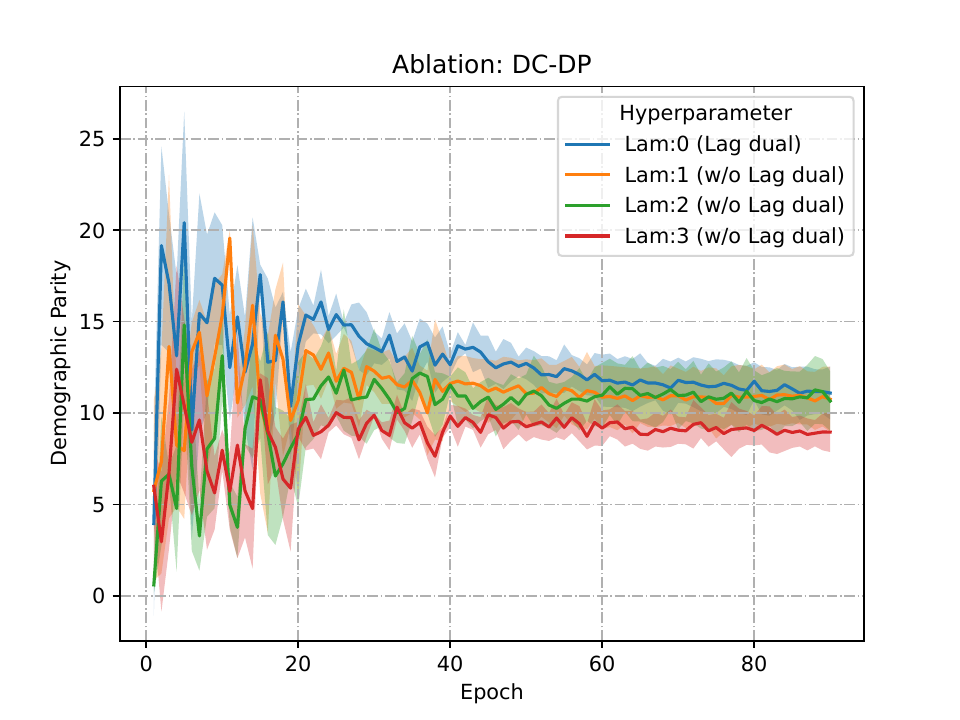}
	\end{minipage}
	\begin{minipage}{0.4\linewidth}
		\centering
		\includegraphics[width=1\linewidth]{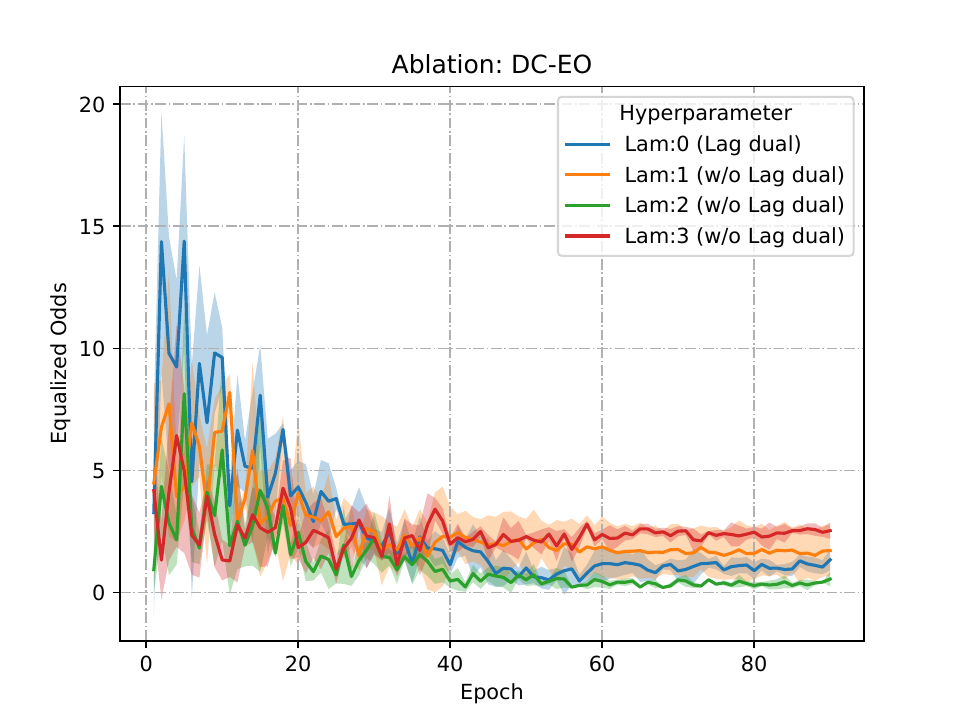}
	\end{minipage}
	\vskip -0.1in
\end{figure}

\begin{figure}[htp]
	\vskip -0.1in
	\centering
	\caption{Trends in accuracy and fairness metrics over 90 epochs for conditional distance covariance (CDC) on the UTKFace dataset with specified hyperparameter settings.}
	\label{ablation:ucdc}
	\begin{minipage}{0.4\linewidth}
		\centering
		\includegraphics[width=1\linewidth]{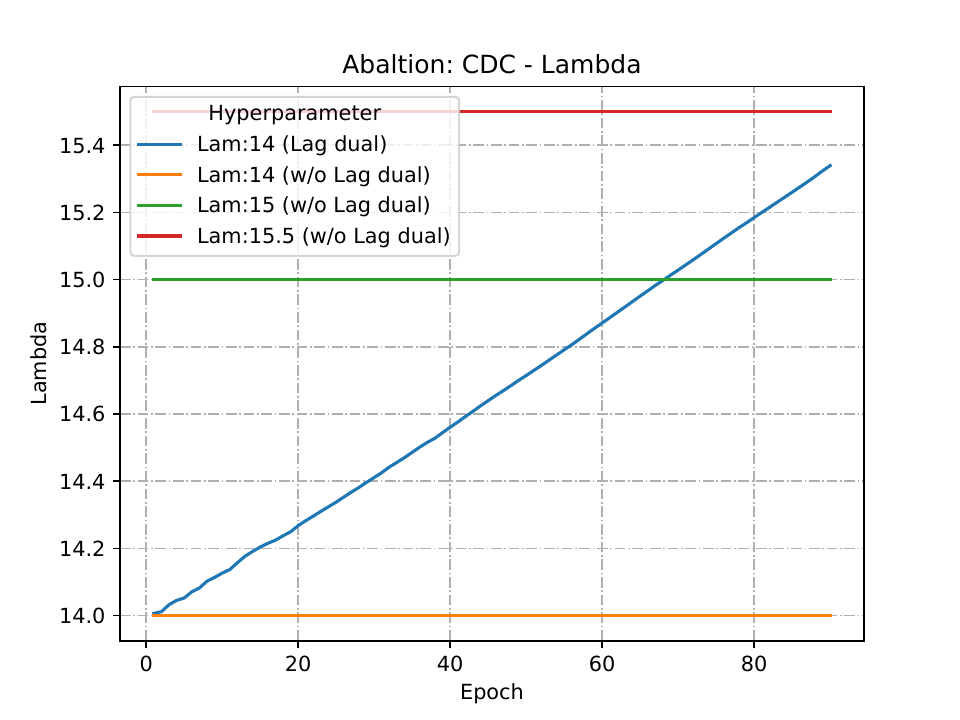}
	\end{minipage}
	\begin{minipage}{0.4\linewidth}
		\centering
		\includegraphics[width=1\linewidth]{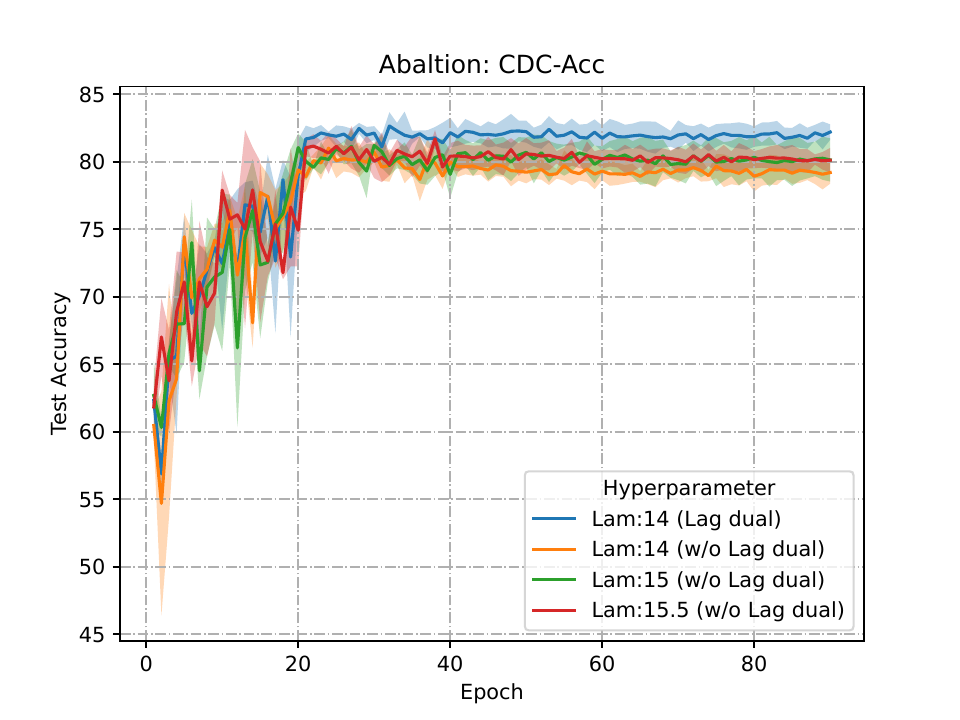}
	\end{minipage}
	\begin{minipage}{0.4\linewidth}
		\centering
		\includegraphics[width=1\linewidth]{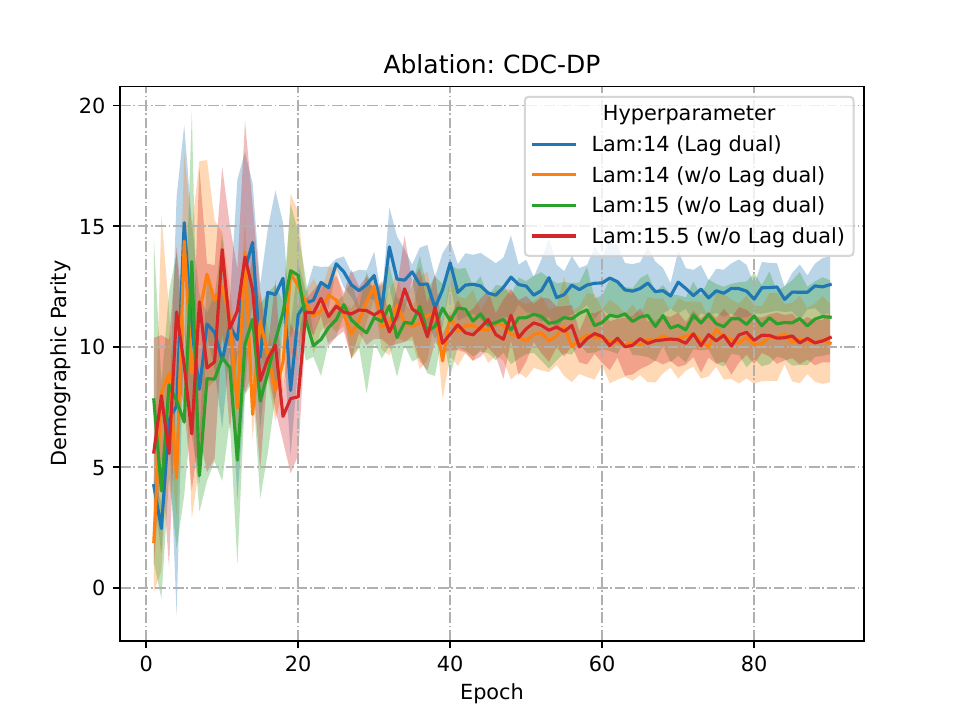}
	\end{minipage}
	\begin{minipage}{0.4\linewidth}
		\centering
		\includegraphics[width=1\linewidth]{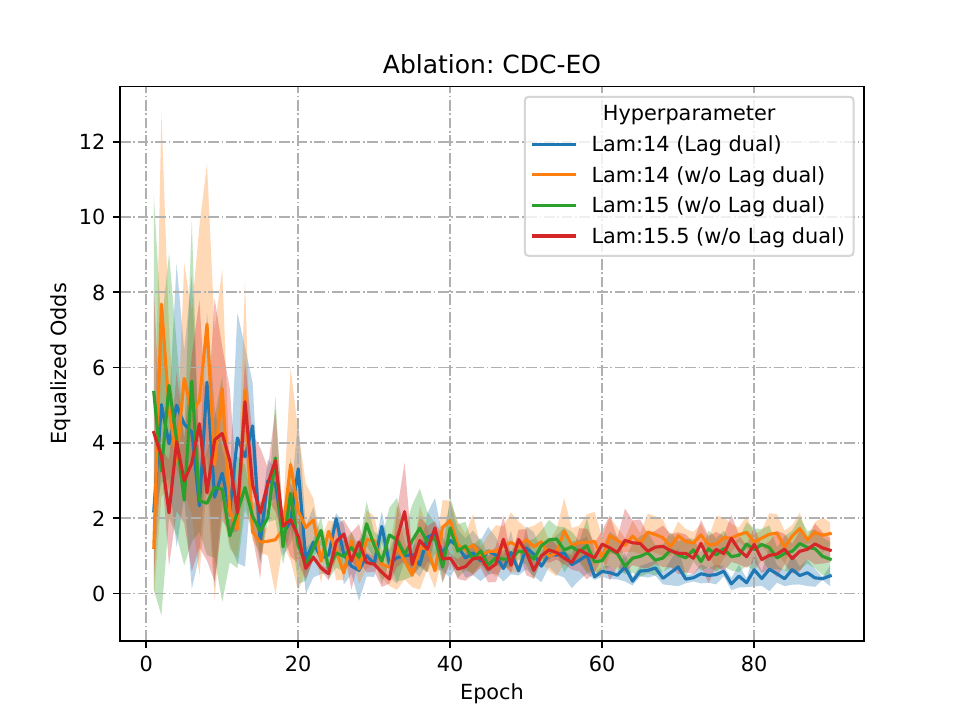}
	\end{minipage}
\end{figure}

For the UTKFace dataset, we select the `gender' attribute as the sensitive attribute. We set the initial hyperparameters as $\lambda=0$ and $\beta=2$ for DC, and $\lambda=14$ and $\beta=10$ for CDC in the Lagrangian dual method to solve the model. The changes in $\lambda$ for the DC and CDC constraints are shown in the first subfigures of \cref{ablation:udc} and \cref{ablation:ucdc}, respectively. For comparison, we choose $\lambda$ within the corresponding intervals for the DC and CDC constraints when using a fixed balancing parameter. The evolution of test accuracy, DP and EO are depicted in \cref{ablation:udc} and \cref{ablation:ucdc}.% during training for these scenarios. 

Through these experiments, we observed that the performance of the Lagrangian dual method achieves better trade-off results compared to methods using a fixed value of $\lambda$. This finding suggests that the Lagrangian dual method contributes to the stability of the optimization process, leading to improved results. 

In practice, users can determine their preference for fairness and accuracy by adjusting the balance parameter $\lambda$. Furthermore, the proposed combination with the Lagrangian dual method enables us to achieve a model that is as accurate as the original preference but more fair, or as fair as the original preference but more accurate.

\section{Complexity Analysis}\label{appendix:complexity}
Since the parameter updates of neural networks dominate the time/space complexity for single-step training (as other computations are negligible), we use the number of parameters (NPs) as a criterion to estimate the time/space complexity for different algorithms. 

 \textbf{Tabular datasets}: we use a four-layer Multilayer Perceptron (MLP) model with ReLU activation function and hidden layers $200,200,20$. Let $d_1,d_{21}$ be the dimensions of input and output, then the number of parameters for a four-layer MLP is $n_{11}=d_1\times200+200\times200+200\times20+20\times d_{21}$.
 
 \textbf{Image datasets}: To encode the input data and extract meaningful representations for image datasets, we utilized the ResNet-18 architecture. Following the encoding step, we employed a two-layer neural network for prediction. The neural network consisted of a hidden layer with a size of 128 neurons, and the Rectified Linear Unit (ReLU) activation function was applied. We use $n_{12}$ as the number of parameters of a ResNet18 and a two-layer MLP.

In the following we will give the notations and their descriptions used in various methods we compare.

 \begin{table}[ht]
    \centering
    \caption{Notations and their descriptions.}
    \label{table:notations}
    \resizebox{1\columnwidth}{!}{
	\renewcommand\arraystretch{1.1}
    \begin{tabular}{c|cl}
        \hline
        &Notations & Descriptions
        \\ \hline
         \multirow{11}{*}{Tabular} & $n_{11}$ & the number of parameters for a four-layer MLP in ours \\\cmidrule(r){2-3}
         & \multirow{2}{*}{$n_{21}$} & the number of parameters for the NNs used in the encoding from the predicted label to sensitive\\
         && attribute representation in FERMI \\\cmidrule(r){2-3}
         &$d_{21}$ & the dimension of  output\\\cmidrule(r){2-3}
         &$d_{31}$& the dimension of the one-hot encodings of the predicted label in FERMI\\\cmidrule(r){2-3}
         & $n_{31}$ & the number of parameters for tabular and image estimating means and variances in FairDisCo \\\cmidrule(r){2-3}
         & $n_{41}$ & the total number of parameters in the encoder and decoder networks used in Dist-Fair \\\cmidrule(r){2-3}
         & \multirow{2}{*}{$n_{51}$} & the total number of parameters in two encoders of both the predicted labels and the sensitive\\
         && attributes used in Soft-HGR \\\hline
         \multirow{11}{*}{Image} & $n_{12}$ & the number of parameters for a Resnet18 and a two-layer MLP in ours \\ \cmidrule(r){2-3}
         &\multirow{2}{*}{ $n_{22}$} & the number of parameters for the nn used in the encoding from the predicted label to sensitive\\
         && attribute representation in FERMI \\\cmidrule(r){2-3}
         &$d_{22}$ & the dimension of  output\\\cmidrule(r){2-3}
         &$d_{32}$ & the dimension of the one-hot encodings of the predicted label in FERMI\\\cmidrule(r){2-3}
         & $n_{32}$ & the number of parameters for tabular and image estimating means and variances in FairDisCo \\\cmidrule(r){2-3}
         & $n_{42}$ & the total number of parameters in the encoder and decoder networks used in Dist-Fair \\
         \cmidrule(r){2-3}
         & \multirow{2}{*}{$n_{52}$} & the total number of parameters in two encoders of both the predicted labels and the sensitive\\
         && attributes used in Soft-HGR \\\hline
    \end{tabular}
    }
\end{table}
 Table \ref{table:nps} reports the NPs used in each algorithm.
 \begin{table}[ht]
    \centering
    \caption{the NPs used in each algorithm.}
    \label{table:nps}
    \resizebox{1\columnwidth}{!}{
	\renewcommand\arraystretch{1.1}
    \begin{tabular}{ccccccccc}
        \hline
        \multicolumn{1}{c}{\diagbox{Dataset}{Method}} & HGR & Fair Mixup & FERMI & FairDisCo & Dist-Fair & Soft-HGR & DC & CDC
        \\ \hline
         Tabular & $n_{11}$ & $n_{11}$ & $n_{11}+d_{21}\times d_{31}+n_{21}$ & $n_{11}+n_{31}$ & $n_{41}$ &$n_{11}+n_{51}$& $n_{11}$ & $n_{11}$\\
         Image & $n_{12}$ & $n_{12}$ & $n_{12}+d_{22}\times d_{32} +n_{22}$& $n_{12}+n_{32}$ & $n_{42}$ & $n_{12}+n_{52}$ & $n_{12}$& $n_{12}$\\ \hline
    \end{tabular}
    }
\end{table}

In general, the numbers of parameters in encoder/decoder networks are much huger than a simple ResNet18.

\end{document}